\documentclass[acmtog,british,screen=true]{acmart}

\usepackage[utf8]{inputenc}
\usepackage[T1]{fontenc}
\usepackage[british]{babel}

\usepackage{pifont}
\newcommand{\cmark}{\ding{51}}
\newcommand{\xmark}{\ding{55}}

\usepackage{epsfig}
\usepackage{adjustbox} 
\graphicspath{{./figures/}}
\DeclareGraphicsExtensions{.pdf,.png,.jpeg,.jpg}

\usepackage{bm} 
\usepackage{esint} 
\usepackage{commath} 

\usepackage{siunitx} 
\usepackage{subcaption} 

\usepackage{booktabs} 
\usepackage{array} 

\usepackage{multirow} 
\usepackage{rotating} 

\usepackage{enumitem} 

\usepackage[capitalize]{cleveref} 
\Crefname{section}{Section}{Sections}
\crefname{section}{Sec.}{Secs.}
\Crefname{table}{Table}{Tables}
\crefname{table}{Tab.}{Tabs.}

\newcommand{\given}[2]{\left(#1\,\middle|\,#2\right)}

\renewcommand{\mid}{\,\ifnum\currentgrouptype=16 \middle\fi|\,}

\newcommand{\x}{\boldsymbol{x}}

\newcommand{\webpageurl}{https://www.speech.kth.se/research/listen-denoise-action/}

\newcommand{\new}[1]{{#1}} 

\newcommand{\setnewcolor}{} 
\newcommand{\setoldcolor}{}

\newcommand{\newer}[1]{{#1}} 

\crefname{table}{Table}{Tables}
\Crefname{table}{Table}{Tables}
\usepackage{colortbl} 
\let\oldmarginpar\marginpar
\renewcommand\marginpar[1]{\-\oldmarginpar[\raggedleft\footnotesize #1]%
{\raggedright\footnotesize #1}}

\newcommand{\tablebf}[1]{%
\pdfliteral direct {2 Tr 0.5 w}
#1%
\pdfliteral direct {0 Tr 0 w}%
}

\setcopyright{rightsretained}
\copyrightyear{2023}
\acmYear{2023}
\acmDOI{10.1145/3592458}

\acmJournal{TOG}
\acmVolume{42}
\acmNumber{4}
\acmMonth{8}
\acmPrice{15.00}

\acmSubmissionID{papers\_879s1} 

\begin{document}



\title[Listen, Denoise, Action! Audio-Driven Motion Synthesis with Diffusion Models]{Listen, Denoise, Action!\\{}Audio-Driven Motion Synthesis with Diffusion Models}

\author{Simon Alexanderson}
\orcid{0000-0002-7801-7617}
\email{simonal@kth.se}
\affiliation{%
  \institution{Division of Speech, Music and Hearing, KTH Royal~Institute of Technology}
  \city{Stockholm}
  \country{Sweden}}
\affiliation{%
  \institution{\mbox{Motorica} AB}
  \country{Sweden}}
\author{Rajmund Nagy}
\orcid{0000-0002-9653-6699 }
\email{rajmundn@kth.se}
\affiliation{%
  \institution{Division of Speech, Music and Hearing, KTH Royal~Institute of Technology}
  \city{Stockholm}
  \country{Sweden}}
\author{Jonas Beskow}
\orcid{0000-0003-1399-6604}
\email{beskow@kth.se}
\affiliation{%
  \institution{Division of Speech, Music and Hearing, KTH Royal~Institute of Technology}
  \city{Stockholm}
  \country{Sweden}}
\author{Gustav Eje Henter}
\orcid{0000-0002-1643-1054}
\email{ghe@kth.se}
\affiliation{%
  \institution{Division of Speech, Music and Hearing, KTH Royal~Institute of Technology}
  \city{Stockholm}
  \country{Sweden}}
\affiliation{%
  \institution{Motorica AB}
  \country{Sweden}}


\begin{abstract}
Diffusion models have experienced a surge of interest as highly expressive yet efficiently trainable probabilistic models. We show that these models are an excellent fit for synthesising human motion that co-occurs with audio, e.g., dancing and co-speech gesticulation, since motion is complex and highly ambiguous given audio, calling for a probabilistic description. Specifically, we adapt the DiffWave architecture to model 3D pose sequences, putting Conformers in place of dilated convolutions for improved modelling power. We also demonstrate control over motion style, using classifier-free guidance to adjust the strength of the stylistic expression. Experiments on gesture and dance generation confirm that the proposed method achieves top-of-the-line motion quality, with distinctive styles whose expression can be made more or less pronounced. We also synthesise path-driven locomotion using the same model architecture. Finally, we generalise the guidance procedure to obtain product-of-expert ensembles of diffusion models and demonstrate how these may be used for, e.g., style interpolation, a contribution we believe is of independent interest.
\end{abstract}




\begin{teaserfigure}
\centering
\includegraphics[width=\textwidth]{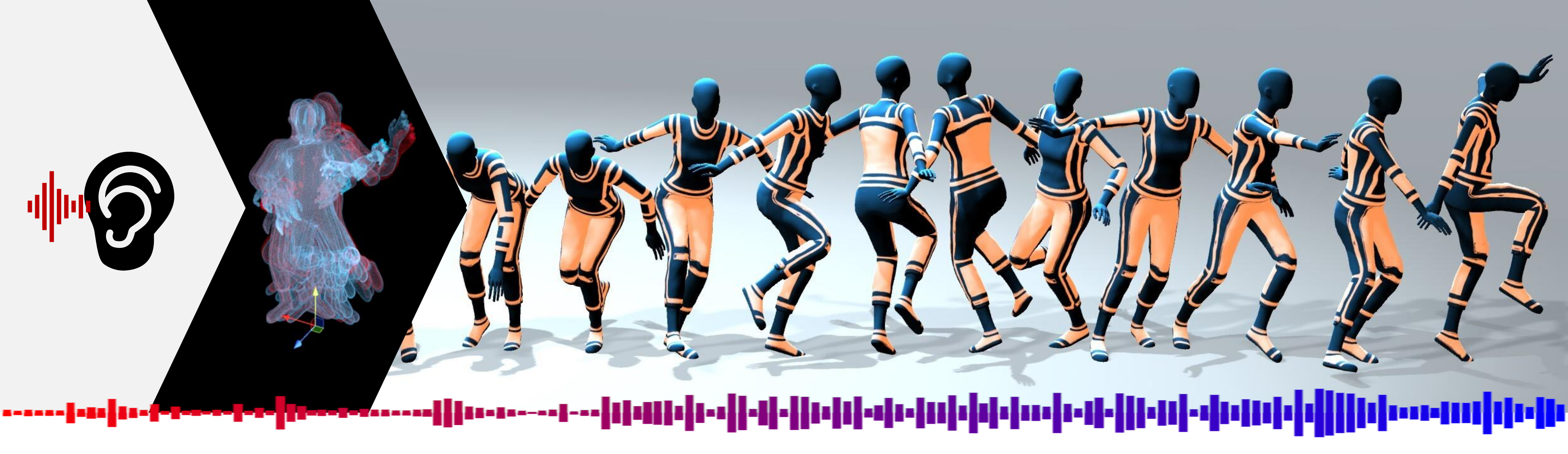}
\vspace{-1.5\baselineskip}
\caption{Listen, denoise, action! Audio-driven Jazz dance motion synthesised from our proposed model. The 3D avatar is \copyright{} Motorica AB.}
\Description{Teaser image illustrating the process of listening to audio and then removing noise from motion, followed by a number of poses of (generated) dancing motion.}
\label{fig:teaser}
\end{teaserfigure}
\maketitle 

\section{Introduction}
\label{sec:intro}
Automated generation of human motion holds great promise for a wide array of applications such as films and special effects, computer games, crowd simulation, architecture and urban planning, and virtual agents and social robots.
Typically, motion occurs in context of other modalities such as audio and vision, and moving appropriately requires taking contextual information into account.
Two motion-generation problems where audio information plays an important role for human behaviour are dancing and co-speech gestures.
\emph{Co-speech gesticulation} -- that is, hand, arm, and body motion that co-occur with speaking -- is an integral part of embodied human communication, and can enhance both human-computer interaction (e.g., avatars and social robots) and human-human digital communication (e.g., in VR and telepresence).
Dancing is a social and deeply human activity that transcends cultural barriers, with some of the most watched content on platforms like YouTube and TikTok specifically involving dance.

However, generating audio-driven motion has proved to be a difficult problem.
Central to the challenge is the fact that such motion is not well predicted by the associated audio:
gestures are highly individual,
nondeterministic, and generally not well-determined by the speech.
The same is true for dancing, which typically synchronises with music structure such as beats and measures, but otherwise can take a vast array of forms even for a single piece of music and genre of performance.
Machine-learning has struggled to cope with this ambiguity and great variability, which can only be accurately captured by a very strong probabilistic model.
In the absence of convincing and controllable motion synthesis models, applications remain reliant on manual labour in the form of expensive motion capture or even more expensive hand animation.


Fortunately, the recent emergence of diffusion models \cite{sohl2015deep,song2019generative,ho2020denoising} offers a general and principled way to learn arbitrary probability distributions using the entire arsenal of deep learning architectures, without issues such as network restrictions (as required for invertibility in normalising flows) or difficult minimax optimisation problems (as required for GANs).
In this paper, we demonstrate the advantages of diffusion probabilistic models for generating high-quality, audio-driven 3D human motion.
Our concrete contributions are:
\begin{itemize}
\item We pioneer diffusion models for audio-driven human motion generation, specifically gestures and dance, using Conformers (\cref{sec:method}).
\item We demonstrate style control with the proposed approach, using classifier-free guidance to adjust the strength of the stylistic expression (\cref{sec:experiments}).
\item We make available a new dataset with audio and high-quality 3D motion capture from diverse genres of dance (\cref{ssec:dance-experiments}).
\item We describe and demonstrate how to leverage product-of-experts ensembles of diffusion models for tasks such as interpolation (\cref{sec:interpolation}).
\end{itemize}
Experiments confirm that the proposed approach outperforms leading baseline systems on multiple datasets in gesture generation and dance, and furthermore is capable of distinctive and adjustable expression of different motion styles.
For code, data, and pre-trained systems, please see \href{https://www.speech.kth.se/research/listen-denoise-action/}{speech.kth.se/research/listen-denoise-action/}.

\section{Background and prior work}
\label{sec:background}
We now review data-driven gesture generation and
dance synthesis, as well as the use of diffusion models for the same.
For a review of human motion generation with deep learning more broadly, please see \citet{ye2021human}.

\subsection{Gesture generation}
Automatic gesture generation makes for more lifelike and engaging artificial agents \cite{salem2012generation}.
It can also aid learning \cite{novack2015learning} and can communicate social information such as personality \cite{neff2010evaluating,durupinar2016perform,smith2017understanding}, and emotion \cite{normoyle2013effect,fourati2016perception,castillo2019we}.
Early work in gesture generation focussed on rule-based approaches \cite{cassell2001beat,kopp2003max,lee2006nonverbal,lhommet2015cerebella} that typically would play pre-recorded gesture clips (or ``lexemes''), at timings selected by hand-crafted rules; see \citet{wagner2014gesture} for a review.
Alternatively, machine learning can be used to learn when to trigger gestures \cite{kucherenko2022multimodal}, even if the gestures themselves still are rendered using pre-determined clips, e.g., \citet{levine2010gesture,chiu2015predicting,sadoughi2019speech,zhou2022gesturemaster}, or via motion matching \cite{habibie2022motion} (where clips only consist of a single frame each \cite{clavet2016motion}).
However, designing a rule-based system requires much manual labour and expert knowledge.
Clip-based models are furthermore fundamentally limited in that they may struggle to synthesise previously unseen motion.
Many of these systems are driven by text rather than audio.

The rise of deep learning has brought increased attention to the problem of audio-driven 3D gesture generation, as a more scalable and generalisable approach to gesture-system creation.
Several relatively early deep-learning systems used recurrent neural networks \cite{takeuchi2017speech,hasegawa2018evaluation,ferstl2018trinity} or convolutional approaches \cite{kucherenko2019analyzing,kucherenko2021moving}.
These were generally based on 3D joint positions in Cartesian coordinates, whereas the field nowadays tends to favour pose representations in terms of joint rotations, since the latter can more easily drive skinned and textured characters in 3D graphics.
They were also limited by treating gesture generation as a regression problem, with one single output, typically leading to underarticluated, averaged output and/or other artefacts.

For a more general approach, research has looked to probabilistic models.
These approaches hold substantial promise, since they can describe an entire range of motion from which distinct realisations can be sampled and furthermore have the potential to generalise much better beyond the available data.
Example approaches have leveraged hidden semi-Markov models \cite{bozkurt2020affective}, \new{combinations of adversarial learning and regression losses \cite{ferstl2019multi,habibie2021learning,liu2022beat}}, normalising flows \cite{alexanderson2020style}, VAEs \cite{ghorbani2022zeroeggs}, VQ-VAEs \cite{yazdian2022gesture2vec}, combinations of flows and VAEs \cite{taylor2021speech}, and different GAN techniques \cite{wu2021modeling,wu2021probabilistic}.
For more in-depth reviews see \citet{nyatsanga2023comprehensive,liu2021speech}.


\subsection{Gesture style control}
Embodied human communication is not merely about what and when we gesture, but also how we do it.
Adding control over style such as identity and mood/emotion to synthetic gestures is thus an important tool for enhanced communication.
Perhaps the dominant approach for style control today, used by, e.g., \citet{yoon2020speech,ahuja2020style,fares2022zero,ghorbani2022exemplar,ghorbani2022zeroeggs}, involves learning an encoder that maps a motion clip to a space of different gesturing styles, often a latent space in a VAE framework.
This setup can be used for one-shot style transfer/adaptation, by feeding examples of novel gestures styles into the encoder to obtain a latent representation of their style.
The method in \citet{alexanderson2020style} differs in its style control, by instead proposing to define styles in terms of continuous-valued kinematic properties such as average hand height, in order to be able to generate different kinds of gesticulation even from a dataset where differences in style had neither been annotated nor deliberately elicited.

This paper uses a different type of style control from all of the above works, which both allows controlling style intensity independently of style identity, and unlocks a new and probabilistically principled way to combine and interpolate between styles.

\subsection{Data-driven dance generation}
Dance generation is perhaps a less explored topic than gesture generation, and will be reviewed more briefly.
The field has seen a similar trajectory as gesture generation in terms of modelling approaches, trying new machine-learning methods as they become available.
Perhaps reflecting the complexity of the probability distribution of dance motion, many approaches generate output by stitching together discrete units, such as concatenating motion segments \cite{fukayama2015music,chen2021choreomaster}, dance figures \cite{ofli2011learn2dance}, or the choreographic action units in ChoreoNet \cite{ye2020choreonet}.
Motion matching has also been used \cite{fan2011example}.
Deep-learning techniques used for directly generating dance pose sequences include recurrent neural networks and autoencoders \cite{crnkovic2016generative,tang2018dance,papillon2022pirounet}, GANs \cite{hu2022multi}, combinations of GANs and VAEs \cite{lee2019dancing}, dilated convolutions and gated activation units \cite{zhuang2022music2dance}, Transformer architectures \cite{li2020learning,valle2021transflower,li2021ai,li2022danceformer}, VQ-VAEs and Transformers with reinforcement learning \cite{siyao2022bailando}, and Conformers \cite{zhang2022music}.

The most straightforward application of style control in deep-learning dance synthesis is to use discrete labels representing different dance genres as an additional model input.
GANs and Transformers have also been used for dance style transfer \cite{yin2022dance}, where style is defined by a motion example rather than a simple label.
Oftentimes, published systems use a large set of audio features together with their style control, which might impact the ability to generalise across different-sounding music pieces and genres, in the sense that it risks creating models that chose their style of dance based on the music rather than the style label supplied by the user.
A goal of our approach is to be able to generate any style of dance to any music.

The prior publications most similar to our work might be  
\citet{zhang2022music} since it integrates Conformers \cite{gulati2020conformer} in the architecture, but with adversarial learning in a non-probabilistic setting, or \citet{valle2021transflower}, which leverages Transformers and probabilistic deep generative modelling, albeit in the form of normalising flows.

\subsection{Diffusion models for motion and audio}
Diffusion models \cite{sohl2015deep,song2019generative,ho2020denoising} are a new paradigm in deep generative modelling that is setting new standards in terms of perceptual quality scores \cite{dhariwal2021diffusion} and also demonstrating very competitive log-likelihood numbers \cite{kingma2021variational}.
Central to this success is that diffusion models combine two very powerful properties: the ability to describe highly general probability distributions (only limited by the expressivity of arbitrary deep-learning architectures) with the ability to efficiently learn these distributions from data by minimising a simple squared-error loss on a carefully designed denoising task.

\new{Diffusion models have recently produced blockbuster results in text-conditioned generation of images \cite{rombach2022high,saharia2022photorealistic,ramesh2022hierarchical} and video \cite{ho2022video,ho2022imagen, hoppe2022diffusion,voleti2022mcvd}.
Whilst
text-conditioned 3D motion generation has been demonstrated without diffusion models \cite{ghosh2021synthesis,petrovich2022temos,guo2022tm2t}, diffusion models have quickly been adopted for that task \cite{zhang2022motiondiffuse,kim2022flame,tevet2023human}.}
These models have been trained on short clips paired with written descriptions of the motion performed and use Transformer architectures \cite{vaswani2017attention} under the hood.
Speech audio is not considered as a model input, nor are text transcriptions of speech.
Style control is possible by specifying desired gesture properties as part of the text prompt, assuming these properties can be adequately captured and articulated in words.

In a parallel development, diffusion models have been applied to generate audio waveforms from acoustic information
(a.k.a.\ neural vocoding)
\cite{kong2021diffwave,chen2021wavegrad}.
\new{However, prior models with} audio-derived input (i.e., acoustic features) \new{do not} generate motion as the output.
This is the contribution of this paper, with speech-driven gesture generation and music-driven dance synthesis as example applications.

\new{We note that, in the time span between
our initial arXiv preprint}
and the camera-ready version of our paper, several concurrent works on audio-driven motion synthesis have been made public.
\newer{These consider either gesture generation \cite{zhang2023diffmotion,ao2023gesturediffuclip} or dance synthesis \cite{dabral2023mofusion,tseng2023edge,ma2022pretrained}, but never both.
It has not been possible to include these works in our empirical comparison, but videos of their motion are available with their respective papers for informal comparison to our results.}
\section{Method}
\label{sec:method}
The task in this paper is to generate a sequence of human poses $\x_{1:T}$ given a sequence of audio features $\bm{a}_{1:T}$ for the same time instances, and optionally a style vector $\bm{s}$.
This section presents the mathematical properties of diffusion models, the new diffusion-model architecture we demonstrate for audio-based motion generation, and how we include style into the models.
Our proposal for style interpolation using products of experts is presented separately in \cref{sec:interpolation}.
In the exposition, bold type signifies vectors and non-bold type scalars.
Limits of summation are written in upper case (e.g., $T$), with lower case denoting indexing operations and colons delimiting ranges of indexing for sequences.

\subsection{Diffusion models}
\label{ssec:diffusion}
Let
$\x$ be distributed according to an unknown density $q(\x)$.
To construct a diffusion model of $\x$, we first define a \emph{diffusion process}, a Markov chain $q\given{\x_n}{\x_{n-1}}$ for $n\in\{1,\,\ldots,\,N\}$ that progressively adds noise to an observation $\x_0$ ($n=0$), eventually erasing all traces of the original observation, so that $q\given{\x_N}{\x_0}$ has a standard normal distribution.
The idea is then to train a network to reverse the $q$-process and ``undo'' the diffusion steps, creating observations out of noise.
This produces the so-called \emph{reverse} or \emph{denoising process}, $p$.

We assume that the noise added by each step of the diffusion process $q$ is zero-mean Gaussian, so that
\begin{align}
q\given{\x_n}{\x_{n-1}}
& = \mathcal{N}(\x_{n};\,\alpha_n \x_{n-1},\,\beta_n\bm{I})
\end{align}
for some $\{\alpha_n,\,\beta_n\}_{n=1}^N$, where $\mathcal{N}$ denotes a multivariate Gaussian density function evaluated at $\x_{n}$.
In this paper we also set $\alpha_n=\sqrt{1-\beta_n}$ \cite{sohl2015deep,ho2020denoising}, in which case $\{\beta_n\}_{n=1}^N$ completely defines the diffusion process.

If the noise added in step $n$ is small relative to $\x_n$, the reverse distribution is also Gaussian \cite{sohl2015deep}.
A Gaussian approximation 
\begin{align}
p(\x_N)
& = \mathcal{N}(\x_N;\,\bm{0},\,\bm{I})\\
p \given{\x_{n-1}}{\x_n}
& = \mathcal{N}(\x_{n-1};\,\bm{\mu}(\x_n,\,n),\,\bm{\Sigma}(\x_n,\,n))
\text{,}
\end{align}
where $\mathcal{N}$ denotes a multivariate Gaussian density function evaluated at $\x_{n-1}$, is therefore likely to be accurate.
In practice, good results have been achieved by setting $\bm{\Sigma}$ equal to a scaled identity matrix \cite{ho2020denoising}.
The learnt distribution is then completely specified by the mean $\bm{\mu}(\x_n,\,n)$, which is predicted by a neural network.
\begin{figure*}
  \centering
  \subcaptionbox{Denoising network\label{sfig:denoiser}}{%
      \includegraphics[width=0.9\columnwidth]{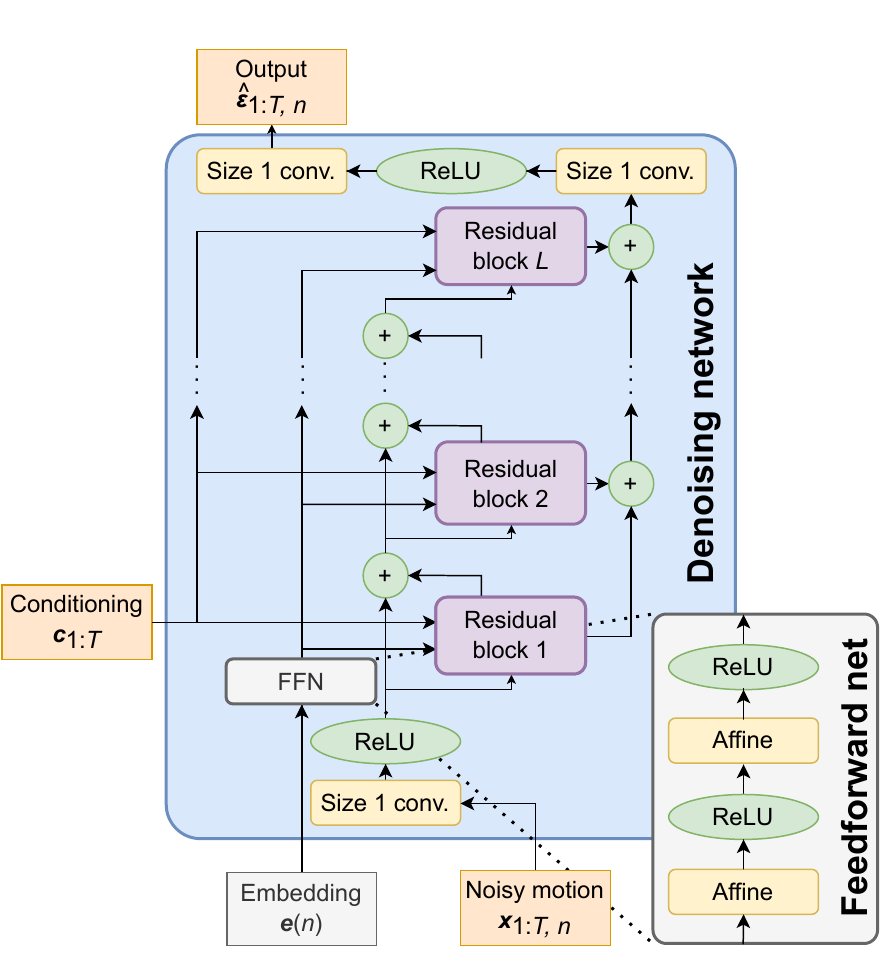}}%
  \hfill
  \subcaptionbox{Residual block and Conformer\label{sfig:conformer}}{%
      \includegraphics[width=1.111\columnwidth]{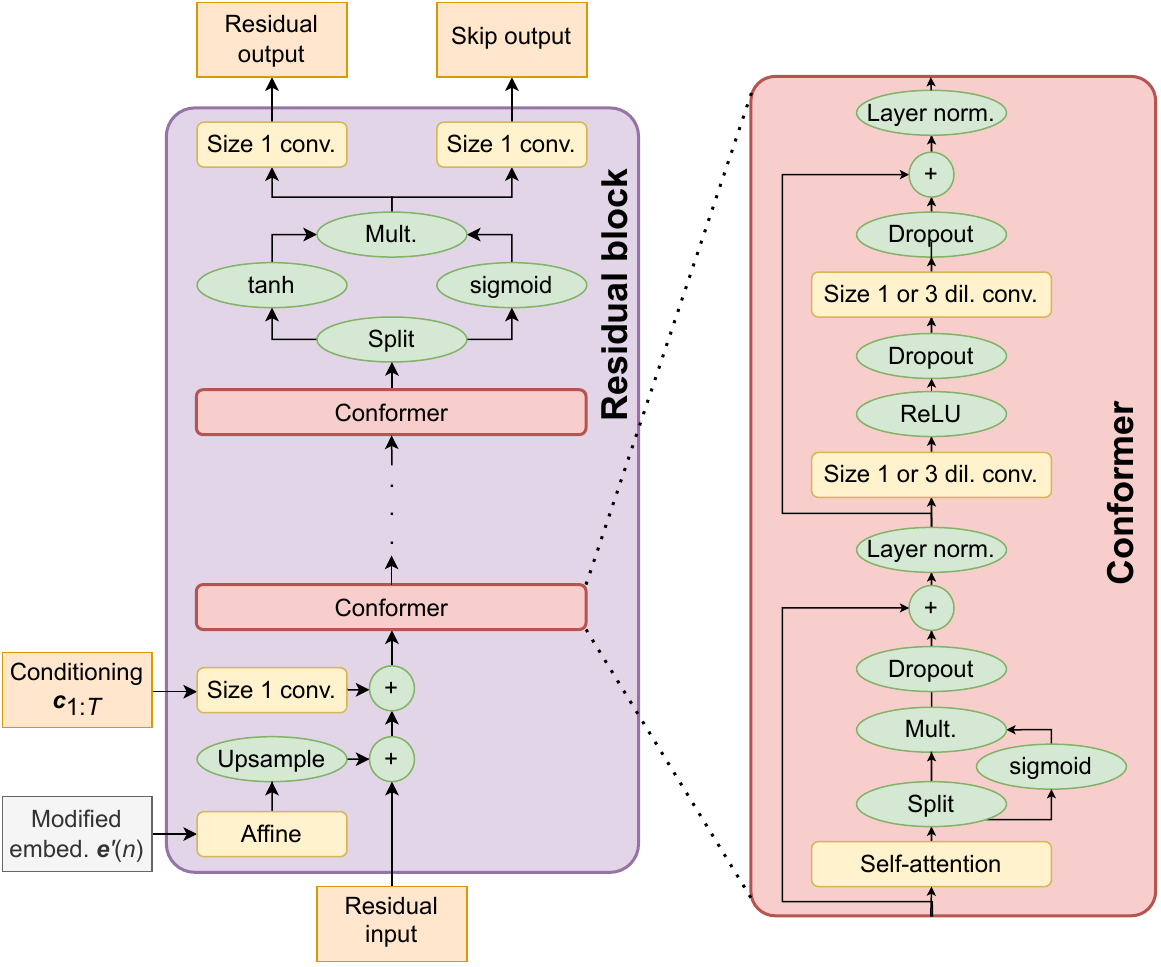}}%
\caption{Architecture diagrams. Each subfigure illustrates a component of the prior subfigure. Rectangular boxes are vectors and scalars, rounded boxes are neural networks or learnt operations, and ovals are fixed mathematical operations.}
\Description{Three architecture diagrams, one for the denoising network, then for the residual block, and finally, the conformer module of the proposed system. The denoising network contains a series of residual blocks, each of them taking the conditioning array and a learned representation of the noisy motion as their input, alongside with a time-step embedding. The residual blocks each contain a stack of Conformer layers, followed by a gating operation. Finally, the conformer layers are shown to contain a self-attention layer, then a sequence of dilated convolutions, with layer norm at the very end.}
\label{fig:architecture}
\vspace{-1\baselineskip}
\end{figure*}

Diffusion models are usually trained through so-called score-matching \cite{hyvarinen2005estimation}, which (similar to energy-based models) does not require knowing the normalisation constant of a distribution.
With a diffusion model, score-matching leads \cite{ho2020denoising} to minimising a loss of the form
\begin{align}
\mathcal{L}\given{\theta}{\mathcal{D}}
& = \mathbb{E}_{\x_0,\,n,\,\bm{\varepsilon}}[\kappa_n\Vert\bm{\varepsilon}-\widehat{\bm{\varepsilon}}(\widetilde{\alpha}_n\x_0+\widetilde{\beta}_n\bm{\varepsilon},\,n)\Vert_2^2]
\label{eq:loss}
\text{,}
\end{align}
in addition to a minor loss term based on the negative log likelihood $-\ln p\given{\x_0}{\x_1}$.
In Eq.\ \eqref{eq:loss}, $\x_0$ is uniformly drawn from the training data $\mathcal{D}$, $n$ is uniform on $\{1,\,\ldots,\,N\}$, $\bm{\varepsilon}\sim\mathcal{N}(\bm{0},\,\bm{I})$, and -- critically -- $\widehat{\bm{\varepsilon}}(\x,\,n)$ is a neural network that predicts the noise $\bm{\varepsilon}$ that was added to $\x_0$.
This is the neural network that defines the denoising process, and thus also the learnt probability density $p(\x_0)$.
$\widetilde{\alpha}_n$ and $\widetilde{\beta}_n$ are constants that depend on $\{\beta_n\}_{n=1}^N$, while $\kappa_n$ are weights. However, a differently-weighted version of the loss, which sets $\kappa_n=1$ as proposed by \cite{ho2020denoising}, is widely used in practice, since it tends to achieve better subjective results \cite{kingma2021variational}.
For a conditional probability model conditioned on some variable $\bm{c}$, one simply trains a network $\widehat{\bm{\varepsilon}}(\x,\,\bm{c},\,n)$ with $\bm{c}$ as an additional input.

It is interesting to note that the training objective in Eq.\ \eqref{eq:loss} contains a simple squared-error minimisation.
Under normal circumstances, the squared-error loss is minimised by the (conditional) expected value, which would regress towards the mean pose and give unnatural motion.
However, the presence of the random variable $\bm{\varepsilon}$ distinguishes the setup from classic minimum mean-squared error, and one can show that score-matching as in Eq.\ \eqref{eq:loss} in fact corresponds to maximising a variational lower bound on the data log-likelihood \cite{kingma2021variational}.
This means that diffusion models actually learn an entire probability distribution.
The variational bound gets tighter as the number of diffusion steps $N$
grows large, obtaining a stochastic differential equation (SDE)  in the limit $N\to\infty$ \cite{song2021score}.

Sampling from a diffusion model starts from $\x_N\sim\mathcal{N}(\bm{0},\,\bm{I})$ and works backwards through the $N$ steps of the reverse process $p$, which may be time-consuming.
Accelerated sampling from trained diffusion models is currently a focus of intense research, e.g., \citet{nichol2021improved,dhariwal2021diffusion,salimans2022progressive,lam2022bddm,meng2022distillation}.
As our aim in this paper is to advance the state of the art in audio-driven motion generation,
exploring faster synthesis is left as future work.

\subsection{Model architecture}
\label{ssec:diffwave}
To generate motion conditioned on audio information, we build on the DiffWave architecture \cite{kong2021diffwave} from conditional audio synthesis.
This model takes acoustic feature vectors as input (usually sampled at 80 Hz) and uses a conditional diffusion model to generate a scalar-valued audio waveform at 22.5 kHz as output.
The model generates all output in parallel, without autoregression or recurrent connections.
Internally, it uses a residual stack of dilated convolutions with skip connections, similar to the trendsetting WaveNet model \cite{oord2016wavenet} for audio synthesis, except that the model (unlike WaveNet) is parallel rather than autoregressive and therefore can use non-causal convolutions.

Denote the sequence of input acoustic feature vectors by $\bm{a}_{1:T}$, where $T$ is the number of frames.
We adapt DiffWave to generate output at the same frame rate as the input (i.e., remove the upsampling), and change to vector-valued rather than scalar output.
In other words, we learn a distribution of the form $p\given{\x_{1:T}}{\bm{a}_{1:T}}$, where $\x_{1:T}=\x_{1:T,\,0}$ is the output of a diffusion process ($0$ indicates the last denoising step) and $\x_t$ is a representation of the pose at time $t$.
In the experiments of this paper, the acoustic features $\bm{a}$ are standard audio features such as mel-frequency cepstrum coefficients (MFCCs) \cite{davis1980comparison},
while the output poses are skeletal joint rotations parameterised using an exponential map representation \cite{grassia1998practical} relative to a T-pose like in \citet{alexanderson2020style}, but this setup is not dictated by our model and other design choices are likely to work similarly well.

The dilated convolutions and skip connections in the original WaveNet together integrate information across many time scales \cite{hua2018wavenets} and allow for an exceptionally wide receptive field that can reach several thousand samples \cite{oord2016wavenet}.
In DiffWave, dilated convolutions (one per residual block $l$) have kernel size 3 and cycle between different dilations $d(l)=2^{l\,\mathrm{mod}\,\bar{l}}$ with $\bar{l}=9$.
Each block also uses a gating mechanism when integrating the conditioning information $\bm{c}_{1:T}$ (here the acoustics $\bm{a}_{1:T}$) on the output.
This can be considered a generalisation of FiLM conditioning \cite{perez2018film}.

Another highly successful mechanism for effectively integrating information over long time scales is multi-head neural self-attention (e.g., \citet{cheng2016long}), especially as used in Transformer architectures \cite{vaswani2017attention}.
That said, convolutional networks (CNNs) still retain an advantage over Transformers in computing kinematically important properties such as finite differences between time frames to represent motion speed and acceleration (force).
Fortunately, the two approaches can be combined by replacing the feedforward networks (convolutions with kernel size 1) inside Transformers with CNNs with kernel sizes larger than one.
This architecture is known as a \emph{Conformer} and outperforms both Transformers and CNNs on tasks like speech recognition \cite{gulati2020conformer}.
To harness these mechanisms, we replace the dilated convolution in the residual blocks of DiffWave with a stack of Transformers or Conformers.
In our experiments, we stack 4 of these in each residual block, all using the same dilation $d(l)=2^{(l\,\mathrm{mod}\,\bar{l})-1}$ with $\bar{l}=3$, where $d(l)<1$ here means that the residual block uses a Transformer (i.e., CNN kernel size 1) rather than a Conformer (kernel size 3), and compare to a similar DiffWave architecture with conventional dilated convolutions and no Transformer/Conformer stack (henceforth just ``Conformers'').

Conformers have no inherent concept of time $t$, and require a position-encoding mechanism to use time information in the computations.
Since we expect motion to be invariant to translation in time, we decided to use a translation-invariant scheme \cite{wennberg2021case,raffel2020exploring}, specifically a version of translation-invariant self-attention (TISA) \cite{wennberg2021case}, to parameterise the impact of sequence position $t$ on the self-attention activations.
This ensures that invariance to temporal translation does not need to be learnt.

Our final architecture is shown in \cref{fig:architecture}, with the specific Conformer architecture shown in \cref{sfig:conformer}.
Like DiffWave, we use sinusoidal embeddings \cite{vaswani2017attention} passed through a feedforward net to encode the diffusion step input $\bm{e}(n)$ to the denoising network.
The Conformers in our experiments use
ReLU nonlinearities and a gating operation.
While \citet{kong2021diffwave} decided to use an L1 loss to train the denoising model in the original DiffWave paper, we only consider the L2 loss, since that is consistent with the theory in Eq.\ \eqref{eq:loss} and seemed to give better results.

Beyond the fact that other recent diffusion models for motion \cite{zhang2022motiondiffuse,kim2022flame,tevet2023human} condition on text rather than audio\new{, and focus on generating motion corresponding to simple actions rather than gestures or dancing that align with the timing and rhythm of an audio signal}, our proposal also differs in the architecture of the denoising network.
Specifically, the cited works all use a stack of Transformers (with \citet{zhang2022motiondiffuse} adding cross-modal attention), but none use Conformers nor dilations.
Furthermore, none of the models use translation-invariant schemes for
incorporating
positional information, with \citet{kim2022flame,tevet2023human} both using sinusoidal position encodings that are known to generalise poorly to sequences longer than those used during training \cite{press2022train}.

\subsection{Style control with guided diffusion}
\label{ssec:control}

For many applications, it is not only important to obtain motion that matches the context -- here the audio that co-occurs with the motion -- but also to have control over the expression of the motion, e.g., generating motion in different styles.
Conditional diffusion models offer a compelling mechanism for controlling not only which style to express (by conditioning $\widehat{\bm{\varepsilon}}$ on style), but independently also the strength of stylistic expression.
The latter is accomplished by a technique called \emph{guided diffusion} \cite{dhariwal2021diffusion}, which corresponds to tuning the temperature $\gamma>0$ of part of the denoising process distribution as
\begin{align}
p_\gamma\given{\x_{n-1}}{\x_n,\,c}
& \propto p\given{\x_{n-1}}{\x_n} p\given{c}{\x_n}^\gamma
\text{,}
\end{align}
so as to focus synthesis on the most distinctive examples of the given class $c$ \cite{dhariwal2021diffusion,dieleman2022guidance}.

Interestingly, the above effect can be achieved by combining the predictions of a conditional and an unconditional diffusion model, which is called \emph{classifier-free guidance} \cite{ho2021classifier}.
This has been used to great effect in models such as GLIDE \cite{nichol2021glide}, DALL$\cdot$E 2 \cite{ramesh2022hierarchical}, and Imagen \cite{saharia2022photorealistic}.
If we let $\bm{s}_{1:T}$ be a vector representing the style input throughout the pose sequence, and define the style-added conditioning $\bm{c}_{1:T}$ with $\bm{c}_t=[\bm{a}^\intercal_t,\,\bm{s}^\intercal_t]^\intercal$,
then classifier-free guidance for stylistic expression can be achieved by combining the prediction of a style-conditional model $\widehat{\bm{\varepsilon}}(\x_{1:T},\,\bm{c}_{1:T},\,n)$ and that of a style-unconditional model $\widehat{\bm{\varepsilon}}(\x_{1:T},\,\bm{a}_{1:T},\,n)$ during the reverse diffusion process as
\begin{multline}
\widehat{\bm{\varepsilon}}_\gamma(\x_{1:T},\,\bm{c}_{1:T},\,n)
= \widehat{\bm{\varepsilon}}(\x_{1:T},\,\bm{a}_{1:T},\,n)\\
+ \gamma(\widehat{\bm{\varepsilon}}(\x_{1:T},\,\bm{c}_{1:T},\,n) - \widehat{\bm{\varepsilon}}(\x_{1:T},\,\bm{a}_{1:T},\,n))
\label{eq:guidance}
\text{.}
\end{multline}
This de-emphasises the effect of the style input $\bm{s}$ for $0\leq\gamma<1$ and exaggerates the effect for $\gamma>1$.
$\gamma=0$ recovers the unconditional model.
Our experiments study style control both with and without guidance, where $\bm{s}$ is a one-hot vector that encodes discrete style labels from the data, which do not change with the time $t$ in a sequence.
It is possible to train one single network to describe both the style-conditional and style-unconditional models, by randomly dropping out the style information $\bm{s}_{1:T}$ from some sequences during training, in order to represent the unconditional case,
but this is not necessary to apply the method.
In our experiments, we instead used separate style-conditional and style-unconditional models, since that led to better-looking results in preliminary tests.

\section{Experiments}
\label{sec:experiments}
We conducted a number of experiments to demonstrate the capabilities of our proposed approach.
After describing the data processing and modelling approach (in \cref{ssec:data}) and the general evaluation framework (in \cref{ssec:evaluation}),
we begin by comparing our proposal to the best available alternatives on two gesture-generation datasets
(\cref{ssec:gesture-generation-experiments}).
We thereafter describe an additional user study that compared our approach to the state of the art in music-driven dance synthesis \newer{(\cref{ssec:dance-experiments})}.
\new{Objective metrics are reported in \cref{ssec:objective} whilst \cref{ssec:locomotion-experiments}
shows} that the approach generalises to
path-driven locomotion generation. 
\cref{ssec:experiments-summary} summarises the findings.
Please see \href{https://www.speech.kth.se/research/listen-denoise-action/}{speech.kth.se/research/listen-denoise-action/} for video clips of generated motion.
Demonstrations of our proposal to create product-of-expert diffusion models are reserved for \cref{sec:interpolation}.

\subsection{Data and modelling}
\label{ssec:data}
Our experiments used five different datasets from high-quality 3D motion capture.
An overview of these datasets is provided in \Cref{tab:datasets}, with additional information provided in the section where each dataset is used.

\begin{table}
\centering
\caption{Overview of datasets and hyperparameters.
Numbers after a slash are for style-unconditional models.}
\begin{tabular}{@{}llllll@{}}
\toprule 
Dataset & TSG & Zero- & \multicolumn{2}{c}{Motorica} & 100STYLE\tabularnewline
 &  & EGGS & Dance & MMA & \tabularnewline
\midrule
Motion & Gestures & Gestures & Dancing & Fighting & Locomotion\tabularnewline
Duration & 244 min & 135 min & 373 min & 60 min & 1125 min\tabularnewline
Performers & 1 male & 1 female & 3 F, 2 M & 1 male & 1\tabularnewline
No.\ styles & \hphantom{00}1 & \hphantom{0}19 & \hphantom{00}8 & \hphantom{00}1 & 100\tabularnewline
\midrule
$\bm{c}_{t}$ dim. & \hphantom{0}20 & \hphantom{0}35/16 & \hphantom{0}11/3 & \hphantom{00}8 & 103\tabularnewline
~~~Audio & \hphantom{0}20 & \hphantom{0}16/16 & \hphantom{00}3/3 & \hphantom{00}0 & \hphantom{00}0\tabularnewline
~~~Other & \hphantom{00}0 & \hphantom{0}19/\hphantom{0}0 & \hphantom{00}8/0 & \hphantom{00}8 & 103\tabularnewline
$\bm{x}_{t}$ dim. & \hphantom{0}70 & \hphantom{0}70 & \hphantom{0}61 & \hphantom{0}61 & \hphantom{0}58\tabularnewline
~~~\new{Pose} & \new{\hphantom{0}67} & \new{\hphantom{0}67} & \new{\hphantom{0}58} & \new{\hphantom{0}58} & \new{\hphantom{0}58}\tabularnewline
~~~\new{Root offset} & \new{\hphantom{00}3} & \new{\hphantom{00}3} & \new{\hphantom{00}3} & \new{\hphantom{00}3} & \new{\hphantom{00}0}\tabularnewline
Steps $N$ & 100 & 100 & 150 & 150 & 150\tabularnewline
No.\ updates & 112k & 67k & 180k & 52k & 292k\tabularnewline
\bottomrule
\end{tabular}
\label{tab:datasets}
\vspace{-1\baselineskip}
\end{table}

All experiments considered full-body motion only.
This is a more challenging problem than only generating upper-body motion (cf.\ \citet{yoon2022genea}), both due to the increased dimensionality of the output space and due to the visual prominence of artefacts such as foot-sliding and ground penetration if highly nonlinear constrains due to foot-ground interactions are not accurately modelled.

All datasets were initially 60 frames per second or more, but were converted to 30 fps for the modelling.
We used skeletal joint rotations to represent poses, and parameterised these rotations using an exponential map representation \cite{grassia1998practical} relative to a T-pose.
All models except for path-driven locomotion generated three additional outputs (namely instantaneous rotation and forwards and lateral translation, similar to \citet{holden2016deep,habibie2017recurrent}) that describe the motion of the root node along the ground plane, to enable the model to move the character around.
Because root-node rotation is parameterised using change of heading relative to the previous frame it is not subject to phase-wrapping discontinuities, simplifying the learning task.
For path driven locomotion, the root-node motion was used as an input to the synthesis, rather than an output, and was then smoothed across 10 frames to emulate the smooth paths typically used in animation systems.

Audio conditioning inputs differed between models and were dependent on domain (speech vs.\ music).
We used one hot-encodings for all style inputs, since we only considered discrete style labels.
The fighting model in \cref{ssec:interpolationdemo} took binary flags for a set of fighting moves as input.
See \Cref{tab:datasets} for an overview of the input and output feature dimensionalities of different models.
For most systems with style control, we also trained one identical model but without the style input, to enable adjusting the stylistic expression via classifier-free guidance as described in \cref{ssec:control}.
Values for these style-unconditional systems are shown on the right side of the slashes (`/') in the table.
Additional details on the data processing are provided in the appendix.

Initial hyperparameter tuning was performed on the dance dataset in \cref{ssec:dance-experiments} using the \href{https://optuna.org/}{Optuna} framework.
We found that the most important aspect to tune was the noise scheduling (the $\beta_n$-values), where we could achieve significantly improved motion by increasing the end-value of the linear schedule. This leads to more time spent using high-noise data in the diffusion process.
The proposed models were trained on a single GPU with 80 5-second motion sequences per batch.
The number of training updates for each dataset is specified in \Cref{tab:datasets}.
Additional details on model training and hyperparameters are provided in the appendix.

\definecolor{lightgray}{rgb}{0.83, 0.83, 0.83}

\subsection{
Evaluation methodology}
\label{ssec:evaluation}

The gold standard in evaluating motion naturalness, style expression, etc.\ is to perform subjective evaluation (i.e., user studies), and careful subjective studies are hence the core of the evaluations in this paper.
The reason for this is that these fields (unlike core machine learning) lack common benchmark datasets and have no \new{widely validated} objective metrics.
\new{Although we compute and report objective metrics in \cref{ssec:objective}, the core of our evaluation is our user studies, since these directly measure how gestures appear to humans.}


\new{For the task of gesture generation}, the most successful instances of common benchmarking are the GENEA challenges \cite{kucherenko2021large,yoon2022genea}, which compared a large number of contemporary gesture-generation methods under a common setup in several best-practices user studies that are the largest in the field as of yet.
These challenges have been held twice thus far, once (in 2020) using data from the Trinity speech-gesture (TSG) dataset \cite{ferstl2018trinity}, and once (in 2022) using data from Talking With Hands 16.2M \cite{lee2019talking}.
We compared our proposed method to the deep generative model with the highest output quality in each instance of the challenge, as described in \cref{sssec:stylegestures-baseline} and \cref{sssec:zeggs-baseline}.

\begin{figure}
    \centering
    \includegraphics[width=\columnwidth]{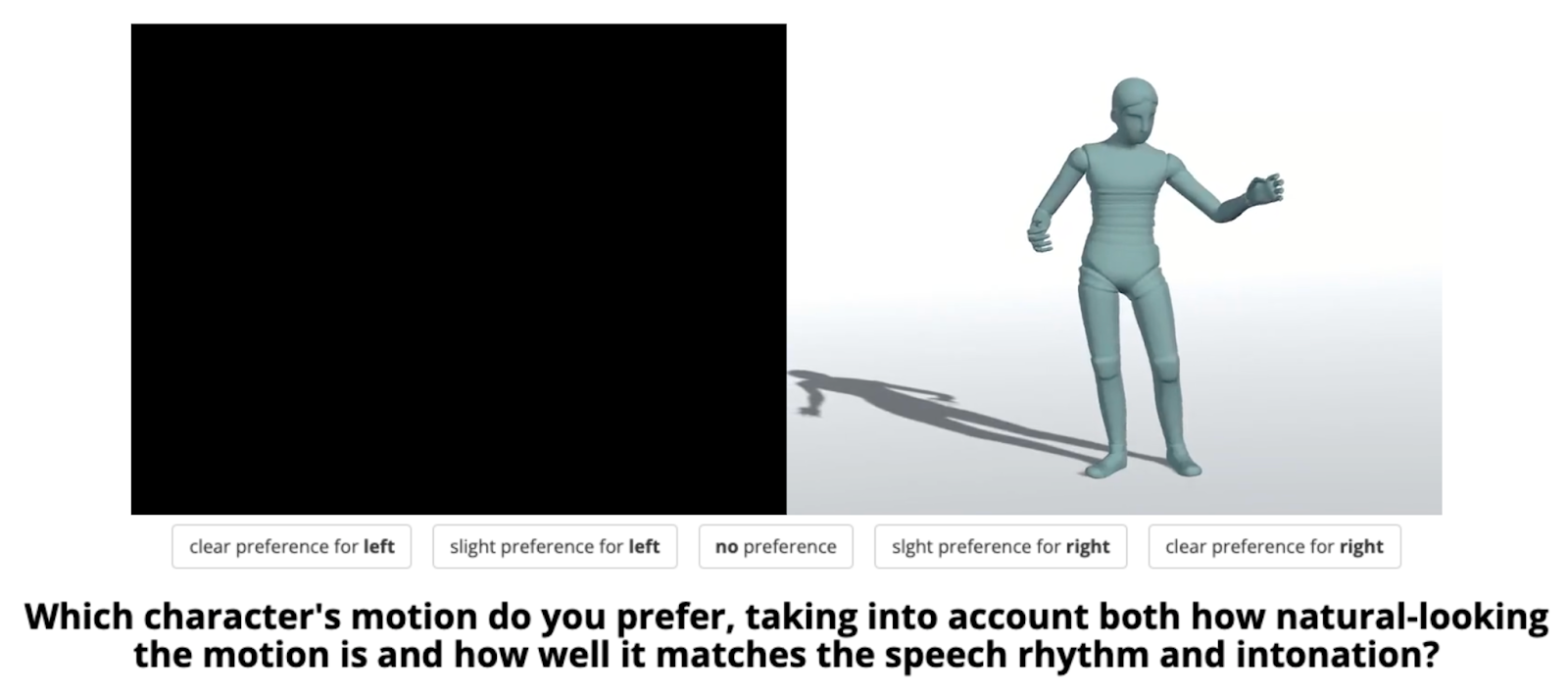}
    \caption{Screenshot of the user interface used for subjective evaluations.}
    \Description{An interface with two videos side by side. The left video appears to be fully black in this example while the right video shows a gesticulating 3D avatar standing on the ground, seen from the front. There are five buttons below the videos, indicating "clear preference for left"; "slight preference for left"; "no preference"; "slight preference for right"; and "clear preference for left". Below the buttons, there's a text asking "Which character's motion do you prefer, taking into account both how natural-looking the motion is and how well it matches the speech rhythm and intonation?".)}
    \label{fig:evaluation_interface_screenshot}
    \vspace{-1\baselineskip}
\end{figure}



Concretely, we performed a total of four user studies to assess both motion quality and the distinctiveness of stylistic expression.
Three of the studies considered gesture generation and one considered dance, but all shared the same setup and analysis method.
Each study was based on pairwise comparisons, where two 10-second video clips were played consecutively side by side.
A number of 20-second long comparison videos were assembled, first showing an animation on the left side, with a black frame on the right, followed by an animation on the right and a black frame on the left (see \cref{fig:evaluation_interface_screenshot}).
The audio driving the animation was always the same for the left and the right clip, and all clips within a study used the same avatar, but each of the two videos in a comparison was from a different system (condition).
The main experimental screen consisted of a video with a text question and five response buttons underneath, where the full video had to be played for the response buttons to become active.
Participants for the user studies were recruited using the Prolific crowdsourcing platform%
\footnote{\href{https://www.prolific.co/}{https://www.prolific.co/}}
and filtered using attention checks prior to analysis.
Please see the appendix for more details, e.g., regarding the attention checks.

Our main method of analysis for each user study relied on merit scores \cite{parizet2005comparison}.
In this methodology, responses with a slight preference for a system were encoded numerically as a score of 1, and clear preferences were assigned a 2, whereas ties or a preference for the other system in the comparison gave a score of 0.
The data was then analysed by means of a one-way ANOVA and a post-hoc Tukey multiple-comparisons test for statistical significance.
In addition to reporting the merit scores with 95\% confidence intervals for each condition, we also report the win rate (excluding ties) of our main proposed system in every study.

\subsection{Gesture-generation experiments}
\label{ssec:gesture-generation-experiments}
\subsubsection{Experiment on the Trinity speech-gesture dataset}
\label{sssec:stylegestures-baseline}
In the first experiment, we compared ourselves to \emph{StyleGestures}%
\footnote{\href{https://github.com/simonalexanderson/StyleGestures}{https://github.com/simonalexanderson/StyleGestures}}
\cite{alexanderson2020stylegestures}, an autoregressive normalising-flow model for speech-conditioned gesture synthesis that achieved the best motion quality in the GENEA Challenge 2020 \cite{alexanderson2020style, kucherenko2021large}.
For this evaluation, we used the TSG dataset (also used as the basis for the GENEA Challenge 2020), which contains around four hours of full-body motion capture and speech audio from a male actor delivering spontaneous monologues.
We used the same train/test split as the 2020 GENEA Challenge and evaluated only on data from the test set.
Motion was visualised using the GENEA 2022 avatar seen in \cref{fig:evaluation_interface_screenshot}.
The avatar design lacks mouth and gaze, to instead draw attention to the rest of the body.

The TSG data did not deliberately elicit different styles and does not contain any style labels.
Although StyleGestures described the optional capability to control the motion style, this was demonstrated using kinematically derived style correlates such as hand height, hand speed, and gesticulation radius \cite{alexanderson2020stylegestures}.
As StyleGestures has not been validated on the discrete style control
we consider
in this paper, we therefore only compare to StyleGestures in terms of motion quality, using only
speech audio conditioning for all models in the experiment.
To represent the speech acoustics, we used 20-dimensional mel-frequency cepstrum coefficients (MFCCs) \cite{davis1980comparison} for each frame.
\begin{table}[!t]
\centering
\caption{Overview of the design of the different user studies. ``No.\ ordered pairs'' is the number of possible pairwise comparisons of distinct conditions, taking presentation order into account. ``Ordered pairs seen'' refers to how many times each participant was exposed to each ordered condition pair.}
\begin{tabular}{@{}lcccc@{}}
\toprule 
Dataset & TSG & \multicolumn{2}{c}{ZeroEGGS} & Dance\tabularnewline
Evaluation & Pref. & Pref. & Style & Pref.\tabularnewline
\midrule
No.\ conditions & \hphantom{00}4 & \hphantom{00}3 & \hphantom{00}3 & \hphantom{00}4\tabularnewline
No.\ ordered pairs & \hphantom{0}12 & \hphantom{00}6 & \hphantom{00}6 & \hphantom{0}12\tabularnewline
Styles evaluated & \hphantom{00}1 & \hphantom{00}4 & \hphantom{00}4 & \hphantom{00}6\tabularnewline
Videos per condition & \hphantom{0}36 & \hphantom{0}36 & \hphantom{0}36 & \hphantom{0}36\tabularnewline
~~~\ldots{}and style & \hphantom{0}36 & \hphantom{00}9 & \hphantom{00}9 & \hphantom{00}6\tabularnewline
Total video stimuli & 432 & 216 & 216 & 432 \tabularnewline
Audio included? & \cmark & \cmark & \xmark & \cmark\tabularnewline
\midrule
Participants recruited & 131 & \hphantom{0}40 & \hphantom{0}40 & \hphantom{0}40\tabularnewline
Participants included & 129 & \hphantom{0}38 & \hphantom{0}38 & \hphantom{0}40\tabularnewline
Ordered pairs seen & \hphantom{00}3 & \hphantom{00}6 & \hphantom{00}6 & \hphantom{00}3\tabularnewline
\bottomrule
\end{tabular}

\label{tab:setup}
\vspace{-1\baselineskip}
\end{table}
\begin{table*}[!t]
\centering
\caption{Results from user studies. Studies include a proposed system (LDA), a dataset-dependent variant thereof, a baseline, and (where indicated) ground truth (GT). We report merit scores \cite{parizet2005comparison} with 95\% confidence intervals and the win rate excluding ties of
\new{every condition versus LDA.}}
\begin{tabular}{@{}ll|cc|cccc|cc@{}}
    \toprule 
    \multicolumn{2}{l|}{Dataset} & \multicolumn{2}{c|}{TSG} & \multicolumn{4}{c|}{ZeroEGGS} & \multicolumn{2}{c}{Dance}\tabularnewline
    \multicolumn{2}{l|}{Evaluation} & \multicolumn{2}{c|}{Preference test} & \multicolumn{2}{c}{Preference test} & \multicolumn{2}{c|}{Style control} & \multicolumn{2}{c}{Preference test}\tabularnewline
    \multicolumn{2}{l|}{Measure} & Merit score & Win rate & Merit score & Win rate & Merit score & Win rate & Merit score & Win rate\tabularnewline
    \midrule
     & GT & 0.95$\pm$0.04 & \new{67.3}\% &\cellcolor{lightgray} -   &\cellcolor{lightgray} -   &\cellcolor{lightgray} -  &\cellcolor{lightgray} -  & 0.87$\pm$0.06 & \new{61.1}\% \tabularnewline
     & LDA & \tablebf{0.59$\pm$0.03} &\cellcolor{lightgray} -  & \tablebf{0.64$\pm$0.08} &\cellcolor{lightgray} -  & 0.36$\pm$0.04  &\cellcolor{lightgray} -  & \tablebf{0.71$\pm$0.06}  &\cellcolor{lightgray} - \tabularnewline
    \midrule
    \multirow{3}{*}{\begin{turn}{90}
    Variants
    \end{turn}} & LDA-DW & 0.50$\pm$0.03 & \new{42.5}\% &\cellcolor{lightgray} -  &\cellcolor{lightgray} -  &\cellcolor{lightgray} -  &\cellcolor{lightgray} -  &\cellcolor{lightgray} -  &\cellcolor{lightgray} - \tabularnewline
     & LDA-G &\cellcolor{lightgray} -  &\cellcolor{lightgray} -  &0.41$\pm$0.07 & \new{37.2}\% &\tablebf{0.61$\pm$0.05} & \new{62.2}\%  &\cellcolor{lightgray} -  &\cellcolor{lightgray} - \tabularnewline
     & LDA-U &\cellcolor{lightgray} -  &\cellcolor{lightgray} -  &\cellcolor{lightgray} -  &\cellcolor{lightgray} -  &\cellcolor{lightgray} -  &\cellcolor{lightgray} -  & 0.51$\pm$0.05 & \new{36.7}\% \tabularnewline
    \midrule
    \multirow{3}{*}{\begin{turn}{90}
    Baselin.
    \end{turn}} & SG & 0.31$\pm$0.03 & \new{28.0}\% &\cellcolor{lightgray} -  &\cellcolor{lightgray} -  &\cellcolor{lightgray} -  &\cellcolor{lightgray} -  &\cellcolor{lightgray} -  &\cellcolor{lightgray} - \tabularnewline
     & ZE &\cellcolor{lightgray} -  &\cellcolor{lightgray} -  & 0.43$\pm$0.07 & \new{34.7}\% &0.50$\pm$0.04  &\new{56.5}\%  &\cellcolor{lightgray} -  &\cellcolor{lightgray} - \tabularnewline
     & BL &\cellcolor{lightgray} -  &\cellcolor{lightgray} -  &\cellcolor{lightgray} -  &\cellcolor{lightgray} -  &\cellcolor{lightgray} -  &\cellcolor{lightgray} -  & 0.24$\pm$0.06 & \new{23.5}\% \tabularnewline
    \bottomrule
    \end{tabular}
\label{tab:userstudy}
\vspace{-1\baselineskip}
\end{table*}

For our experiment on the TSG data we performed a preference test to evaluate the quality of the gesture motion.
Four conditions were compared: ground-truth motion capture (labelled GT), the proposed system (LDA), an ablation that used the original DiffWave architecture instead of our proposed Conformers (LDA-DW), and a StyleGestures system without style input (SG).
36 audio segments of 10 s from the TSG test set were used to generate animation with each of the three models (LDA, LDA-DW, SG) plus ground truth (GT).
%
These animations were then evaluated in a user study, as described in \cref{ssec:evaluation}.
For each comparison video in the study, participants were asked ``\textit{Which character's motion do you prefer, taking into account both how natural-looking the motion is and how well it matches the speech rhythm and intonation?}''.
Answers were given using a 5-point Likert-style preference response scale, 
with the response alternatives ``clear preference for left'', ``slight preference for left'', ``no preference'', ``slight preference for right'', and ``clear preference for right''; see \cref{fig:evaluation_interface_screenshot}.
Additional information about the user study is provided in \Cref{tab:setup}.


\Cref{tab:userstudy} summarises the results of the user study on the TSG dataset.
The ground-truth motion capture (GT) achieved the best merit score, followed by LDA, LDA-DW, and SG, in that order.
Our statistical analysis found all differences to be significant, with $p<0.01$ for LDA vs. LDA-DW and $p<0.001$ for all other differences.
Whilst not reaching the same level as human motion capture, our proposed diffusion model thus outperformed StyleGestures.
We also found that the proposed architecture with Conformers contributed to this.
The difference between LDA and LDA-DW was furthermore larger in pilot studies we performed using different hyperparameters, including a different noise schedule, suggesting that the Conformers also are less sensitive to hyperparameter settings.
Since LDA-DW performed less well, it is not considered in subsequent studies.


\subsubsection{Experiments on the ZeroEGGS dataset}
\label{sssec:zeggs-baseline}
In the second and third experiments, we compared ourselves to the best-performing deep generative submission to the 2022 GENEA Challenge \cite{yoon2022genea}, namely a model \cite{ghorbani2022exemplar} later released as \emph{ZeroEGGS} \cite{ghorbani2022zeroeggs}.
ZeroEGGS is an autoregressive model consisting of a feedforward speech encoder, a probabilistic style encoder (using a Transformer) with a standard Gaussian prior, and a recurrent gesture-generation module (using gated recurrent units, GRUs).

ZeroEGGS was released alongside a new gesture dataset of the same name \cite{ghorbani2022zeroeggs} with two hours of speech and full-body motion capture of a female actor delivering monologues in 19 diverse styles.
This data allows us to demonstrate control over a range of style expressions.
(The GENEA 2022 data has no such styles.)
We selected four styles for our evaluation, spanning emotional states (\emph{happy} and \emph{angry}), speaking styles (\emph{public speaking} a.k.a.\ \emph{oration}), and age (\emph{old}).
\new{This experiment used 16-dimensional MFCCs as input, instead of 20 dimensions as used for TSG.
This is because the Madmom audio-processing library%
\footnote{\href{https://github.com/CPJKU/madmom}{https://github.com/CPJKU/madmom}}
used for feature extraction defaults to a different number of MFCCs depending on the sampling rate of the audio, which differs between the two datasets.}


To prevent the style of the speech audio from interfering with the effect of style control in our evaluations (either by biasing the generated styles or raters' perceptions of these styles), we held out two speech recordings in the \emph{neutral} style (specifically \texttt{004\_Neutral\_3} and \texttt{005\_Neutral\_4}), so as to have sufficient neutral speech material for our evaluation, and used the remaining data for training.

We trained and compared three conditions on this data: our proposed model (labelled LDA), the same model using classifier-free guidance of the style control during synthesis (LDA-G), and Zero\-EGGS (ZE).
The latter used the original model hyperparameters and code from the official codebase and data repository.%
\footnote{\href{https://github.com/ubisoft/ubisoft-laforge-ZeroEGGS}{https://github.com/ubisoft/ubisoft-laforge-ZeroEGGS}}
For LDA-G, we used a guidance factor $\gamma=1.5$, to study the impact of exaggerated stylistic expression obtained through classifier-free guidance.

Nine 10-second audio segments from the neutral-style test set of ZeroEGGS were used to generate animation in four different styles (\emph{happy}, \emph{angry}, \emph{old}, and \emph{public speaking}), yielding 36 segment-style combinations.
Motion was visualised using the female avatar that was released together with the ZeroEGGS dataset.


\paragraph{Motion preference evaluation}
We conducted the same type of preference test as in the first experiment, with the same question and response alternatives, but for the three systems trained on the ZeroEGGS data.
Additional information about the user study is provided in \Cref{tab:setup}, whilst
\Cref{tab:userstudy} shows the results.
The best performer was LDA, with ZE and LDA-G being statistically tied in terms of merit scores.
In addition to SG earlier, the proposed model thus also outperforms ZeroEGGS in terms of preference $(p<0.001)$.
That difference disappears when using guided diffusion to exaggerate style expression, but it is not surprising to find exaggerated motion less natural and less preferred.

\paragraph{Style-control evaluation}
To measure the distinctiveness of stylistic expression, we conducted a another experiment, where we displayed the same motion clips as in the previous experiment, but instead of asking for user preference we asked ``\textit{Based on the \textbf{body movements alone} (disregarding the face), which of the two clips looks most like {\normalfont{\textbf{\texttt{STYLE}}}}}?'', where \textbf{\texttt{STYLE}} was one of the following: \{``\textbf{the person is happy}'', ``\textbf{the person is angry}'', ``\textbf{an old person}'', ``\textbf{a person giving a speech to the public}''\}, using bold font to highlight key parts of the instructions as shown.
The five response alternatives were ``clearly the left one'', ``probably the left one'', ``I can not tell'', ``probably the right one'', and ``clearly the right one''.
The instruction to disregard the facial expression was added since the ZeroEGGS avatar has a quite stern facial expression, which might contradict, e.g., the \textit{happy} gesturing style.

In this experiment, the videos were silent in order to prevent speech content from affecting the judgement of gesturing style.
(That
one modality can affect the perception of the other has been confirmed in multiple studies, e.g., \citet{jonell2020let,kucherenko2021large,bosker2021beat}.)
Presentation was grouped by style, meaning that all nine \emph{happy}-style comparisons were presented first, followed by the \emph{angry}, \emph{old}, and \emph{public speaking} comparisons in that order.
Before each new style, an interstitial screen was presented, informing the participant that the subsequent stimuli were going to be judged according to the style in question.
The order of comparisons within a style was randomised for each participant.

An overview of the style-control user study is provided in \Cref{tab:setup}, with the results shown in \Cref{tab:userstudy}.
LDA-G received the highest style-matching score, ahead of ZE ($p<0.01$), which was ahead of LDA ($p<0.001$).
This confirms that we were able to successfully moderate style strength by using guided diffusion, to an extent that can surpass ZeroEGGS in distinctiveness.
%
\begin{figure}[!t]
\centering
\includegraphics[width=\columnwidth]{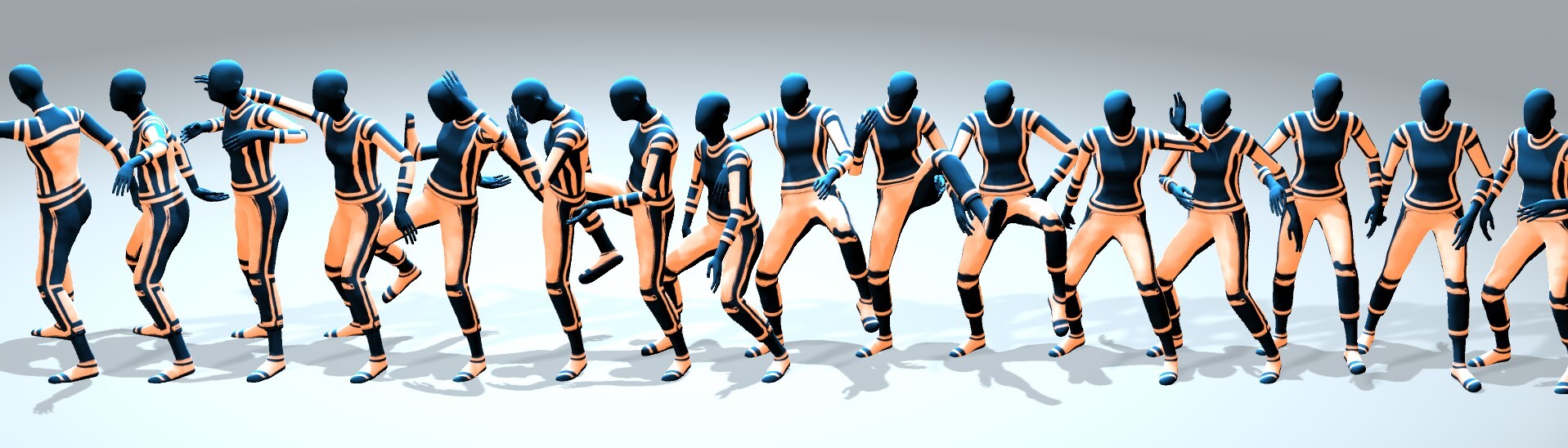}
\includegraphics[width=\columnwidth]{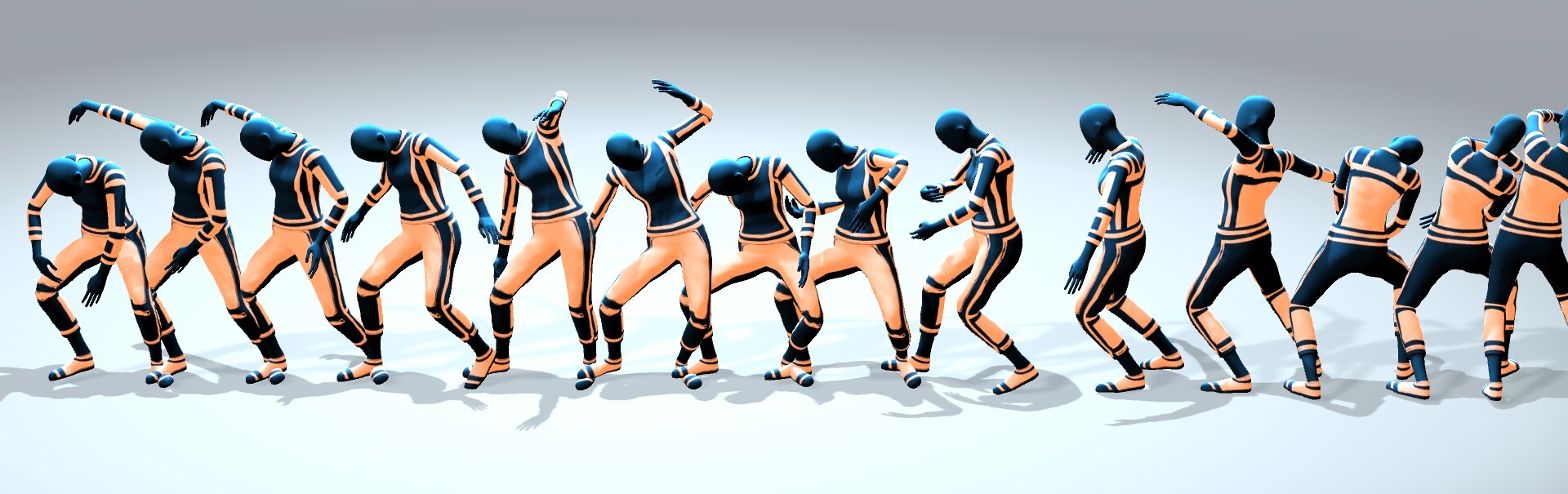}
\caption{Dance synthesised from our trained diffusion model in the \emph{Locking} and \emph{Krumping} styles. See \cref{fig:teaser} for the \emph{Jazz} style. Avatar \copyright{} Motorica AB.}
\Description{Two rows representing smooth and diverse dancing movement. Each row contains frames extracted from a different motion clip; in the top row, the dancer is slightly turning and making an exaggerated stomping dance move, while in the bottom row, the dancer is performing a "water wave" motion with their upper body, turning by 180 degrees near the end.}
\label{fig:dance}
\vspace{-1\baselineskip}
\end{figure}

\subsection{Music-driven dance synthesis}
\label{ssec:dance-experiments}
As a second application of audio-driven motion generation, we also train and evaluate a music-driven dance model based on our approach.
For this, we used a new dataset that combines new and existing high-quality motion-capture data.
Specifically, we extracted a subset of the highest-quality recordings from the PSMD dataset from \citet{valle2021transflower}, using only material recorded using optical motion capture.
The selected material contained the \emph{Casual} style and three street dance styles (\emph{Hip-Hop}, \emph{Popping}, and \emph{Krumping}).
These were combined with an additional dataset of approximately 3.5 hours of high-quality recordings from three accomplished dancers in the genres \emph{Jazz}, \emph{Charleston}, \emph{Tap dancing}, and \emph{Locking}, yielding the 373 minutes of parallel music audio and dance motion-capture described in \Cref{tab:datasets}.
We intend to release this dataset upon paper acceptance.

\new{Compared to the most commonly used recent dance dataset, AIST++ from \citet{li2021ai}, our dataset is larger, captures entire songs, and adds a completely new set of styles.
AIST++ was also not captured using marker-based mocap, leading to data artefacts such as floating and jittery motion, that may be reproduced by (or otherwise degrade) generative models trained on such data.}
This may explain why \citet{tseng2023edge}, a concurrent work on diffusion models for dance generation that was trained on AIST++, incorporated additional loss terms relating to, e.g., foot contacts, during training.
Our models used only vanilla diffusion model loss terms and no foot stabilisation, smoothing, or other post-processing.

We trained both style-conditional and style-unconditional proposed models on this data (labelled LDA and LDA-U, respectively).
These models were conditioned on a minimalist set of three audio features from music information retrieval, specifically spectral flux (1 feature), chroma \cite{muller2005chroma} (1 feature),
and the activation of the RNNDownBeat-processor \cite{bock2016joint} (1 feature), all obtained using the Madmom
audio signal processing library, as before.
This parsimonious feature set was chosen \new{with the} goal of reflecting the structure of the music whilst being sufficiently nonspecific generalisable across music genres, so as to enable dancing in any style to any music.
It also likely reduces the risk of overfitting, and we found that versions of our model that used much larger music-feature sets gave worse results in an informal comparison.
\begin{figure}[!t]
\centering
\includegraphics[width=\columnwidth]{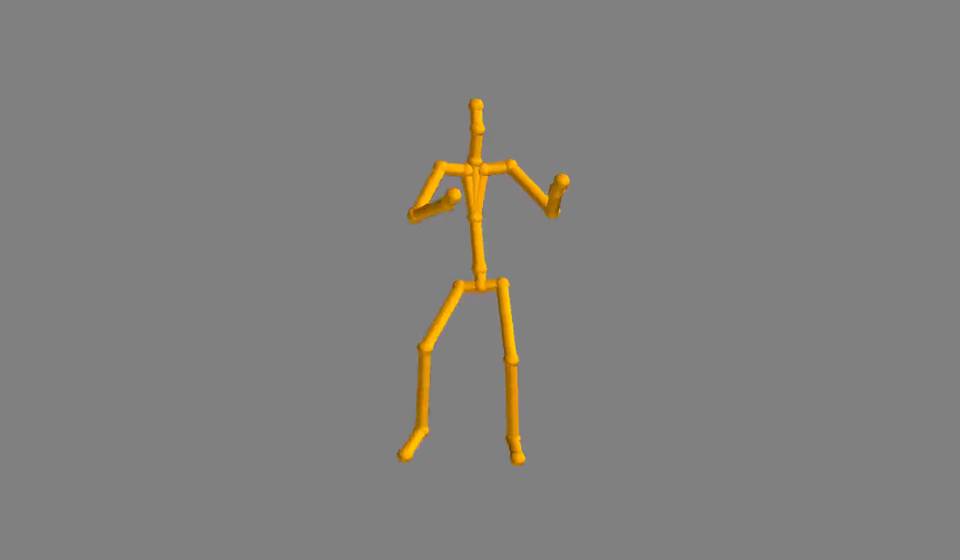}
\caption{3D stick-figure skeleton visualisation excerpted from dance-evaluation video.}
\Description{A 3D stick figure is floating at the center of the figure, with empty grey background.}
\label{fig:skeleton}
\vspace{-1\baselineskip}
\end{figure}

%
We also trained a \emph{Bailando} \cite{siyao2022bailando} model on the same dataset, as a representative of the state of the art in the field.
\citet{siyao2022bailando} report strong results in subjective evaluations, being rated as better than other recent baseline systems at least 80\% of the time, and even beating the ground truth in 40\% of comparisons.

Bailando is an autoregressive probabilistic model with Transformers that operates on a discretised version of the pose sequence using ideas from language modelling.
Compared to the other models in this paper, it has a quite involved training procedure with four distinct steps that have to be performed in order.
We used the official codebase%
\footnote{\href{https://github.com/lisiyao21/Bailando/}{https://github.com/lisiyao21/Bailando}} and the original hyperparameters for training.
The implementation uses 438-dimensional music features, but does not include any explicit style input, so the dance style is instead tightly coupled to what the music sounds like.
In that sense, Bailando is more similar to our style-unconditional model (LDA-U) than our main proposed system (LDA).

We trained Bailando (BL) on 3D Cartesian joint positions following the training procedure of \citeauthor{siyao2022bailando}. This restricts possible visualisations for the user studies, therefore we visualised the dance motion for these using videos with stick-figure animations of the skeleton motion, similar what the Bailando paper did \cite{siyao2022bailando}.
A screenshot of our 3D model is shown in \cref{fig:skeleton}.
The absence of a ground plane and shadow in our visualisation is to level the playing field for Bailando, which often generates ground penetration or floating motion, since it parameterises the root motion in terms of vertical delta values, meaning that the height of the character in absolute coordinates easily may drift over time.
Our proposed approach does not have this issue.

The two proposed model variants (LDA and LDA-U) were trained on joint rotations, as before, to be able to directly apply model output motion to skinned characters. However, for the user-study visualisation, the outputs of the proposed models were converted into 3D joint positions.
An example of a skinned avatar driven by our style-conditional dance-generation model is provided in \cref{fig:dance}.
Even in a still image the displayed styles are visually distinctive.
Dance videos are provided on our project page.
\new{Our demonstrations include long dance sequences of over a thousand frames all generated in one single go.
These show that our models, despite only being trained on sequences 150 frames in length, are able to generalise well to much longer sequence lengths.
This is likely attributable to the use of translation-invariant self-attention in our architecture, cf.\ \citet{wennberg2021case,press2022train}.}


For our subjective evaluation, we compared the trained models (LDA, LDA-U, and BL) to each other and to the ground truth (GT) using held-out motion and songs.
As in \cref{sssec:zeggs-baseline}, we selected a subset of distinctive styles for the subjective comparison, namely \emph{Jazz}, \emph{Charleston}, \emph{Hip-Hop}, \emph{Locking}, \emph{Krumping}, and \emph{Casual}.
We then generated dance motion for 36 10-second audio segments, 6 for each style of dance, using each of the models.
Although each style used different musical accompaniment, there is no guarantee that style-unconditional models like LDA-U or BL will generate dance motion in the style of the music, especially for styles of dance that are performed to similar music.
In contrast, our style-conditioned model can dance to a given piece of music in any style we request, as shown in our presentation video.

We ran a subjective preference test with the overall design and numbers resported in \Cref{tab:setup}.
For each comparison video in the study, participants were asked ``\textit{Which character's dancing motion do you prefer, taking into account both how natural-looking
the motion is and how well it matches the music?}'', with the same response alternatives as in previous preference tests.

\Cref{tab:userstudy} summarises the results the from dance preference evaluation.
The ground-truth motion capture (GT) achieved the best scores, followed by LDA, LDA-U, and BL.
Our statistical analysis found all differences to be significant ($p<0.001$).
We can conclude that our proposed diffusion model can generate dancing motion with a quality that surpasses the state of the art in the field, as represented by Bailando.
Whilst not reaching the same level as human motion capture, LDA still exhibited almost 40\% win rate against GT.
The lower rating of LDA-U compared to LDA may be explained by the relatively small number of input music features, which might not be sufficient to provoke genre-appropriate dancing for LDA-U, whereas LDA has more information about what dancing would be appropriate owing to the style control we supplied.

It is our opinion that dances generated by our Bailando baseline BL actually have a better subjective motion quality than the dance examples presented with the original Bailando paper.
We believe this difference is due to the consistently high quality of the motion capture in the new database we have assembled, suggesting that our data release will be of value to the community.


\setnewcolor
\begin{table}[!t]
\centering
\caption{\new{Objective metrics.
The Fréchet distances (FGD, $\text{FID}_k$, $\text{FID}_g$) are computed with respect to the full dataset.
Low values are better.
GT is a top-line condition based on the recorded motion capture in the test set.
Higher beat-alignment scores (BAS) mean a closer match to the beats.}}
\begin{tabular}{@{}ll|c|c|ccccc@{}}
\toprule 
\multicolumn{2}{l|}{Dataset} & TSG & ZeroEGGS & \multicolumn{5}{c}{Dance}\tabularnewline
\multicolumn{2}{l|}{Evaluation} & Quality & Quality & \multicolumn{2}{c}{Quality} & \multicolumn{2}{c}{Diversity} & Rhythm\tabularnewline
\multicolumn{2}{l|}{Measure} & FGD & FGD & $\text{FID}_{\text{k}}$ & $\text{FID}_{\text{g}}$ & $\text{Div}_{\text{k}}$ & $\text{Div}_{\text{g}}$ & BAS \tabularnewline
\midrule
 & GT & \hphantom{00}9.94 & 31.84 & \hphantom{0}3.00 & \hphantom{0}5.22 & 9.55 & 7.12 & 0.2662\tabularnewline
 & LDA & \hphantom{0}25.77 & 59.07 & \hphantom{0}6.62 & \hphantom{0}7.41 & 9.78 & 5.77 & 0.2559\tabularnewline
\midrule
\multirow{3}{*}{\begin{turn}{90}
Variants
\end{turn}} & LDA-DW & 126.92 & \cellcolor{lightgray}- & \cellcolor{lightgray}- & \cellcolor{lightgray}- & \cellcolor{lightgray}- & \cellcolor{lightgray}- & \cellcolor{lightgray}- \tabularnewline
 & LDA-G & \cellcolor{lightgray}- & 47.68 & \cellcolor{lightgray}- & \cellcolor{lightgray}- & \cellcolor{lightgray}- & \cellcolor{lightgray}- & \cellcolor{lightgray}- \tabularnewline
 & LDA-U & \cellcolor{lightgray}- & \cellcolor{lightgray}- & \hphantom{0}7.69 & \hphantom{0}8.98 & 9.13 & 6.13 & 0.2600 \tabularnewline
\midrule
\multirow{3}{*}{\begin{turn}{90}
Baselin.
\end{turn}} & SG & 118.61 & \cellcolor{lightgray}- & \cellcolor{lightgray}- & \cellcolor{lightgray}- & \cellcolor{lightgray}- & \cellcolor{lightgray}- & \cellcolor{lightgray}- \tabularnewline
 & ZE & \cellcolor{lightgray}- & 52.99 & \cellcolor{lightgray}- & \cellcolor{lightgray}- & \cellcolor{lightgray}- & \cellcolor{lightgray}- & \cellcolor{lightgray}- \tabularnewline
 & BL & \cellcolor{lightgray}- & \cellcolor{lightgray}- & 28.15 & 12.70 & 5.22 & 4.35 & 0.2311\tabularnewline
\bottomrule
\end{tabular}
\label{tab:objective}
\vspace{-1\baselineskip}
\end{table}
\setoldcolor


\subsection{\new{Objective metrics}}
\label{ssec:objective}
\new{
Although our main evaluation directly quantified how human observers rate the different motion-generation approaches, we also computed a selection of objective metrics for the various conditions in our experiments.
The majority of these metrics are based on the Fréchet Inception Distance (FID) \cite{heusel2017gans}, which is the most accepted objective metric for evaluating generative models.
The FID between a dataset $\mathcal{D}$ and a set of synthetic samples $\mathcal{S}$ is defined as the 2-Wasserstein distance between two multivariate Gaussian distributions, whose means and variances are estimated from features extracted from the individual samples in $\mathcal{D}$ and $\mathcal{S}$ using a domain-specific feature extractor.
The FID thus quantifies how much the real and synthetic data distributions differ from each other, providing a proxy for perceived synthesis quality or convincingness.
It cannot, however, capture whether or not synthesised examples accurately reflect the associated control input -- for example whether motion accurately reflects a desired style, or how appropriate the motion is for the rhythm and content of the audio.

For the gesture-synthesis models, we used the pre-trained autoencoder network provided by \citet{yoon2020speech} as the feature extractor.\footnote{\href{https://github.com/ai4r/Gesture-Generation-from-Trimodal-Context}{https://github.com/ai4r/Gesture-Generation-from-Trimodal-Context}}
The resulting metric is known as the Fréchet Gesture Distance (FGD), and \newer{has been found to have} moderate but nonzero correlation with gesture human-likeness ratings \citep{kucherenko2023evaluating}.
For each gesture dataset in our experiments, we divided the test set into evenly spaced, non-overlapping 10-second sequences, leaving gaps for past- or future input context windows that the baseline models required.
(For example, SG requires access to future speech audio during generation, and also needs an initial past context due to its use of autoregression.)
After that we randomly generated three 10-second motion clips with each model for all of these test sequences, and used these to compute the FGD with respect to the full dataset.

For the dance-synthesis models, we also divided the test dataset into non-overlapping 10-second windows, generated motion clips with each model for each window, and computed the same metrics as \citet{siyao2022bailando}: an FID score using kinematic features \cite{onuma2008fmdistance}, denoted $\text{FID}_k$; an FID score using geometric features \cite{Mueller2005Efficientcontentbased}, denoted $\text{FID}_g$; the corresponding diversity scores, $\text{Div}_k$ and $\text{Div}_g$, computed as the average Euclidean distance between the features of each possible pair of motion clips; and a beat-alignment score (``BAS'') \newer{based on the inverse of the average time} between each music beat and its closest dance beat.

\cref{tab:objective} summarises the objective results.
We see that the best FGD and FID scores come from comparing the test-set motion capture to the natural motion capture in the database (row ``GT''), which can be considered a kind of top line.
Although the objective scores mostly align with human perception in our main experiments, there are a few outliers, e.g., LDA-DW has an FGD on par with the SG baseline, even though it significantly outperformed that baseline in the user study (\Cref{tab:userstudy}).
Similar outlying scores were observed in experiments to validate the FGD metric on large amounts of human judgements, as reported in \citet{kucherenko2023evaluating}.
This can possibly be related to the fact that motion-capture datasets are relatively small, with the low amount of test-set data available limiting the statistical accuracy of objective metrics, compared to what we are used to in large-data tasks such as image generation.

From the table, we can further see that our method produced more diverse dancing than the BL baseline, with the kinematic diversity on par with the natural dance.
Our method also exhibited greater beat alignment that the baseline, coming close to that of the recorded professional dance motion, but we caution that good dancing is not a beat-matching task \citep{valle2021transflower} and natural dance and multimodal communication often contain a wide diversity of nested rhythms \citep{miller2013you,pouw2021multilevel}.
}
\begin{figure}[!t]
\centering
\includegraphics[width=\columnwidth]{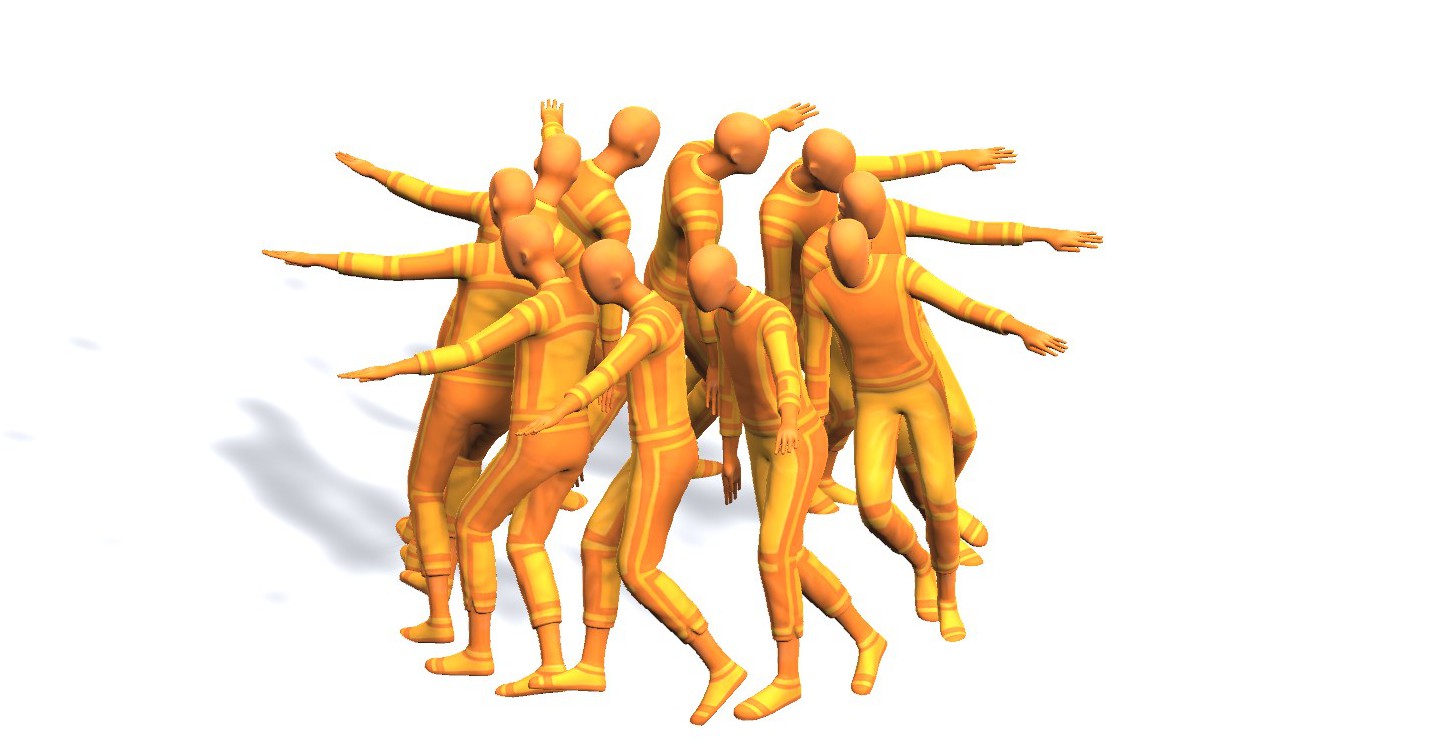}
\caption{Locomotion generated by our model, conditioned on a circular path and a style that always holds the left arm out. Avatar \copyright{} Motorica AB. 
\Description{Around ten frames extracted from a motion clip where the 3D avatar is walking in a circle. The left hand is held out straight away from the body in every frame.}
}
\label{fig:loco}
\vspace{-1\baselineskip}
\end{figure}

\subsection{Path-driven locomotion synthesis}
\label{ssec:locomotion-experiments}
To highlight the generality of our modelling approach, we also trained our diffusion model to perform
path-driven locomotion generation.
In this scenario, the root-node motion was constrained to follow along a path defined on the ground plane, which was parameterised similar to \citet{holden2016deep,habibie2017recurrent}, i.e., root-node rotation and forwards and sideways translation, a total of three numbers per frame.
We trained these models on the recent 100STYLE dataset \cite{mason2022realtime}, which contains high-quality 3D motion capture in 100 diverse styles of locomotion.
In addition to style information, the three numbers per frame that describe the root-node path were provided as conditioning information to the model.

The motion in \cref{fig:loco} was generated from a circular path (constant forward and rotational velocity) in the style \emph{Raised\-Left\-Arm}.
The motion is clearly consistent with both the style and the path control.
Video examples are on the project page.
\new{The videos show that the model is able to generate motion without noticeable foot-skating (no foot stabilisation was applied to any motion in our paper), and that the model also is able to generate transitions between different root-node speeds, even whilst turning.}

Path-driven locomotion was previously generated using diffusion models by \citet{findlay2022denoising}, but only for stick figures.
It is our personal opinion that our motion is noticeable steadier and better quality than what they demonstrated.

\subsection{Summary}
\label{ssec:experiments-summary}
To conclude, our subjective evaluations find that our proposed audio-driven motion-generation models are significantly preferred over a number of strong baseline methods, on three different datasets covering both gesture generation and dance.
We have also shown that classifier-free guidance can successfully moderate the stylistic expression strength of our models.



\section{Products of expert diffusion models}
\label{sec:interpolation}
In this section, we extend the ideas behind classifier-free guidance in \cref{ssec:control} to describe general product-of-expert ensembles from combining the predictions of several diffusion models.
We first describe the theory, then compare the ideas to prior work, and finally present an experimental investigation.

\subsection{Theory}
Mathematically, classifier-free guidance in Eq.\ \eqref{eq:guidance} comprises a barycentric combination of two diffusion-model predictions guiding the process expression to be more or less similar to -- i.e., towards or away from -- one of the models, namely the predictions of an unconditional model.
This suggests a generalisation instead using a barycentric combination of two conditional models
\begin{multline}
\widehat{\bm{\varepsilon}}_{(1-\gamma)\bm{s}_1+\gamma\bm{s}_2}(\x_{1:T},\,\bm{a}_{1:T},\,n)\\
= (1-\gamma) \widehat{\bm{\varepsilon}}(\x_{1:T},\,[\bm{a}^\intercal_{1:T},\,\bm{s}^\intercal_{1\,1:T}]^\intercal,\,n)\\
+ \gamma\widehat{\bm{\varepsilon}}(\x_{1:T},\,[\bm{a}^\intercal_{1:T},\,\bm{s}^\intercal_{2\,1:T}]^\intercal,\,n)
\text{,}
\label{eq:interp}
\end{multline}
to blend (i.e., interpolate) between two different styles $\bm{s}_1$ and $\bm{s}_2$ during the diffusion steps.
We call this setup \emph{guided interpolation}.
We can furthermore extrapolate by choosing $\gamma\notin[0,\,1]$, leading to style expressions that exaggerate one style in a manner that makes it even more distinctive from the other style.

\new{We emphasise that guided interpolation is not the same as inpainting.
Inpainting is a common use case of diffusion models, with \citet{tevet2023human} and \citet{zhang2022motiondiffuse} both demonstrating the use of diffusion models for inpainting in the motion domain (i.e., \emph{motion editing} or \emph{inbetweening}), by filling in missing poses or joint data at different time frames.
Guided interpolation, in contrast, interpolates between different probability distributions described by diffusion models, such as different motion styles.}

The idea behind guided interpolation readily generalises to more than two models.
Consider $M$ diffusion models $\widehat{\bm{\varepsilon}}_m$ with different conditioning information $\bm{c}_{m,\,1:T}$.
These can be combined as 
\begin{align}
\widehat{\bm{\varepsilon}}_{\bm{\gamma}}(\x_{1:T},\,\{\bm{c}_{m,\,1:T}\}_m,\,n)
& = \sum_{m=1}^M \gamma_m \widehat{\bm{\varepsilon}}_m(\x_{1:T},\,\bm{c}_{m,\,1:T},\,n)
\text{,}
\label{eq:ensemble}
\end{align}
where $\sum_{m=1}^M \gamma_m=1$, as required for a barycentric combination.
(We do not require $\gamma_m\in[0,\,1]$, to permit extrapolation and not only interpolation.)
Note that $\widehat{\bm{\varepsilon}}_m$ can be different models, or instances of the same model with different conditioning input $\bm{c}_{m,\,1:T}$ (e.g., different style input, as in Eq.\ \eqref{eq:interp}).
It is not required that $\widehat{\bm{\varepsilon}}_m$ is always the same model nor that they were trained on the same data.
They do not even have to accept the same conditioning information, and instead the dimensionality and content of $\bm{c}_{m,\,1:T}$ can depend on $m$.
All that is needed is that the output spaces match.
This construction is highly general and allows combining an arbitrary number of potentially heterogeneous diffusion models into an ensemble, interpolating between their predictions at will \new{during denoising}.

The score-matching objective used for training diffusion models means that the diffusion models at every step approximate $\nabla_{\x_{n-1}} \ln p\given{\x_{n-1}}{\x_n,\,\bm{c}}$ \cite{song2019generative,dieleman2022guidance}.
By working backwards through the steps in \citet{dieleman2022guidance},
reverting the gradient operation and taking the exponent%
\footnote{This derivation is not completely rigorous, e.g., one needs to assume that the network predictions $\widehat{\bm{\varepsilon}}\given{\x_{n-1}}{\x_n,\,\bm{c},\,n}$ form a divergence-free vector field on $\x_{n-1}$, which happens if score matching is optimal, but need not be exactly true in practice, but the result is nonetheless elucidating.}, we can see that the proposed ensemble corresponds to performing denoising steps based on a product of multiple different denoising distributions,
\begin{multline}
p_{\bm{\gamma}}\given{\x_{1:T,\,n-1}}{\x_{1:T,\,n},\,\{\bm{c}_{m,\,1:T}\}_m}\\
\propto \prod_{m=1}^M p_m\given{\x_{1:T,\,n-1}}{\x_{1:T,\,n},\,\bm{c}_{m,\,1:T}}^{\gamma_m}
\label{eq:prodofexperts}
\text{.}
\end{multline}
This is a product of several probability density functions, i.e., a \emph{product of experts} \cite{hinton2002training}.
The fact that the exponents sum to one avoids the probability-concentration issues seen with na{\"\i}ve product-of-experts models discussed in \citet{shannon2011effect}.

We note that guided interpolation through products of experts is different from conventional mechanisms for \new{creating intermediate conditioning} information in deep-learning models, such as averaging two inputs (which
may lead to previously unseen input and undefined behaviour, especially if style labels are discrete) or averaging two latent-space embeddings (whose average similarly may fall into regions of latent space with poor coverage during training; cf.\ \citet{tomczak2018vae}).
Instead of interpolating between two conditional inputs, our proposal effectively interpolates between the \emph{behaviour} of several conditional diffusion processes, each of which is well defined.
This is less likely to produce unnatural output, since both diffusion processes in isolation are good at driving noisy $\x_n$-values towards regions of outcome space perceived as natural.

Products of experts are a compelling paradigm for ensembling synthesis models.
The combination of multiple experts restricts output away from regions that any single expert considers to be unnatural or otherwise inappropriate, i.e., that have low probability.
\new{(This happens because the experts in Eq.\ \eqref{eq:ensemble} steer the denoising process towards values that have high probability according to each the expert.
Specifically, since the largest updates in the combination come from experts that find $\x_{1:T}$ to be improbable, the process will converge toward a point where all experts make small updates, and thus all agree that the probability of $\x_{1:T}$ is high.)
An inductive bias that favours output that no component model considers unnatural} is appealing for synthesis applications \cite{henter2016minimum,theis2016note}, since it is easier for human observers to notice the presence of something undesirable than to notice the absence of something desirable (e.g., reduced output diversity).
(This same principle also explains why GANs can yield high-quality samples even in the presence of high degrees of mode collapse.)
This inductive bias contrasts against conventional additive mixtures like Gaussian mixture models (GMMs) and mixture density networks (MDNs) more generally \cite{bishop1994mixture}, where each expert adds rather than takes away behaviours/outcomes, which can be problematic if the individual experts do not all maintain high output quality.

Despite their advantages, products of experts have hitherto seen little use for synthesis, as they have been considered difficult to sample from \cite{mackay2003information}, especially in high dimensions.
Diffusion models can sample from unnormalised models and give good results with few steps \cite{nichol2021improved,dhariwal2021diffusion}, removing this stumbling block and making highly general probabilistic product-of-experts models feasible for synthesis applications.

\subsection{Related work on mixtures of experts}
Mixtures of experts in general have been used before in motion generation using deep learning.
Aside from conventional additive mixtures like the GMM mixture density networks in \citet{fragkiadaki2015recurrent}, mixtures of experts have have notably been used in mode-adaptive neural networks and their extensions, e.g., \citet{zhang2018mode,ling2020character,starke2020local,starke2022deep,xie2022learning}.
Those experts correspond to different weight matrices, allowing the network weights to change, e.g., during a walking cycle.
However, these ideas are not a product of experts and have not yet been demonstrated for diffusion models.

An ensemble of diffusion model experts was presented in \citet{balaji2022ediffi}, but the experts have disjunct domains
(they correspond to different steps $n$ in the denoising process),
so unlike a product of experts only one expert is used at any given point (denoising step).
\new{The most relevant prior works we have found
\cite{zhao2022egsde,liu2022compositional} are based on treating diffusion models as energy-based models.
Of these, \citet{zhao2022egsde} combine a single diffusion model with several non-diffusion energy-based models obtained by removing the last layer of two classifiers.
\citet{liu2022compositional}, in contrast, combine multiple diffusion-model predictions at synthesis time, leading to a similar formalism as presented here, although not stated in terms of a product of experts.
All of these works consider image generation rather than motion synthesis.}

\new{Concurrent work by \citet{ma2022pretrained} also considers using multiple diffusion models to create intermediate behaviours, but by randomly choosing the prediction of only a single model at any given denoising step $n$.
They dub this approach \emph{alternating control}.
By making the probability of choosing one of their two different models depend on $n$, the resulting synthesis process can be made to favour one type of motion for the coarse outline of the motion, but with details more closely based on another motion type.}

\begin{figure}[!t]
  \centering
      \includegraphics[width=\columnwidth]{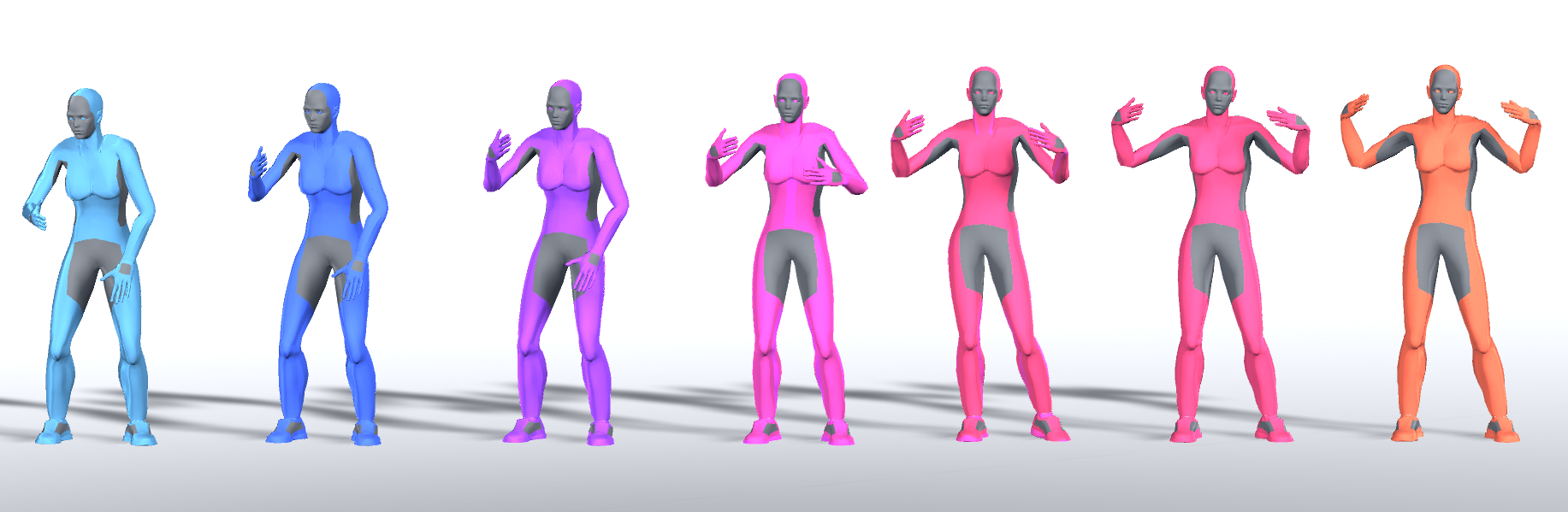}
\caption{Snapshot of \emph{old} (left) to \emph{angry} (right) interpolation for $\gamma=-0.25$ to $\gamma=1.25$. Poses on the far left and right constitute extrapolation. All examples were generated using the same random seed. Avatar \copyright{} Ubisoft.}
\Description{Seven standing 3D avatars are lined up with gaps in between them. The first three avatars, corresponding to the smaller gamma values and the old style, appear to be gesticulating with one hand in a slightly hunched pose, while the other hand is resting on the hip. The remaining avatars appear more and more vivid as gamma is increased, with both hands gesticulating vividly while standing upright.}
\label{fig:interpshape}
\vspace{-1\baselineskip}
\end{figure}
\begin{figure}[!t]
  \centering
      \includegraphics[width=\columnwidth]{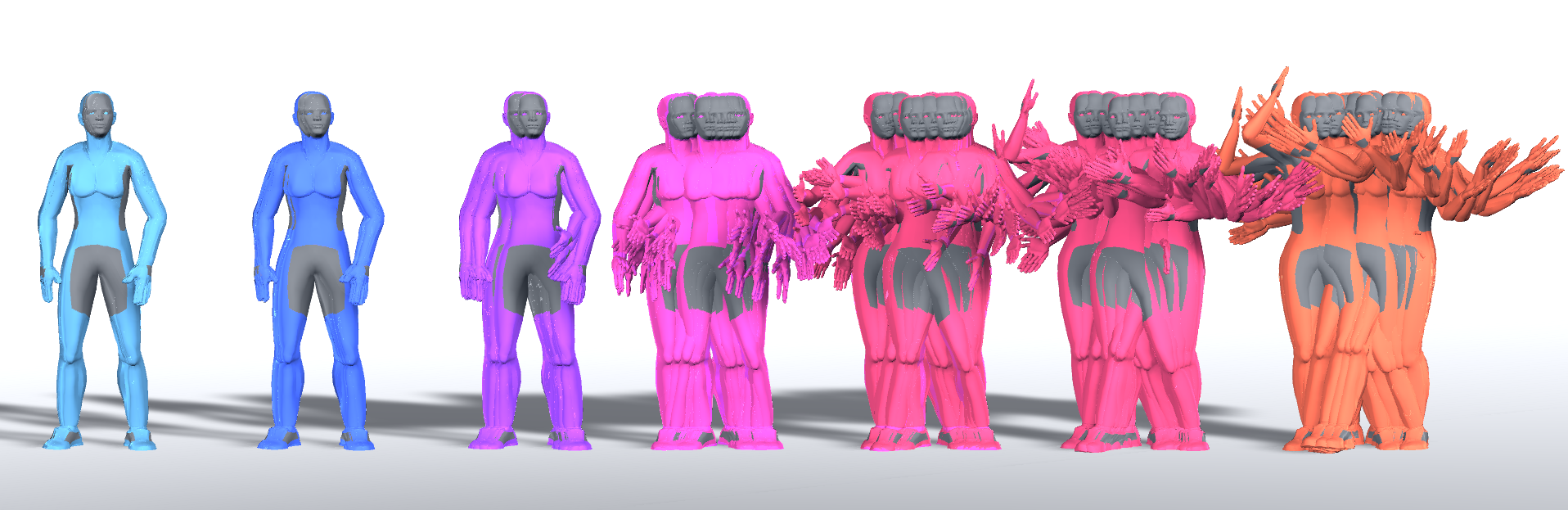}%
\caption{10 s overlay of \emph{still} (left) to \emph{public speaking} (right) interpolation for $\gamma=-0.25$ to $\gamma=1.25$. Extrapolation is seen on the far left and right. All examples were generated using the same random seed. Avatar \copyright{} Ubisoft.}
\Description{Seven standing 3D avatars are lined up with gaps in between them. The motion overlay on them shows that the first three avatars (with smaller gamma values) are almost completely still, while the rest of the avatars gesture more and more frantically as gamma is increased.}
\label{fig:interptemporal}
\vspace{-1\baselineskip}
\end{figure}

\subsection{Experiments}
\label{ssec:interpolationdemo}
\emph{Interpolating between gesture styles.}
To demonstrate the effects of product-of-expert diffusion models for motion, we performed experiments with guided interpolation between different gesturing styles, by mixing predictions from the style-conditional model in \cref{sssec:zeggs-baseline} under two different style inputs.
We note that stylistic interpolation should both affect shape aspects, such as the overall posture of the character, and temporal aspects, such as the gesture frequency and speed.
To investigate the shape aspects, we interpolated between the \emph{old} (hunched over)
and \emph{angry} (upright)
styles using the same speech and random seed.
Results are visualised in \cref{fig:interpshape}.
We see a definite progression in overall character posture as we interpolate from the hunched-over style to the upright one.

To demonstrate how guided interpolation affects temporal aspects of gesticulation, we interpolated between the \emph{still} and 
\emph{public speaking} styles, where the latter is very animated whereas the former has almost no motion.
The results of 10 s of animation
are shown overlaid in \cref{fig:interptemporal}, and display a clear and steady increase in the range of body movement
throughout the interpolation.
To see interpolations in motion, please refer to the project page.
\begin{figure}[!t]
\centering
\includegraphics[width=\columnwidth]{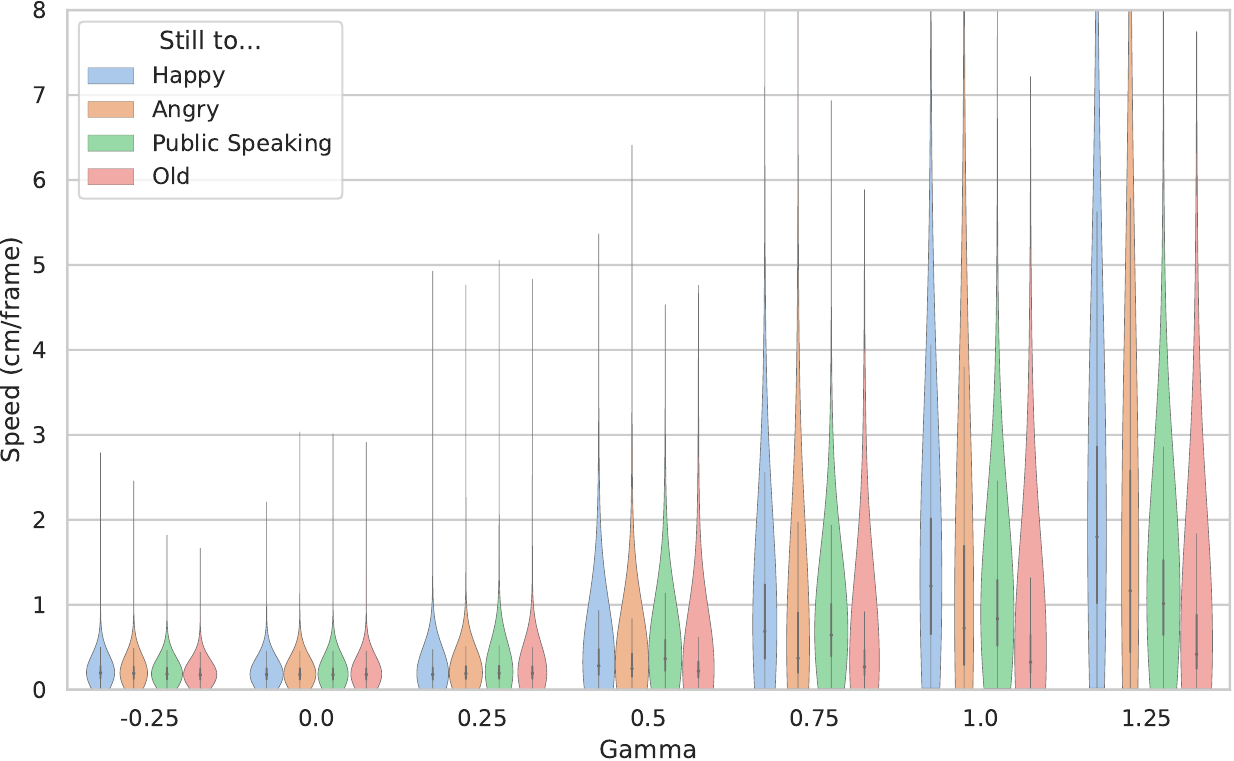}
\caption{Violin plot of wrist-speed distributions during guided interpolation between the still style and four other speaking styles (\emph{happy}, \emph{angry}, \emph{old}, and \emph{public speaking/oration}).}
\Description{Wrist-speed violin plots for gamma values between -0.25 and 1.25 in 0.25 increments. The plot for gamma equals 0 shows that the first speaking style, called Still, uses very small wrist speed, while the plot for gamma equals 1 shows that the wrist speed in the other four speaking styles are greater and have much higher variation. As the gamma value changes from 0 to 1, the violin plots smoothly change from Still into the shape of the second speaking style.}
\label{fig:objective}
\vspace{-1\baselineskip}
\end{figure}

To lend additional substance to the above observations, we performed an objective evaluation of the behaviour of guided interpolation between gesturing styles, specifically when interpolating from the 
\emph{still} style to the four styles used in our user study. We considered both endpoints, three intermediate steps, and one step of extrapolation on either side ($\gamma=\{-0.25,\,0,\,\ldots,\,1.25\}$).
To generate data for the evaluation we randomly sampled 5 motions for each 10-second audio clip 
covering the test data, resulting in 125 clips for each $\gamma$-value and style.
We then computed the instantaneous speed (magnitude of the displacement vector from the previous frame) in 3D space of the two wrists averaged together for each frame.
Hand and arm motion is central to co-speech gestures \cite{nyatsanga2023comprehensive}, and this wrist-speed distribution has been used as the basis of objective gesture evaluation before \cite{kucherenko2021moving}.

The speed distributions of the wrists are graphed in \cref{fig:objective}.
As expected, we see that model behaviour (motion statistics) change monotonically from one distribution to the other along the spectrum of $\gamma$-values in the guided interpolation.
The transition is not linear, however, since we are not interpolating in Euclidean space, but instead trading off between relatively high-probability regions of the two model distributions, to find and sample from the set of motions where the combined (weighted) probability is the greatest.

\paragraph{Interpolating between locomotion styles}
Aside from gestures, we also performed guided interpolation with other models from \cref{sec:experiments}.
Video examples are presented on the project page.
Of note is the demonstration of interpolation between the \emph{Aeroplane} and \emph{HighKnees} locomotion styles, where the exaggerated \emph{HighKnees} style also holds the arms even closer to the body than before, in clear contrast against the outstretched arms of the \emph{Aeroplane} style that has negative weight in the exaggerated barycentric combination.

\new{\paragraph{Dynamically transitioning between styles}
In addition to static style interpolation, we also demonstrate that guided interpolation can be used to dynamically transition between styles in longer motion sequences.
This is done by letting the weights $\gamma_m$ in Eq.\ \eqref{eq:ensemble} depend on $t$, so that the motion distribution gradually transitions between different models for different points in time.
This is an important control mechanism for many graphics applications, where state transitions between different motion types are frequent.
Videos of dynamic style transitions can be found on the project page.}

\paragraph{Combining highly different models} As a final demonstration, we present results from ensembling two models trained on two very different datasets and with different conditional inputs.
Specifically, we combine our style-conditional dance model from \cref{ssec:dance-experiments} with a model trained to perform mixed martial arts (MMA) motion.
The latter model was trained on a motion-capture dataset we recorded of a championship-level MMA athlete moving around in a fighting stance whilst punching, kicking, elbowing, and kneeing a focus mitt held by a sparring partner.
We trained models on this data using a similar recipe as for the dance and locomotion data, but in this case the only conditioning information was eight binary values.
These were always zero except whilst executing one of the four aforementioned combat moves using the extremities on either the left or the right side (a total of eight possible moves).
This yielded a model that could throw punches, kick, etc., on command, using the arm or leg of one's choice.
To combine this with the dancing model, we created a small choreography with a sequence of kicks and punches, whose timings were determined by the beat of a piece of music (by using the beat onset feature), enabling us to generate fighting moves in time with the music.

The results of a 50-50 dance-fighting combination of the two models highlight the nature of the product of experts to seek the intersection of two distributions: the character moves naturally and in time with the beat, but few clear dance moves or fighting moves are present, since these have low probability under at least one of the models, making them improbable also under the combined distribution.
We can see that the character generally holds their arms and hands at a level appropriate for MMA, which is a subset of the hand heights exhibited during dancing, again consistent with expectations.
To see interpolation videos for this model combination,
please consult the project page.


\section{Limitations}
\label{sec:limitations}
The most obvious and talked-about limitation of diffusion models is their slow generation speed, a consequence to the many denoising steps required to sample from the learnt models.
The architectures in our paper take around a second to generate each motion example.
Since our goal has been to advance the quality of audio-driven motion generation using diffusion, with a focus on offline generation, we have not spent any effort on optimising for synthesis speed.
We expect that synthesis time can be sped up considerably using recent innovations for diffusion models \cite{nichol2021improved,dhariwal2021diffusion,salimans2022progressive,lam2022bddm,meng2022distillation}, and that additional approaches for accomplishing this goal also will be published in the near future.
As it stands, however, synthesis speed is clearly an area that can be improved.
Our use of a parallel architecture, whilst appealing for GPU-based generation, is also not suitable for real-time interaction or integration into game engines.
There, autoregressive models like \citet{alexanderson2020style,ghorbani2022zeroeggs,siyao2022bailando} might be a better fit.

Another important limitation of the models trained in this paper is that, whilst being visually strong, they do not learn to capture all aspects that go into purposeful gesturing and dancing, such as semantics for speech or the global structure of music for a dance choreography.
There are many reasons for this, but both of these limitations can be related to the set of input features used.
Whilst speech audio in principle contains everything a human needs to make sense of a spoken message, and thus to put meaning into associated co-speech gestures, it is not possible for models to learn to make sense of language (let alone its grounding in the real world) from our acoustic features and the small amount of material seen during training.
As a consequence, one cannot expect the presented systems to generate gestures that serve a communicative function.
A starting point for improving this situation would be to follow \citet{kucherenko2020gesticulator,ahuja2020no,yoon2020speech,ao2022rhythmic} and provide information derived from text transcriptions, such as semantic word embeddings trained on large written corpora, e.g., BERT \cite{devlin2019bert}.

For dancing, it is similarly true that the information currently provided to the proposed models does not allow higher-level structure in the input to become apparent, as the feature set we used in our demonstrations is both low-dimensional and limited to musical characteristics that change rapidly over time.
These features do not have any explicit indicator of measures, verse or refrain, or the start and end of a performance, to name a few examples.
In addition, the model also only sees a few seconds of dance at a time during training, complicating the task of learning to recreate longer correlations that represent overall choreography.
A path towards enabling our models to generate more cohesive dance performance would be to consider a hierarchical modelling approach with explicit structure information, like in \citet{aristidou2022rhythm}.

For the proposed products of experts, their strong ability to focus on the intersection of a set of probability distributions can be both a strength and a weakness.
Although perfectly consistent with what the mathematics tell us, it can nonetheless sometimes be counterintuitive to see interpolation give rise to less distinctive motion, or less motion overall, than the models on either side of an interpolation spectrum.
Whilst focussing on the intersection is sometimes desirable (for instance when using a strong prior model trained on a lot of motion data to stabilise the output of a model trained on a small amount of material), that is not always the case.
An example of the latter is our mix of dancing and fighting, which most of the time engages in neither activity and appears less exciting as a result.
If one explicitly wants to generate output that alternates between different motion types in an interpolation, a conventional additive mixture model might be a better choice, for example by using a hidden Markov state to switch between different autoregressive motion models.
The individual models in such a construction may still be strong generative models such as normalising flows \cite{ghosh2020robust,ghosh2021normalizing} or diffusion models.
A more continuously adjustable version of the same effect may be achieved by changing the mixing coefficients $\gamma_m$ between the different experts in the guided interpolation framework for different points in time.

\section{Conclusions and future work}
\label{sec:conclusion}
We have introduced
diffusion models for audio-driven probabilistic modelling problems in 3D motion generation.
As part of this, we describe a new diffusion-model architecture with Conformers.
Our experiments consider several applications of audio-driven motion generation, and also path-driven locomotion, and present an extensive gold-standard evaluation of the advantages of the method against leading deep generative model baselines on several datasets.
The experimental results validate that the model outperforms previous state-of-the-art models in terms of motion quality and also enable controlling the strength of stylistic expression.
We additionally describe how to create product-of-expert ensembles with diffusion models, and show how they, for example, enable a novel type of interpolation and combining diverse diffusion models with different control parameters.

Future work includes
accelerated output generation \cite{nichol2021improved,dhariwal2021diffusion,salimans2022progressive,lam2022bddm,meng2022distillation};
conditioning motion on both audio and text, for example to enable generating semantic and communicative gestures
that synchronise with speech
(cf.\ \citet{kucherenko2020gesticulator,ahuja2020no,yoon2020speech,ao2022rhythmic}); and even synthesising multiple modalities such as speech and gesture together \cite{alexanderson2020generating,wang2021integrated}.
Separately, the community should leverage these powerful models to make a difference in applications of audio-driven motion generation, both in research, e.g., in human-agent interaction, and outside academia.

The ideas of guided interpolation and their mathematical formulation, meanwhile, open the door to a vast array of product-of-diffusion models, in ensembling, interpolation, and beyond.
For example, it also seems compelling to use pre-trained diffusion models as strong prior distributions in both supervised tasks and reinforcement learning,
as a possible alternative to using ideas like Motion\-BERT \cite{zhu2022motionbert} or fine-tuning a strong, pre-trained ``foundation model'' \cite{bommasani2021opportunities}; cf.\ \citet{holmquist2022diffpose}.
Unlike conventional fine-tuning, there is no need to train or modify a potentially large existing model.
Because the prior model is used without modification, prior information is furthermore never lost, removing the risk of catastrophic forgetting that may occur if conventional fine-tuning is run for too long.

\begin{acks}
\label{sec:acks}
We thank Esther Ericsson for the 3D characters, Shivam Mehta for the evaluation platform, and the Zero\-EGGS authors for their high-quality data and codebase. We also thank the dancers, including Mario Perez Amigo who, like Stockholm Swing All Stars, gave permission to use their music in scientific research. 

This work was partially supported by a grant from Digital Futures (KTH-RPROJ-0146472) and by the Wallenberg AI, Autonomous Systems and Software Program (WASP) funded by the Knut and Alice Wallenberg Foundation. Some of the computations were enabled by the supercomputing resource Berzelius provided by the National Supercomputer Centre at Linköping University and the Knut and Alice Wallenberg foundation. 
\end{acks}


\bibliographystyle{ACM-Reference-Format}
\bibliography{refs}


\begin{thebibliography}{153}


\ifx \showCODEN    \undefined \def \showCODEN     #1{\unskip}     \fi
\ifx \showDOI      \undefined \def \showDOI       #1{#1}\fi
\ifx \showISBNx    \undefined \def \showISBNx     #1{\unskip}     \fi
\ifx \showISBNxiii \undefined \def \showISBNxiii  #1{\unskip}     \fi
\ifx \showISSN     \undefined \def \showISSN      #1{\unskip}     \fi
\ifx \showLCCN     \undefined \def \showLCCN      #1{\unskip}     \fi
\ifx \shownote     \undefined \def \shownote      #1{#1}          \fi
\ifx \showarticletitle \undefined \def \showarticletitle #1{#1}   \fi
\ifx \showURL      \undefined \def \showURL       {\relax}        \fi
\providecommand\bibfield[2]{#2}
\providecommand\bibinfo[2]{#2}
\providecommand\natexlab[1]{#1}
\providecommand\showeprint[2][]{arXiv:#2}

\bibitem[Ahuja et~al\mbox{.}(2020a)]%
        {ahuja2020no}
\bibfield{author}{\bibinfo{person}{Chaitanya Ahuja}, \bibinfo{person}{Dong~Won
  Lee}, \bibinfo{person}{Ryo Ishii}, {and} \bibinfo{person}{Louis-Philippe
  Morency}.} \bibinfo{year}{2020}\natexlab{a}.
\newblock \showarticletitle{No Gestures Left Behind: Learning Relationships
  Between Spoken Language and Freeform Gestures}. In
  \bibinfo{booktitle}{\emph{Findings of the Association for Computational
  Linguistics}} \emph{(\bibinfo{series}{EMNLP '20})}. \bibinfo{publisher}{ACL},
  \bibinfo{pages}{1884--1895}.
\newblock
\urldef\tempurl%
\url{https://doi.org/10.18653/v1/2020.findings-emnlp.170}
\showDOI{\tempurl}


\bibitem[Ahuja et~al\mbox{.}(2020b)]%
        {ahuja2020style}
\bibfield{author}{\bibinfo{person}{Chaitanya Ahuja}, \bibinfo{person}{Dong~Won
  Lee}, \bibinfo{person}{Yukiko~I. Nakano}, {and}
  \bibinfo{person}{Louis-Philippe Morency}.} \bibinfo{year}{2020}\natexlab{b}.
\newblock \showarticletitle{Style Transfer for Co-Speech Gesture Animation: A
  Multi-Speaker Conditional-Mixture Approach}. In
  \bibinfo{booktitle}{\emph{Proceedings of the European Conference on Computer
  Vision}} \emph{(\bibinfo{series}{ECCCV '20})}. \bibinfo{pages}{248--265}.
\newblock
\urldef\tempurl%
\url{https://doi.org/10.1007/978-3-030-58523-5_15}
\showDOI{\tempurl}


\bibitem[Alexanderson(2020)]%
        {alexanderson2020stylegestures}
\bibfield{author}{\bibinfo{person}{Simon Alexanderson}.}
  \bibinfo{year}{2020}\natexlab{}.
\newblock \showarticletitle{The {S}tyle{G}estures Entry to the {GENEA}
  {C}hallenge 2020}. In \bibinfo{booktitle}{\emph{Proceedings of the
  International Workshop on Generation and Evaluation of Non-verbal Behaviour
  for Embodied Agents}} \emph{(\bibinfo{series}{GENEA '20})}.
\newblock
\urldef\tempurl%
\url{https://doi.org/10.5281/zenodo.4088599}
\showDOI{\tempurl}


\bibitem[Alexanderson et~al\mbox{.}(2020a)]%
        {alexanderson2020style}
\bibfield{author}{\bibinfo{person}{Simon Alexanderson},
  \bibinfo{person}{Gustav~Eje Henter}, \bibinfo{person}{Taras Kucherenko},
  {and} \bibinfo{person}{Jonas Beskow}.} \bibinfo{year}{2020}\natexlab{a}.
\newblock \showarticletitle{Style-Controllable Speech-Driven Gesture Synthesis
  Using Normalising Flows}.
\newblock \bibinfo{journal}{\emph{Comput. Graph. Forum}} \bibinfo{volume}{39},
  \bibinfo{number}{2} (\bibinfo{year}{2020}), \bibinfo{pages}{487--496}.
\newblock
\urldef\tempurl%
\url{https://doi.org/10.1111/cgf.13946}
\showDOI{\tempurl}


\bibitem[Alexanderson et~al\mbox{.}(2020b)]%
        {alexanderson2020generating}
\bibfield{author}{\bibinfo{person}{Simon Alexanderson},
  \bibinfo{person}{\'{E}va Sz\'{e}kely}, \bibinfo{person}{Gustav~Eje Henter},
  \bibinfo{person}{Taras Kucherenko}, {and} \bibinfo{person}{Jonas Beskow}.}
  \bibinfo{year}{2020}\natexlab{b}.
\newblock \showarticletitle{Generating Coherent Spontaneous Speech and Gesture
  from Text}. In \bibinfo{booktitle}{\emph{Proceedings of the International
  Conference on Intelligent Virtual Agents}} \emph{(\bibinfo{series}{IVA
  '20})}. \bibinfo{publisher}{ACM}, \bibinfo{pages}{1:1--1:3}.
\newblock
\urldef\tempurl%
\url{https://doi.org/10.1145/3383652.3423874}
\showDOI{\tempurl}


\bibitem[Ao et~al\mbox{.}(2022)]%
        {ao2022rhythmic}
\bibfield{author}{\bibinfo{person}{Tenglong Ao}, \bibinfo{person}{Qingzhe Gao},
  \bibinfo{person}{Yuke Lou}, \bibinfo{person}{Baoquan Chen}, {and}
  \bibinfo{person}{Libin Liu}.} \bibinfo{year}{2022}\natexlab{}.
\newblock \showarticletitle{Rhythmic Gesticulator: Rhythm-Aware Co-Speech
  Gesture Synthesis with Hierarchical Neural Embeddings}.
\newblock \bibinfo{journal}{\emph{ACM Trans. Graph.}} \bibinfo{volume}{41},
  \bibinfo{number}{6}, Article \bibinfo{articleno}{209} (\bibinfo{year}{2022}),
  \bibinfo{numpages}{19}~pages.
\newblock
\showISSN{0730-0301}
\urldef\tempurl%
\url{https://doi.org/10.1145/3550454.3555435}
\showDOI{\tempurl}


\bibitem[Ao et~al\mbox{.}(2023)]%
        {ao2023gesturediffuclip}
\bibfield{author}{\bibinfo{person}{Tenglong Ao}, \bibinfo{person}{Zeyi Zhang},
  {and} \bibinfo{person}{Libin Liu}.} \bibinfo{year}{2023}\natexlab{}.
\newblock \showarticletitle{{G}esture{D}iffu{CLIP}: Gesture Diffusion Model
  with {CLIP} Latents}.
\newblock \bibinfo{journal}{\emph{arXiv preprint arXiv:2303.14613}}
  (\bibinfo{year}{2023}).
\newblock
\urldef\tempurl%
\url{https://arxiv.org/abs/2303.14613}
\showURL{%
\tempurl}


\bibitem[Aristidou et~al\mbox{.}(2022)]%
        {aristidou2022rhythm}
\bibfield{author}{\bibinfo{person}{Andreas Aristidou},
  \bibinfo{person}{Anastasios Yiannakidis}, \bibinfo{person}{Kfir Aberman},
  \bibinfo{person}{Daniel Cohen-Or}, \bibinfo{person}{Ariel Shamir}, {and}
  \bibinfo{person}{Yiorgos Chrysanthou}.} \bibinfo{year}{2022}\natexlab{}.
\newblock \showarticletitle{Rhythm Is a Dancer: Music-Driven Motion Synthesis
  with Global Structure}.
\newblock \bibinfo{journal}{\emph{IEEE T. Vis. Comput. Gr.}}
  (\bibinfo{year}{2022}).
\newblock
\urldef\tempurl%
\url{https://doi.org/10.1109/TVCG.2022.3163676}
\showDOI{\tempurl}


\bibitem[Balaji et~al\mbox{.}(2022)]%
        {balaji2022ediffi}
\bibfield{author}{\bibinfo{person}{Yogesh Balaji}, \bibinfo{person}{Seungjun
  Nah}, \bibinfo{person}{Xun Huang}, \bibinfo{person}{Arash Vahdat},
  \bibinfo{person}{Jiaming Song}, \bibinfo{person}{Karsten Kreis},
  \bibinfo{person}{Miika Aittala}, \bibinfo{person}{Timo Aila},
  \bibinfo{person}{Samuli Laine}, \bibinfo{person}{Bryan Catanzaro},
  {et~al\mbox{.}}} \bibinfo{year}{2022}\natexlab{}.
\newblock \showarticletitle{{eDiff-I}: Text-to-Image Diffusion Models with an
  Ensemble of Expert Denoisers}.
\newblock \bibinfo{journal}{\emph{arXiv preprint arXiv:2211.01324}}
  (\bibinfo{year}{2022}).
\newblock
\urldef\tempurl%
\url{https://arxiv.org/abs/2211.01324}
\showURL{%
\tempurl}


\bibitem[Bishop(1994)]%
        {bishop1994mixture}
\bibfield{author}{\bibinfo{person}{Christopher~M. Bishop}.}
  \bibinfo{year}{1994}\natexlab{}.
\newblock \bibinfo{booktitle}{\emph{Mixture Density Networks}}.
\newblock \bibinfo{type}{{T}echnical {R}eport} NCRG/94/004.
  \bibinfo{institution}{Neural Computing Research Group, Aston University},
  \bibinfo{address}{Birmingham, UK}.
\newblock
\urldef\tempurl%
\url{https://publications.aston.ac.uk/373/1/NCRG_94_004.pdf}
\showURL{%
\tempurl}


\bibitem[B{\"o}ck et~al\mbox{.}(2016)]%
        {bock2016joint}
\bibfield{author}{\bibinfo{person}{Sebastian B{\"o}ck},
  \bibinfo{person}{Florian Krebs}, {and} \bibinfo{person}{Gerhard Widmer}.}
  \bibinfo{year}{2016}\natexlab{}.
\newblock \showarticletitle{Joint Beat and Downbeat Tracking with Recurrent
  Neural Networks}. In \bibinfo{booktitle}{\emph{Proceedings of the
  International Society for Music Information Retrieval Conference}}
  \emph{(\bibinfo{series}{ISMIR '16})}. \bibinfo{pages}{255--261}.
\newblock
\urldef\tempurl%
\url{https://archives.ismir.net/ismir2016/paper/000186.pdf}
\showURL{%
\tempurl}


\bibitem[Bommasani et~al\mbox{.}(2021)]%
        {bommasani2021opportunities}
\bibfield{author}{\bibinfo{person}{Rishi Bommasani}, \bibinfo{person}{Drew~A.
  Hudson}, \bibinfo{person}{Ehsan Adeli}, \bibinfo{person}{Russ Altman},
  \bibinfo{person}{Simran Arora}, \bibinfo{person}{Sydney von Arx},
  \bibinfo{person}{Michael~S. Bernstein}, \bibinfo{person}{Jeannette Bohg},
  \bibinfo{person}{Antoine Bosselut}, \bibinfo{person}{Emma Brunskill},
  {et~al\mbox{.}}} \bibinfo{year}{2021}\natexlab{}.
\newblock \showarticletitle{On the Opportunities and Risks of Foundation
  Models}.
\newblock \bibinfo{journal}{\emph{arXiv preprint arXiv:2108.07258}}
  (\bibinfo{year}{2021}).
\newblock
\urldef\tempurl%
\url{https://crfm.stanford.edu/assets/report.pdf}
\showURL{%
\tempurl}


\bibitem[Bosker and Peeters(2021)]%
        {bosker2021beat}
\bibfield{author}{\bibinfo{person}{Hans~Rutger Bosker} {and}
  \bibinfo{person}{David Peeters}.} \bibinfo{year}{2021}\natexlab{}.
\newblock \showarticletitle{Beat Gestures Influence Which Speech Sounds You
  Hear}.
\newblock \bibinfo{journal}{\emph{P. R. Soc. B}} \bibinfo{volume}{288},
  \bibinfo{number}{1943} (\bibinfo{year}{2021}), \bibinfo{pages}{20202419}.
\newblock
\urldef\tempurl%
\url{https://doi.org/10.1098/rspb.2020.2419}
\showDOI{\tempurl}


\bibitem[Bozkurt et~al\mbox{.}(2020)]%
        {bozkurt2020affective}
\bibfield{author}{\bibinfo{person}{Elif Bozkurt}, \bibinfo{person}{Y{\"u}cel
  Yemez}, {and} \bibinfo{person}{Engin Erzin}.}
  \bibinfo{year}{2020}\natexlab{}.
\newblock \showarticletitle{Affective Synthesis and Animation of Arm Gestures
  from Speech Prosody}.
\newblock \bibinfo{journal}{\emph{Speech Commun.}}  \bibinfo{volume}{119}
  (\bibinfo{year}{2020}), \bibinfo{pages}{1--11}.
\newblock
\urldef\tempurl%
\url{https://doi.org/10.1016/j.specom.2020.02.005}
\showDOI{\tempurl}


\bibitem[Cassell et~al\mbox{.}(2001)]%
        {cassell2001beat}
\bibfield{author}{\bibinfo{person}{Justine Cassell},
  \bibinfo{person}{Hannes~H{\"o}gni Vilhj{\'a}lmsson}, {and}
  \bibinfo{person}{Timothy Bickmore}.} \bibinfo{year}{2001}\natexlab{}.
\newblock \showarticletitle{{BEAT}: The Behavior Expression Animation Toolkit}.
  In \bibinfo{booktitle}{\emph{Annual Conference on Computer Graphics and
  Interactive Techniques}} \emph{(\bibinfo{series}{SIGGRAPH '01})}.
  \bibinfo{publisher}{ACM}, \bibinfo{pages}{477–486}.
\newblock
\urldef\tempurl%
\url{https://doi.org/10.1145/383259.383315}
\showDOI{\tempurl}


\bibitem[Castillo and Neff(2019)]%
        {castillo2019we}
\bibfield{author}{\bibinfo{person}{Gabriel Castillo} {and}
  \bibinfo{person}{Michael Neff}.} \bibinfo{year}{2019}\natexlab{}.
\newblock \showarticletitle{What Do We Express Without Knowing? Emotion in
  Gesture}. In \bibinfo{booktitle}{\emph{Proceedings of the International
  Conference on Autonomous Agents and Multiagent Systems}}
  \emph{(\bibinfo{series}{AAMAS '19})}. \bibinfo{publisher}{IFAAMAS},
  \bibinfo{pages}{702--710}.
\newblock
\urldef\tempurl%
\url{https://www.ifaamas.org/Proceedings/aamas2019/pdfs/p702.pdf}
\showURL{%
\tempurl}


\bibitem[Chen et~al\mbox{.}(2021a)]%
        {chen2021choreomaster}
\bibfield{author}{\bibinfo{person}{Kang Chen}, \bibinfo{person}{Zhipeng Tan},
  \bibinfo{person}{Jin Lei}, \bibinfo{person}{Song-Hai Zhang},
  \bibinfo{person}{Yuan-Chen Guo}, \bibinfo{person}{Weidong Zhang}, {and}
  \bibinfo{person}{Shi-Min Hu}.} \bibinfo{year}{2021}\natexlab{a}.
\newblock \showarticletitle{{C}horeo{M}aster: Choreography-Oriented
  Music-Driven Dance Synthesis}.
\newblock \bibinfo{journal}{\emph{ACM Trans. Graph.}} \bibinfo{volume}{40},
  \bibinfo{number}{4}, Article \bibinfo{articleno}{145} (\bibinfo{year}{2021}),
  \bibinfo{numpages}{13}~pages.
\newblock
\urldef\tempurl%
\url{https://doi.org/10.1145/3450626.3459932}
\showDOI{\tempurl}


\bibitem[Chen et~al\mbox{.}(2021b)]%
        {chen2021wavegrad}
\bibfield{author}{\bibinfo{person}{Nanxin Chen}, \bibinfo{person}{Yu Zhang},
  \bibinfo{person}{Heiga Zen}, \bibinfo{person}{Ron~J. Weiss},
  \bibinfo{person}{Mohammad Norouzi}, {and} \bibinfo{person}{William Chan}.}
  \bibinfo{year}{2021}\natexlab{b}.
\newblock \showarticletitle{{W}ave{G}rad: Estimating Gradients for Waveform
  Generation}. In \bibinfo{booktitle}{\emph{Proceedings of the International
  Conference on Learning Representations}} \emph{(\bibinfo{series}{ICLR '21})}.
\newblock
\urldef\tempurl%
\url{https://openreview.net/forum?id=NsMLjcFaO8O}
\showURL{%
\tempurl}


\bibitem[Cheng et~al\mbox{.}(2016)]%
        {cheng2016long}
\bibfield{author}{\bibinfo{person}{Jianpeng Cheng}, \bibinfo{person}{Li Dong},
  {and} \bibinfo{person}{Mirella Lapata}.} \bibinfo{year}{2016}\natexlab{}.
\newblock \showarticletitle{Long Short-Term Memory-Networks for Machine
  Reading}. In \bibinfo{booktitle}{\emph{Proceedings of the Conference on
  Empirical Methods in Natural Language Processing}}
  \emph{(\bibinfo{series}{EMNLP '16})}. \bibinfo{publisher}{ACL},
  \bibinfo{pages}{551--561}.
\newblock
\urldef\tempurl%
\url{https://doi.org/10.18653/v1/D16-1053}
\showDOI{\tempurl}


\bibitem[Chiu et~al\mbox{.}(2015)]%
        {chiu2015predicting}
\bibfield{author}{\bibinfo{person}{Chung-Cheng Chiu},
  \bibinfo{person}{Louis-Philippe Morency}, {and} \bibinfo{person}{Stacy
  Marsella}.} \bibinfo{year}{2015}\natexlab{}.
\newblock \showarticletitle{Predicting Co-Verbal Gestures: A Deep and Temporal
  Modeling Approach}. In \bibinfo{booktitle}{\emph{Proceedings of the
  International Conference on Intelligent Virtual Agents}}.
  \bibinfo{pages}{152--166}.
\newblock
\urldef\tempurl%
\url{https://doi.org/10.1007/978-3-319-21996-7_17}
\showDOI{\tempurl}


\bibitem[Clavet(2016)]%
        {clavet2016motion}
\bibfield{author}{\bibinfo{person}{Simon Clavet}.}
  \bibinfo{year}{2016}\natexlab{}.
\newblock \showarticletitle{Motion Matching and the Road to Next-Gen
  Animation}. In \bibinfo{booktitle}{\emph{Proceedings of the Game Developers
  Conference}} \emph{(\bibinfo{series}{GDC '16})}.
\newblock
\urldef\tempurl%
\url{https://www.gdcvault.com/play/1023280/Motion-Matching-and-The-Road}
\showURL{%
\tempurl}


\bibitem[Crnkovic-Friis and Crnkovic-Friis(2016)]%
        {crnkovic2016generative}
\bibfield{author}{\bibinfo{person}{Luka Crnkovic-Friis} {and}
  \bibinfo{person}{Louise Crnkovic-Friis}.} \bibinfo{year}{2016}\natexlab{}.
\newblock \showarticletitle{Generative Choreography Using Deep Learning}. In
  \bibinfo{booktitle}{\emph{International Conference on Computational
  Creativity}} \emph{(\bibinfo{series}{ICCC '16})}. \bibinfo{publisher}{ACC},
  \bibinfo{pages}{272--277}.
\newblock
\urldef\tempurl%
\url{http://www.computationalcreativity.net/iccc2016/wp-content/uploads/2016/01/Generative-Choreography-using-Deep-Learning.pdf}
\showURL{%
\tempurl}


\bibitem[Dabral et~al\mbox{.}(2022)]%
        {dabral2023mofusion}
\bibfield{author}{\bibinfo{person}{Rishabh Dabral},
  \bibinfo{person}{Muhammad~Hamza Mughal}, \bibinfo{person}{Vladislav
  Golyanik}, {and} \bibinfo{person}{Christian Theobalt}.}
  \bibinfo{year}{2022}\natexlab{}.
\newblock \showarticletitle{{M}o{F}usion: A Framework for
  Denoising-Diffusion-Based Motion Synthesis}. In
  \bibinfo{booktitle}{\emph{Proceedings of the IEEE/CVF Conference on Computer
  Vision and Pattern Recognition}} \emph{(\bibinfo{series}{CVPR '23})}.
\newblock
\urldef\tempurl%
\url{https://arxiv.org/abs/2212.04495}
\showURL{%
\tempurl}


\bibitem[Davis and Mermelstein(1980)]%
        {davis1980comparison}
\bibfield{author}{\bibinfo{person}{Steven Davis} {and} \bibinfo{person}{Paul
  Mermelstein}.} \bibinfo{year}{1980}\natexlab{}.
\newblock \showarticletitle{Comparison of Parametric Representations for
  Monosyllabic Word Recognition in Continuously Spoken Sentences}.
\newblock \bibinfo{journal}{\emph{IEEE T. Acoust. Speech}}
  \bibinfo{volume}{28}, \bibinfo{number}{4} (\bibinfo{year}{1980}),
  \bibinfo{pages}{357--366}.
\newblock
\urldef\tempurl%
\url{https://doi.org/10.1109/TASSP.1980.1163420}
\showDOI{\tempurl}


\bibitem[Devlin et~al\mbox{.}(2019)]%
        {devlin2019bert}
\bibfield{author}{\bibinfo{person}{Jacob Devlin}, \bibinfo{person}{Ming-Wei
  Chang}, \bibinfo{person}{Kenton Lee}, {and} \bibinfo{person}{Kristina
  Toutanova}.} \bibinfo{year}{2019}\natexlab{}.
\newblock \showarticletitle{{BERT}: Pre-Training of Deep Bidirectional
  Transformers for Language Understanding}. In
  \bibinfo{booktitle}{\emph{Proceedings of the Conference of the North American
  Chapter of the Association for Computational Linguistics}}
  \emph{(\bibinfo{series}{NAACL '19})}. \bibinfo{publisher}{ACL},
  \bibinfo{pages}{4171--4186}.
\newblock
\urldef\tempurl%
\url{https://doi.org/10.18653/v1/N19-1423}
\showDOI{\tempurl}


\bibitem[Dhariwal and Nichol(2021)]%
        {dhariwal2021diffusion}
\bibfield{author}{\bibinfo{person}{Prafulla Dhariwal} {and}
  \bibinfo{person}{Alexander Nichol}.} \bibinfo{year}{2021}\natexlab{}.
\newblock \showarticletitle{Diffusion Models Beat {GAN}s on Image Synthesis}.
  In \bibinfo{booktitle}{\emph{Advances in Neural Information Processing
  Systems}} \emph{(\bibinfo{series}{NeurIPS '21})}.
  \bibinfo{pages}{8780--8794}.
\newblock
\urldef\tempurl%
\url{https://proceedings.neurips.cc/paper_files/paper/2021/file/49ad23d1ec9fa4bd8d77d02681df5cfa-Paper.pdf}
\showURL{%
\tempurl}


\bibitem[Dieleman(2022)]%
        {dieleman2022guidance}
\bibfield{author}{\bibinfo{person}{Sander Dieleman}.}
  \bibinfo{year}{2022}\natexlab{}.
\newblock \bibinfo{title}{Guidance: A Cheat Code for Diffusion Models}.
\newblock
  \bibinfo{howpublished}{\href{https://sander.ai/2022/05/26/guidance.html}{https://sander.ai/2022/05/26/guidance.html}}.
\newblock


\bibitem[Durupinar et~al\mbox{.}(2016)]%
        {durupinar2016perform}
\bibfield{author}{\bibinfo{person}{Funda Durupinar}, \bibinfo{person}{Mubbasir
  Kapadia}, \bibinfo{person}{Susan Deutsch}, \bibinfo{person}{Michael Neff},
  {and} \bibinfo{person}{Norman~I. Badler}.} \bibinfo{year}{2016}\natexlab{}.
\newblock \showarticletitle{{PERFORM}: Perceptual Approach for Adding {OCEAN}
  Personality to Human Motion Using {L}aban Movement Analysis}.
\newblock \bibinfo{journal}{\emph{ACM Trans. Graph.}} \bibinfo{volume}{36},
  \bibinfo{number}{1}, Article \bibinfo{articleno}{6} (\bibinfo{year}{2016}),
  \bibinfo{numpages}{16}~pages.
\newblock
\urldef\tempurl%
\url{https://doi.org/10.1145/2983620}
\showDOI{\tempurl}


\bibitem[Fan et~al\mbox{.}(2011)]%
        {fan2011example}
\bibfield{author}{\bibinfo{person}{Rukun Fan}, \bibinfo{person}{Songhua Xu},
  {and} \bibinfo{person}{Weidong Geng}.} \bibinfo{year}{2011}\natexlab{}.
\newblock \showarticletitle{Example-Based Automatic Music-Driven Conventional
  Dance Motion Synthesis}.
\newblock \bibinfo{journal}{\emph{IEEE T. Vis. Comput. Gr.}}
  \bibinfo{volume}{18}, \bibinfo{number}{3} (\bibinfo{year}{2011}),
  \bibinfo{pages}{501--515}.
\newblock
\urldef\tempurl%
\url{https://doi.org/10.1109/TVCG.2011.73}
\showDOI{\tempurl}


\bibitem[Fares et~al\mbox{.}(2022)]%
        {fares2022zero}
\bibfield{author}{\bibinfo{person}{Mireille Fares}, \bibinfo{person}{Michele
  Grimaldi}, \bibinfo{person}{Catherine Pelachaud}, {and}
  \bibinfo{person}{Nicolas Obin}.} \bibinfo{year}{2022}\natexlab{}.
\newblock \showarticletitle{Zero-Shot Style Transfer for Gesture Animation
  Driven by Text and Speech Using Adversarial Disentanglement of Multimodal
  Style Encoding}.
\newblock \bibinfo{journal}{\emph{arXiv preprint arXiv:2208.01917}}
  (\bibinfo{year}{2022}).
\newblock
\urldef\tempurl%
\url{https://arxiv.org/abs/2208.01917}
\showURL{%
\tempurl}


\bibitem[Ferstl and McDonnell(2018)]%
        {ferstl2018trinity}
\bibfield{author}{\bibinfo{person}{Ylva Ferstl} {and} \bibinfo{person}{Rachel
  McDonnell}.} \bibinfo{year}{2018}\natexlab{}.
\newblock \showarticletitle{Investigating the Use of Recurrent Motion Modelling
  for Speech Gesture Generation}. In \bibinfo{booktitle}{\emph{Proceedings of
  the ACM International Conference on Intelligent Virtual Agents}}
  \emph{(\bibinfo{series}{IVA '18})}. \bibinfo{publisher}{ACM},
  \bibinfo{pages}{93--98}.
\newblock
\urldef\tempurl%
\url{https://trinityspeechgesture.scss.tcd.ie}
\showURL{%
\tempurl}


\bibitem[Ferstl et~al\mbox{.}(2019)]%
        {ferstl2019multi}
\bibfield{author}{\bibinfo{person}{Ylva Ferstl}, \bibinfo{person}{Michael
  Neff}, {and} \bibinfo{person}{Rachel McDonnell}.}
  \bibinfo{year}{2019}\natexlab{}.
\newblock \showarticletitle{Multi-Objective Adversarial Gesture Generation}. In
  \bibinfo{booktitle}{\emph{Proceedings of the ACM SIGGRAPH Conference on
  Motion, Interaction and Games}} \emph{(\bibinfo{series}{MIG'19})}.
  \bibinfo{publisher}{ACM}, Article \bibinfo{articleno}{3},
  \bibinfo{numpages}{10}~pages.
\newblock
\urldef\tempurl%
\url{https://doi.org/10.1145/3359566.3360053}
\showDOI{\tempurl}


\bibitem[Findlay et~al\mbox{.}(2022)]%
        {findlay2022denoising}
\bibfield{author}{\bibinfo{person}{Edmund J.~C. Findlay},
  \bibinfo{person}{Haozheng Zhang}, \bibinfo{person}{Ziyi Chang}, {and}
  \bibinfo{person}{Hubert P.~H. Shum}.} \bibinfo{year}{2022}\natexlab{}.
\newblock \showarticletitle{Denoising Diffusion Probabilistic Models for Styled
  Walking Synthesis}. In \bibinfo{booktitle}{\emph{Proceedings of the ACM
  SIGGRAPH Conference on Motion, Interaction and Games}}
  \emph{(\bibinfo{series}{MIG '22})}. \bibinfo{publisher}{ACM}.
\newblock
\urldef\tempurl%
\url{https://arxiv.org/abs/2209.14828}
\showURL{%
\tempurl}


\bibitem[Fourati and Pelachaud(2016)]%
        {fourati2016perception}
\bibfield{author}{\bibinfo{person}{Nesrine Fourati} {and}
  \bibinfo{person}{Catherine Pelachaud}.} \bibinfo{year}{2016}\natexlab{}.
\newblock \showarticletitle{Perception of Emotions and Body Movement in the
  {E}milya Database}.
\newblock \bibinfo{journal}{\emph{IEEE T. Affect. Comput.}}
  \bibinfo{volume}{9}, \bibinfo{number}{1} (\bibinfo{year}{2016}),
  \bibinfo{pages}{90--101}.
\newblock
\urldef\tempurl%
\url{https://doi.org/10.1109/TAFFC.2016.2591039}
\showDOI{\tempurl}


\bibitem[Fragkiadaki et~al\mbox{.}(2015)]%
        {fragkiadaki2015recurrent}
\bibfield{author}{\bibinfo{person}{Katerina Fragkiadaki},
  \bibinfo{person}{Sergey Levine}, \bibinfo{person}{Panna Felsen}, {and}
  \bibinfo{person}{Jitendra Malik}.} \bibinfo{year}{2015}\natexlab{}.
\newblock \showarticletitle{Recurrent Network Models for Human Dynamics}. In
  \bibinfo{booktitle}{\emph{Proceedings of the IEEE/CVF International
  Conference on Computer Vision}} \emph{(\bibinfo{series}{ICCV '15})}.
  \bibinfo{publisher}{IEEE}, \bibinfo{pages}{4346--4354}.
\newblock
\urldef\tempurl%
\url{https://doi.org/10.1109/ICCV.2015.494}
\showDOI{\tempurl}


\bibitem[Fukayama and Goto(2015)]%
        {fukayama2015music}
\bibfield{author}{\bibinfo{person}{Satoru Fukayama} {and}
  \bibinfo{person}{Masataka Goto}.} \bibinfo{year}{2015}\natexlab{}.
\newblock \showarticletitle{Music Content Driven Automated Choreography with
  Beat-Wise Motion Connectivity Constraints}.
\newblock \bibinfo{journal}{\emph{IEEE International Conference on Systems,
  Man, and Cybernetics}} (\bibinfo{year}{2015}), \bibinfo{pages}{177--183}.
\newblock


\bibitem[Ghorbani et~al\mbox{.}(2022)]%
        {ghorbani2022exemplar}
\bibfield{author}{\bibinfo{person}{Saeed Ghorbani}, \bibinfo{person}{Ylva
  Ferstl}, {and} \bibinfo{person}{Marc-Andr{\'e} Carbonneau}.}
  \bibinfo{year}{2022}\natexlab{}.
\newblock \showarticletitle{Exemplar-Based Stylized Gesture Generation from
  Speech: An Entry to the {GENEA} {C}hallenge 2022}. In
  \bibinfo{booktitle}{\emph{Proceedings of the ACM International Conference on
  Multimodal Interaction}} \emph{(\bibinfo{series}{ICMI '22})}.
  \bibinfo{publisher}{ACM}, \bibinfo{pages}{778--783}.
\newblock
\urldef\tempurl%
\url{https://doi.org/10.1145/3536221.3558068}
\showDOI{\tempurl}


\bibitem[Ghorbani et~al\mbox{.}(2023)]%
        {ghorbani2022zeroeggs}
\bibfield{author}{\bibinfo{person}{Saeed Ghorbani}, \bibinfo{person}{Ylva
  Ferstl}, \bibinfo{person}{Daniel Holden}, \bibinfo{person}{Nikolaus~F.
  Troje}, {and} \bibinfo{person}{Marc-André Carbonneau}.}
  \bibinfo{year}{2023}\natexlab{}.
\newblock \showarticletitle{{Z}ero{EGGS}: Zero-Shot Example-Based Gesture
  Generation from Speech}.
\newblock \bibinfo{journal}{\emph{Comput. Graph. Forum}} \bibinfo{volume}{42},
  \bibinfo{number}{1} (\bibinfo{year}{2023}), \bibinfo{pages}{206--216}.
\newblock
\urldef\tempurl%
\url{https://doi.org/10.1111/cgf.14734}
\showDOI{\tempurl}


\bibitem[Ghosh et~al\mbox{.}(2021a)]%
        {ghosh2021synthesis}
\bibfield{author}{\bibinfo{person}{Anindita Ghosh}, \bibinfo{person}{Noshaba
  Cheema}, \bibinfo{person}{Cennet Oguz}, \bibinfo{person}{Christian Theobalt},
  {and} \bibinfo{person}{Philipp Slusallek}.} \bibinfo{year}{2021}\natexlab{a}.
\newblock \showarticletitle{Synthesis of Compositional Animations from Textual
  Descriptions}. In \bibinfo{booktitle}{\emph{Proceedings of the IEEE/CVF
  International Conference on Computer Vision}} \emph{(\bibinfo{series}{ICCV
  '21})}. \bibinfo{pages}{1396--1406}.
\newblock
\urldef\tempurl%
\url{https://doi.org/10.1109/ICCV48922.2021.00143}
\showDOI{\tempurl}


\bibitem[Ghosh et~al\mbox{.}(2020)]%
        {ghosh2020robust}
\bibfield{author}{\bibinfo{person}{Anubhab Ghosh}, \bibinfo{person}{Antoine
  Honor{\'e}}, \bibinfo{person}{Dong Liu}, \bibinfo{person}{Gustav~Eje Henter},
  {and} \bibinfo{person}{Saikat Chatterjee}.} \bibinfo{year}{2020}\natexlab{}.
\newblock \showarticletitle{Robust Classification Using Hidden {M}arkov Models
  and Mixtures of Normalizing Flows}. In \bibinfo{booktitle}{\emph{Proceedings
  of the IEEE International Workshop on Machine Learning for Signal
  Processing}} \emph{(\bibinfo{series}{MLSP '20})}. \bibinfo{publisher}{IEEE}.
\newblock
\urldef\tempurl%
\url{https://doi.org/10.1109/MLSP49062.2020.9231775}
\showDOI{\tempurl}


\bibitem[Ghosh et~al\mbox{.}(2021b)]%
        {ghosh2021normalizing}
\bibfield{author}{\bibinfo{person}{Anubhab Ghosh}, \bibinfo{person}{Antoine
  Honor{\'e}}, \bibinfo{person}{Dong Liu}, \bibinfo{person}{Gustav~Eje Henter},
  {and} \bibinfo{person}{Saikat Chatterjee}.} \bibinfo{year}{2021}\natexlab{b}.
\newblock \showarticletitle{Normalizing Flow Based Hidden {M}arkov Models for
  Classification of Speech Phones with Explainability}.
\newblock \bibinfo{journal}{\emph{arXiv preprint arXiv:2107.00730}}
  (\bibinfo{year}{2021}).
\newblock
\urldef\tempurl%
\url{https://arxiv.org/abs/2107.00730}
\showURL{%
\tempurl}


\bibitem[Grassia(1998)]%
        {grassia1998practical}
\bibfield{author}{\bibinfo{person}{F.~Sebastian Grassia}.}
  \bibinfo{year}{1998}\natexlab{}.
\newblock \showarticletitle{Practical Parameterization of Rotations Using the
  Exponential Map}.
\newblock \bibinfo{journal}{\emph{J. Graph. Tools}} \bibinfo{volume}{3},
  \bibinfo{number}{3} (\bibinfo{year}{1998}), \bibinfo{pages}{29--48}.
\newblock
\urldef\tempurl%
\url{https://doi.org/10.1080/10867651.1998.10487493}
\showDOI{\tempurl}


\bibitem[Gulati et~al\mbox{.}(2020)]%
        {gulati2020conformer}
\bibfield{author}{\bibinfo{person}{Anmol Gulati}, \bibinfo{person}{James Qin},
  \bibinfo{person}{Chung-Cheng Chiu}, \bibinfo{person}{Niki Parmar},
  \bibinfo{person}{Yu Zhang}, \bibinfo{person}{Jiahui Yu}, \bibinfo{person}{Wei
  Han}, \bibinfo{person}{Shibo Wang}, \bibinfo{person}{Zhengdong Zhang},
  \bibinfo{person}{Yonghui Wu}, {et~al\mbox{.}}}
  \bibinfo{year}{2020}\natexlab{}.
\newblock \showarticletitle{Conformer: Convolution-Augmented Transformer for
  Speech Recognition}. In \bibinfo{booktitle}{\emph{Proceedings of the Annual
  Conference of the International Speech Communication Association}}
  \emph{(\bibinfo{series}{Interspeech '20})}. \bibinfo{publisher}{ISCA},
  \bibinfo{pages}{5036--5040}.
\newblock
\urldef\tempurl%
\url{https://doi.org/10.21437/Interspeech.2020-3015}
\showDOI{\tempurl}


\bibitem[Guo et~al\mbox{.}(2022)]%
        {guo2022tm2t}
\bibfield{author}{\bibinfo{person}{Chuan Guo}, \bibinfo{person}{Xinxin Zuo},
  \bibinfo{person}{Sen Wang}, {and} \bibinfo{person}{Li Cheng}.}
  \bibinfo{year}{2022}\natexlab{}.
\newblock \showarticletitle{{TM2T}: Stochastic and Tokenized Modeling for the
  Reciprocal Generation of 3{D} Human Motions and Texts}. In
  \bibinfo{booktitle}{\emph{Proceedings of the European Conference on Computer
  Vision}} \emph{(\bibinfo{series}{ECCV '22})}. \bibinfo{pages}{580--597}.
\newblock
\urldef\tempurl%
\url{https://doi.org/10.1007/978-3-031-19833-5_34}
\showDOI{\tempurl}


\bibitem[Habibie et~al\mbox{.}(2022)]%
        {habibie2022motion}
\bibfield{author}{\bibinfo{person}{Ikhsanul Habibie}, \bibinfo{person}{Mohamed
  Elgharib}, \bibinfo{person}{Kripasindhu Sarkar}, \bibinfo{person}{Ahsan
  Abdullah}, \bibinfo{person}{Simbarashe Nyatsanga}, \bibinfo{person}{Michael
  Neff}, {and} \bibinfo{person}{Christian Theobalt}.}
  \bibinfo{year}{2022}\natexlab{}.
\newblock \showarticletitle{A Motion Matching-Based Framework for Controllable
  Gesture Synthesis from Speech}. In \bibinfo{booktitle}{\emph{ACM Special
  Interest Group on Computer Graphics and Interactive Techniques Conference
  Proceedings}} \emph{(\bibinfo{series}{SIGGRAPH '22})}.
  \bibinfo{publisher}{ACM}, Article \bibinfo{articleno}{46},
  \bibinfo{numpages}{9}~pages.
\newblock
\urldef\tempurl%
\url{https://doi.org/10.1145/3528233.3530750}
\showDOI{\tempurl}


\bibitem[Habibie et~al\mbox{.}(2017)]%
        {habibie2017recurrent}
\bibfield{author}{\bibinfo{person}{Ikhansul Habibie}, \bibinfo{person}{Daniel
  Holden}, \bibinfo{person}{Jonathan Schwarz}, \bibinfo{person}{Joe Yearsley},
  {and} \bibinfo{person}{Taku Komura}.} \bibinfo{year}{2017}\natexlab{}.
\newblock \showarticletitle{A Recurrent Variational Autoencoder for Human
  Motion Synthesis}. In \bibinfo{booktitle}{\emph{Proceedings of the British
  Machine Vision Conference}} \emph{(\bibinfo{series}{BMVC '17})}.
  \bibinfo{publisher}{BMVA Press}, Article \bibinfo{articleno}{119},
  \bibinfo{numpages}{12}~pages.
\newblock
\urldef\tempurl%
\url{https://doi.org/10.5244/C.31.119}
\showDOI{\tempurl}


\bibitem[Habibie et~al\mbox{.}(2021)]%
        {habibie2021learning}
\bibfield{author}{\bibinfo{person}{Ikhsanul Habibie}, \bibinfo{person}{Weipeng
  Xu}, \bibinfo{person}{Dushyant Mehta}, \bibinfo{person}{Lingjie Liu},
  \bibinfo{person}{Hans-Peter Seidel}, \bibinfo{person}{Gerard Pons-Moll},
  \bibinfo{person}{Mohamed Elgharib}, {and} \bibinfo{person}{Christian
  Theobalt}.} \bibinfo{year}{2021}\natexlab{}.
\newblock \showarticletitle{Learning Speech-Driven 3{D} Conversational Gestures
  from Video}. In \bibinfo{booktitle}{\emph{Proceedings of the ACM
  International Conference on Intelligent Virtual Agents}}
  \emph{(\bibinfo{series}{IVA '21})}. \bibinfo{publisher}{ACM},
  \bibinfo{pages}{101--108}.
\newblock
\urldef\tempurl%
\url{https://doi.org/10.1145/3472306.3478335}
\showDOI{\tempurl}


\bibitem[Hasegawa et~al\mbox{.}(2018)]%
        {hasegawa2018evaluation}
\bibfield{author}{\bibinfo{person}{Dai Hasegawa}, \bibinfo{person}{Naoshi
  Kaneko}, \bibinfo{person}{Shinichi Shirakawa}, \bibinfo{person}{Hiroshi
  Sakuta}, {and} \bibinfo{person}{Kazuhiko Sumi}.}
  \bibinfo{year}{2018}\natexlab{}.
\newblock \showarticletitle{Evaluation of Speech-to-Gesture Generation Using
  Bi-Directional {LSTM} Network}. In \bibinfo{booktitle}{\emph{Proceedings of
  the ACM International Conference on Intelligent Virtual Agents}}
  \emph{(\bibinfo{series}{IVA '18})}. \bibinfo{publisher}{ACM},
  \bibinfo{pages}{79--86}.
\newblock
\urldef\tempurl%
\url{https://doi.org/10.1145/3267851.3267878}
\showDOI{\tempurl}


\bibitem[Henter and Kleijn(2016)]%
        {henter2016minimum}
\bibfield{author}{\bibinfo{person}{Gustav~Eje Henter} {and}
  \bibinfo{person}{W.~Bastiaan Kleijn}.} \bibinfo{year}{2016}\natexlab{}.
\newblock \showarticletitle{Minimum Entropy Rate Simplification of Stochastic
  Processes}.
\newblock \bibinfo{journal}{\emph{IEEE T. Pattern Anal.}} \bibinfo{volume}{38},
  \bibinfo{number}{12} (\bibinfo{year}{2016}), \bibinfo{pages}{2487--2500}.
\newblock
\urldef\tempurl%
\url{https://doi.org/10.1109/TPAMI.2016.2533382}
\showDOI{\tempurl}


\bibitem[Heusel et~al\mbox{.}(2017)]%
        {heusel2017gans}
\bibfield{author}{\bibinfo{person}{Martin Heusel}, \bibinfo{person}{Hubert
  Ramsauer}, \bibinfo{person}{Thomas Unterthiner}, \bibinfo{person}{Bernhard
  Nessler}, {and} \bibinfo{person}{Sepp Hochreiter}.}
  \bibinfo{year}{2017}\natexlab{}.
\newblock \showarticletitle{{GAN}s Trained by a Two Time-Scale Update Rule
  Converge to a Local {N}ash Equilibrium}. In
  \bibinfo{booktitle}{\emph{Advances in Neural Information Processing Systems}}
  \emph{(\bibinfo{series}{NIPS '17})}.
\newblock
\urldef\tempurl%
\url{https://proceedings.neurips.cc/paper_files/paper/2017/file/8a1d694707eb0fefe65871369074926d-Paper.pdf}
\showURL{%
\tempurl}


\bibitem[Hinton(2002)]%
        {hinton2002training}
\bibfield{author}{\bibinfo{person}{Geoffrey~E. Hinton}.}
  \bibinfo{year}{2002}\natexlab{}.
\newblock \showarticletitle{Training Products of Experts by Minimizing
  Contrastive Divergence}.
\newblock \bibinfo{journal}{\emph{Neural Comput.}} \bibinfo{volume}{14},
  \bibinfo{number}{8} (\bibinfo{year}{2002}), \bibinfo{pages}{1771--1800}.
\newblock
\urldef\tempurl%
\url{https://doi.org/10.1162/089976602760128018}
\showDOI{\tempurl}


\bibitem[Ho et~al\mbox{.}(2022a)]%
        {ho2022imagen}
\bibfield{author}{\bibinfo{person}{Jonathan Ho}, \bibinfo{person}{William
  Chan}, \bibinfo{person}{Chitwan Saharia}, \bibinfo{person}{Jay Whang},
  \bibinfo{person}{Ruiqi Gao}, \bibinfo{person}{Alexey Gritsenko},
  \bibinfo{person}{Diederik~P. Kingma}, \bibinfo{person}{Ben Poole},
  \bibinfo{person}{Mohammad Norouzi}, \bibinfo{person}{David~J. Fleet},
  {et~al\mbox{.}}} \bibinfo{year}{2022}\natexlab{a}.
\newblock \showarticletitle{{I}magen {V}ideo: High Definition Video Generation
  with Diffusion Models}.
\newblock \bibinfo{journal}{\emph{arXiv preprint arXiv:2210.02303}}
  (\bibinfo{year}{2022}).
\newblock
\urldef\tempurl%
\url{https://arxiv.org/abs/2210.02303}
\showURL{%
\tempurl}


\bibitem[Ho et~al\mbox{.}(2020)]%
        {ho2020denoising}
\bibfield{author}{\bibinfo{person}{Jonathan Ho}, \bibinfo{person}{Ajay Jain},
  {and} \bibinfo{person}{Pieter Abbeel}.} \bibinfo{year}{2020}\natexlab{}.
\newblock \showarticletitle{Denoising Diffusion Probabilistic Models}. In
  \bibinfo{booktitle}{\emph{Advances in Neural Information Processing Systems}}
  \emph{(\bibinfo{series}{NeurIPS '20})}. \bibinfo{pages}{6840--6851}.
\newblock
\urldef\tempurl%
\url{https://proceedings.neurips.cc/paper_files/paper/2020/file/4c5bcfec8584af0d967f1ab10179ca4b-Paper.pdf}
\showURL{%
\tempurl}


\bibitem[Ho and Salimans(2021)]%
        {ho2021classifier}
\bibfield{author}{\bibinfo{person}{Jonathan Ho} {and} \bibinfo{person}{Tim
  Salimans}.} \bibinfo{year}{2021}\natexlab{}.
\newblock \showarticletitle{Classifier-Free Diffusion Guidance}. In
  \bibinfo{booktitle}{\emph{Proceedings of the NeurIPS Workshop on DGMs and
  Applications}} \emph{(\bibinfo{series}{NeurIPS '21 Workshop})}.
\newblock
\urldef\tempurl%
\url{https://openreview.net/forum?id=qw8AKxfYbI}
\showURL{%
\tempurl}


\bibitem[Ho et~al\mbox{.}(2022b)]%
        {ho2022video}
\bibfield{author}{\bibinfo{person}{Jonathan Ho}, \bibinfo{person}{Tim
  Salimans}, \bibinfo{person}{Alexey Gritsenko}, \bibinfo{person}{William
  Chan}, \bibinfo{person}{Mohammad Norouzi}, {and} \bibinfo{person}{David~J.
  Fleet}.} \bibinfo{year}{2022}\natexlab{b}.
\newblock \showarticletitle{Video Diffusion Models}. In
  \bibinfo{booktitle}{\emph{Advances in Neural Information Processing Systems}}
  \emph{(\bibinfo{series}{NeurIPS '22})}. \bibinfo{pages}{8633--8646}.
\newblock
\urldef\tempurl%
\url{https://proceedings.neurips.cc/paper_files/paper/2022/file/39235c56aef13fb05a6adc95eb9d8d66-Paper-Conference.pdf}
\showURL{%
\tempurl}


\bibitem[Holden et~al\mbox{.}(2016)]%
        {holden2016deep}
\bibfield{author}{\bibinfo{person}{Daniel Holden}, \bibinfo{person}{Jun Saito},
  {and} \bibinfo{person}{Taku Komura}.} \bibinfo{year}{2016}\natexlab{}.
\newblock \showarticletitle{A Deep Learning Framework for Character Motion
  Synthesis and Editing}.
\newblock \bibinfo{journal}{\emph{ACM Trans. Graph.}} \bibinfo{volume}{35},
  \bibinfo{number}{4}, Article \bibinfo{articleno}{138} (\bibinfo{year}{2016}),
  \bibinfo{numpages}{11}~pages.
\newblock
\urldef\tempurl%
\url{https://doi.org/10.1145/2897824.2925975}
\showDOI{\tempurl}


\bibitem[Holmquist and Wandt(2022)]%
        {holmquist2022diffpose}
\bibfield{author}{\bibinfo{person}{Karl Holmquist} {and}
  \bibinfo{person}{Bastian Wandt}.} \bibinfo{year}{2022}\natexlab{}.
\newblock \showarticletitle{{D}iff{P}ose: Multi-Hypothesis Human Pose
  Estimation Using Diffusion Models}.
\newblock \bibinfo{journal}{\emph{arXiv preprint arXiv:2211.16487}}
  (\bibinfo{year}{2022}).
\newblock
\urldef\tempurl%
\url{https://arxiv.org/abs/2211.16487}
\showURL{%
\tempurl}


\bibitem[H{\"o}ppe et~al\mbox{.}(2022)]%
        {hoppe2022diffusion}
\bibfield{author}{\bibinfo{person}{Tobias H{\"o}ppe}, \bibinfo{person}{Arash
  Mehrjou}, \bibinfo{person}{Stefan Bauer}, \bibinfo{person}{Didrik Nielsen},
  {and} \bibinfo{person}{Andrea Dittadi}.} \bibinfo{year}{2022}\natexlab{}.
\newblock \showarticletitle{Diffusion Models for Video Prediction and
  Infilling}.
\newblock \bibinfo{journal}{\emph{Trans. Mach. Learn. Res.}}
  (\bibinfo{year}{2022}).
\newblock
\urldef\tempurl%
\url{https://openreview.net/forum?id=lf0lr4AYM6}
\showURL{%
\tempurl}


\bibitem[Hu et~al\mbox{.}(2022)]%
        {hu2022multi}
\bibfield{author}{\bibinfo{person}{Hao Hu}, \bibinfo{person}{Changhong Liu},
  \bibinfo{person}{Yong Chen}, \bibinfo{person}{Aiwen Jiang},
  \bibinfo{person}{Zhenchun Lei}, {and} \bibinfo{person}{Mingwen Wang}.}
  \bibinfo{year}{2022}\natexlab{}.
\newblock \showarticletitle{Multi-Scale Cascaded Generator for Music-Driven
  Dance Synthesis}. In \bibinfo{booktitle}{\emph{Proceedings of the
  International Joint Conference on Neural Networks}}
  \emph{(\bibinfo{series}{IJCNN '22})}. \bibinfo{publisher}{IEEE},
  \bibinfo{pages}{1--7}.
\newblock
\urldef\tempurl%
\url{https://doi.org/10.1109/IJCNN55064.2022.9892844}
\showDOI{\tempurl}


\bibitem[Hua(2018)]%
        {hua2018wavenets}
\bibfield{author}{\bibinfo{person}{Kanru Hua}.}
  \bibinfo{year}{2018}\natexlab{}.
\newblock \showarticletitle{Do {W}ave{N}ets Dream of Acoustic Waves?}
\newblock \bibinfo{journal}{\emph{arXiv preprint arXiv:1802.08370}}
  (\bibinfo{year}{2018}).
\newblock
\urldef\tempurl%
\url{https://arxiv.org/abs/1802.08370}
\showURL{%
\tempurl}


\bibitem[Hyv{\"a}rinen and Dayan(2005)]%
        {hyvarinen2005estimation}
\bibfield{author}{\bibinfo{person}{Aapo Hyv{\"a}rinen} {and}
  \bibinfo{person}{Peter Dayan}.} \bibinfo{year}{2005}\natexlab{}.
\newblock \showarticletitle{Estimation of Non-Normalized Statistical Models by
  Score Matching}.
\newblock \bibinfo{journal}{\emph{J. Mach. Learn. Res.}} \bibinfo{volume}{6},
  \bibinfo{number}{4} (\bibinfo{year}{2005}).
\newblock
\urldef\tempurl%
\url{http://jmlr.org/papers/v6/hyvarinen05a.html}
\showURL{%
\tempurl}


\bibitem[Jonell et~al\mbox{.}(2020)]%
        {jonell2020let}
\bibfield{author}{\bibinfo{person}{Patrik Jonell}, \bibinfo{person}{Taras
  Kucherenko}, \bibinfo{person}{Gustav~Eje Henter}, {and}
  \bibinfo{person}{Jonas Beskow}.} \bibinfo{year}{2020}\natexlab{}.
\newblock \showarticletitle{Let's Face It: {P}robabilistic Probabilistic
  Multi-Modal Interlocutor-Aware Generation of Facial Gestures in Dyadic
  Settings}. In \bibinfo{booktitle}{\emph{Proceedings of the ACM International
  Conference on Intelligent Virtual Agents}} \emph{(\bibinfo{series}{IVA
  '20})}. \bibinfo{publisher}{ACM}, \bibinfo{pages}{31:1--31:8}.
\newblock
\urldef\tempurl%
\url{https://doi.org/10.1145/3383652.3423911}
\showDOI{\tempurl}


\bibitem[Kim et~al\mbox{.}(2022)]%
        {kim2022flame}
\bibfield{author}{\bibinfo{person}{Jihoon Kim}, \bibinfo{person}{Jiseob Kim},
  {and} \bibinfo{person}{Sungjoon Choi}.} \bibinfo{year}{2022}\natexlab{}.
\newblock \showarticletitle{{FLAME}: Free-Form Language-Based Motion Synthesis
  \& Editing}.
\newblock \bibinfo{journal}{\emph{arXiv preprint arXiv:2209.00349}}
  (\bibinfo{year}{2022}).
\newblock
\urldef\tempurl%
\url{https://arxiv.org/abs/2209.00349}
\showURL{%
\tempurl}


\bibitem[Kingma et~al\mbox{.}(2021)]%
        {kingma2021variational}
\bibfield{author}{\bibinfo{person}{Diederik Kingma}, \bibinfo{person}{Tim
  Salimans}, \bibinfo{person}{Ben Poole}, {and} \bibinfo{person}{Jonathan Ho}.}
  \bibinfo{year}{2021}\natexlab{}.
\newblock \showarticletitle{Variational Diffusion Models}. In
  \bibinfo{booktitle}{\emph{Advances in Neural Information Processing Systems}}
  \emph{(\bibinfo{series}{NeurIPS '21})}. \bibinfo{pages}{21696--21707}.
\newblock
\urldef\tempurl%
\url{https://proceedings.neurips.cc/paper_files/paper/2021/file/b578f2a52a0229873fefc2a4b06377fa-Paper.pdf}
\showURL{%
\tempurl}


\bibitem[Kingma and Ba(2015)]%
        {kingma2015adam}
\bibfield{author}{\bibinfo{person}{Diederik~P. Kingma} {and}
  \bibinfo{person}{Jimmy Ba}.} \bibinfo{year}{2015}\natexlab{}.
\newblock \showarticletitle{Adam: A Method for Stochastic Optimization}. In
  \bibinfo{booktitle}{\emph{Proceedings of the International Conference on
  Learning Representations}} \emph{(\bibinfo{series}{ICLR '15})}.
  \bibinfo{numpages}{15}~pages.
\newblock
\urldef\tempurl%
\url{http://arxiv.org/abs/1412.6980}
\showURL{%
\tempurl}


\bibitem[Kong et~al\mbox{.}(2021)]%
        {kong2021diffwave}
\bibfield{author}{\bibinfo{person}{Zhifeng Kong}, \bibinfo{person}{Wei Ping},
  \bibinfo{person}{Jiaji Huang}, \bibinfo{person}{Kexin Zhao}, {and}
  \bibinfo{person}{Bryan Catanzaro}.} \bibinfo{year}{2021}\natexlab{}.
\newblock \showarticletitle{{D}iff{W}ave: A Versatile Diffusion Model for Audio
  Synthesis}. In \bibinfo{booktitle}{\emph{Proceedings of the International
  Conference on Learning Representations}} \emph{(\bibinfo{series}{ICLR '21})}.
\newblock
\urldef\tempurl%
\url{https://openreview.net/forum?id=a-xFK8Ymz5J}
\showURL{%
\tempurl}


\bibitem[Kopp et~al\mbox{.}(2003)]%
        {kopp2003max}
\bibfield{author}{\bibinfo{person}{Stefan Kopp}, \bibinfo{person}{Bernhard
  Jung}, \bibinfo{person}{Nadine Le{\ss}mann}, {and} \bibinfo{person}{Ipke
  Wachsmuth}.} \bibinfo{year}{2003}\natexlab{}.
\newblock \showarticletitle{{M}ax -- A Multimodal Assistant in Virtual Reality
  Construction}.
\newblock \bibinfo{journal}{\emph{KI -- K{\"u}nstliche Intelligenz}}
  \bibinfo{volume}{4}, \bibinfo{number}{03} (\bibinfo{year}{2003}),
  \bibinfo{pages}{11--17}.
\newblock
\urldef\tempurl%
\url{https://www.techfak.uni-bielefeld.de/~skopp/download/Max-KI_ECA.pdf}
\showURL{%
\tempurl}


\bibitem[Kucherenko et~al\mbox{.}(2019)]%
        {kucherenko2019analyzing}
\bibfield{author}{\bibinfo{person}{Taras Kucherenko}, \bibinfo{person}{Dai
  Hasegawa}, \bibinfo{person}{Gustav~Eje Henter}, \bibinfo{person}{Naoshi
  Kaneko}, {and} \bibinfo{person}{Hedvig Kjellstr{\"o}m}.}
  \bibinfo{year}{2019}\natexlab{}.
\newblock \showarticletitle{Analyzing Input and Output Representations for
  Speech-Driven Gesture Generation}. In \bibinfo{booktitle}{\emph{Proceedings
  of the ACM International Conference on Intelligent Virtual Agents}}
  \emph{(\bibinfo{series}{IVA '19})}. \bibinfo{publisher}{ACM},
  \bibinfo{pages}{97--104}.
\newblock
\urldef\tempurl%
\url{https://doi.org/10.1145/3308532.3329472}
\showDOI{\tempurl}


\bibitem[Kucherenko et~al\mbox{.}(2021a)]%
        {kucherenko2021moving}
\bibfield{author}{\bibinfo{person}{Taras Kucherenko}, \bibinfo{person}{Dai
  Hasegawa}, \bibinfo{person}{Naoshi Kaneko}, \bibinfo{person}{Gustav~Eje
  Henter}, {and} \bibinfo{person}{Hedvig Kjellstr{\"o}m}.}
  \bibinfo{year}{2021}\natexlab{a}.
\newblock \showarticletitle{Moving Fast and Slow: {A}nalysis of Representations
  and Post-Processing in Speech-Driven Automatic Gesture Generation}.
\newblock \bibinfo{journal}{\emph{Int. J. Hum.-Comput. Int.}}
  \bibinfo{volume}{37}, \bibinfo{number}{14} (\bibinfo{year}{2021}),
  \bibinfo{pages}{1300--1316}.
\newblock
\urldef\tempurl%
\url{https://doi.org/10.1080/10447318.2021.1883883}
\showDOI{\tempurl}


\bibitem[Kucherenko et~al\mbox{.}(2020)]%
        {kucherenko2020gesticulator}
\bibfield{author}{\bibinfo{person}{Taras Kucherenko}, \bibinfo{person}{Patrik
  Jonell}, \bibinfo{person}{Sanne van Waveren}, \bibinfo{person}{Gustav~Eje
  Henter}, \bibinfo{person}{Simon Alexanderson}, \bibinfo{person}{Iolanda
  Leite}, {and} \bibinfo{person}{Hedvig Kjellstr{\"o}m}.}
  \bibinfo{year}{2020}\natexlab{}.
\newblock \showarticletitle{Gesticulator: A Framework for Semantically-Aware
  Speech-Driven Gesture Generation}. In \bibinfo{booktitle}{\emph{Proceedings
  of the ACM International Conference on Multimodal Interaction}}
  \emph{(\bibinfo{series}{ICMI '20})}. \bibinfo{publisher}{ACM},
  \bibinfo{pages}{242--250}.
\newblock
\urldef\tempurl%
\url{https://doi.org/10.1145/3382507.3418815}
\showDOI{\tempurl}


\bibitem[Kucherenko et~al\mbox{.}(2021b)]%
        {kucherenko2021large}
\bibfield{author}{\bibinfo{person}{Taras Kucherenko}, \bibinfo{person}{Patrik
  Jonell}, \bibinfo{person}{Youngwoo Yoon}, \bibinfo{person}{Pieter Wolfert},
  {and} \bibinfo{person}{Gustav~Eje Henter}.} \bibinfo{year}{2021}\natexlab{b}.
\newblock \showarticletitle{A Large, Crowdsourced Evaluation of Gesture
  Generation Systems on Common Data: {T}he {GENEA} {C}hallenge 2020}. In
  \bibinfo{booktitle}{\emph{Proceedings of the Annual Conference on Intelligent
  User Interfaces}} \emph{(\bibinfo{series}{IUI '21})}.
  \bibinfo{publisher}{ACM}, \bibinfo{pages}{11--21}.
\newblock
\urldef\tempurl%
\url{https://doi.org/10.1145/3397481.3450692}
\showDOI{\tempurl}


\bibitem[Kucherenko et~al\mbox{.}(2022)]%
        {kucherenko2022multimodal}
\bibfield{author}{\bibinfo{person}{Taras Kucherenko}, \bibinfo{person}{Rajmund
  Nagy}, \bibinfo{person}{Michael Neff}, \bibinfo{person}{Hedvig
  Kjellstr{\"o}m}, {and} \bibinfo{person}{Gustav~Eje Henter}.}
  \bibinfo{year}{2022}\natexlab{}.
\newblock \showarticletitle{Multimodal Analysis of the Predictability of
  Hand-Gesture Properties}. In \bibinfo{booktitle}{\emph{Proceedings of the
  International Conference on Autonomous Agents and Multiagent Systems}}
  \emph{(\bibinfo{series}{AAMAS '22})}. \bibinfo{publisher}{IFAAMAS},
  \bibinfo{pages}{770--779}.
\newblock
\urldef\tempurl%
\url{https://ifaamas.org/Proceedings/aamas2022/pdfs/p770.pdf}
\showURL{%
\tempurl}


\bibitem[Kucherenko et~al\mbox{.}(2023)]%
        {kucherenko2023evaluating}
\bibfield{author}{\bibinfo{person}{Taras Kucherenko}, \bibinfo{person}{Pieter
  Wolfert}, \bibinfo{person}{Youngwoo Yoon}, \bibinfo{person}{Carla Viegas},
  \bibinfo{person}{Teodor Nikolov}, \bibinfo{person}{Mihail Tsakov}, {and}
  \bibinfo{person}{Gustav~Eje Henter}.} \bibinfo{year}{2023}\natexlab{}.
\newblock \showarticletitle{Evaluating Gesture-Generation in a Large-Scale Open
  Challenge: The {GENEA} {C}hallenge 2022}.
\newblock \bibinfo{journal}{\emph{arXiv preprint arXiv:2303.08737}}
  (\bibinfo{year}{2023}).
\newblock
\urldef\tempurl%
\url{https://arxiv.org/abs/2303.08737}
\showURL{%
\tempurl}


\bibitem[Lam et~al\mbox{.}(2022)]%
        {lam2022bddm}
\bibfield{author}{\bibinfo{person}{Max W.~Y. Lam}, \bibinfo{person}{Jun Wang},
  \bibinfo{person}{Dan Su}, {and} \bibinfo{person}{Dong Yu}.}
  \bibinfo{year}{2022}\natexlab{}.
\newblock \showarticletitle{{BDDM}: Bilateral Denoising Diffusion Models for
  Fast and High-Quality Speech Synthesis}. In
  \bibinfo{booktitle}{\emph{Proceedings of the International Conference on
  Learning Representations}} \emph{(\bibinfo{series}{ICLR '22})}.
\newblock
\urldef\tempurl%
\url{https://openreview.net/forum?id=L7wzpQttNO}
\showURL{%
\tempurl}


\bibitem[Lee et~al\mbox{.}(2019a)]%
        {lee2019talking}
\bibfield{author}{\bibinfo{person}{Gilwoo Lee}, \bibinfo{person}{Zhiwei Deng},
  \bibinfo{person}{Shugao Ma}, \bibinfo{person}{Takaaki Shiratori},
  \bibinfo{person}{Siddhartha Srinivasa}, {and} \bibinfo{person}{Yaser
  Sheikh}.} \bibinfo{year}{2019}\natexlab{a}.
\newblock \showarticletitle{{T}alking {W}ith {H}ands 16.2{M}: A Large-Scale
  Dataset of Synchronized Body-Finger Motion and Audio for Conversational
  Motion Analysis and Synthesis}. In \bibinfo{booktitle}{\emph{Proceedings of
  the IEEE/CVF International Conference on Computer Vision}}
  \emph{(\bibinfo{series}{ICCV '19})}. \bibinfo{pages}{763--772}.
\newblock
\urldef\tempurl%
\url{https://doi.org/10.1109/ICCV.2019.00085}
\showDOI{\tempurl}


\bibitem[Lee et~al\mbox{.}(2019b)]%
        {lee2019dancing}
\bibfield{author}{\bibinfo{person}{Hsin-Ying Lee}, \bibinfo{person}{Xiaodong
  Yang}, \bibinfo{person}{Ming-Yu Liu}, \bibinfo{person}{Ting-Chun Wang},
  \bibinfo{person}{Yu-Ding Lu}, \bibinfo{person}{Ming-Hsuan Yang}, {and}
  \bibinfo{person}{Jan Kautz}.} \bibinfo{year}{2019}\natexlab{b}.
\newblock \showarticletitle{Dancing to Music}.
\newblock \bibinfo{journal}{\emph{Advances in Neural Information Processing
  Systems}}  \bibinfo{volume}{32} (\bibinfo{year}{2019}).
\newblock
\urldef\tempurl%
\url{https://proceedings.neurips.cc/paper_files/paper/2019/file/7ca57a9f85a19a6e4b9a248c1daca185-Paper.pdf}
\showURL{%
\tempurl}


\bibitem[Lee and Marsella(2006)]%
        {lee2006nonverbal}
\bibfield{author}{\bibinfo{person}{Jina Lee} {and} \bibinfo{person}{Stacy
  Marsella}.} \bibinfo{year}{2006}\natexlab{}.
\newblock \showarticletitle{Nonverbal Behavior Generator for Embodied
  Conversational Agents}. In \bibinfo{booktitle}{\emph{Proceedings of the
  International Conference on Intelligent Virtual Agents}}
  \emph{(\bibinfo{series}{IVA '06})}. \bibinfo{publisher}{Springer},
  \bibinfo{pages}{243--255}.
\newblock
\urldef\tempurl%
\url{https://doi.org/10.1007/11821830_20}
\showDOI{\tempurl}


\bibitem[Levine et~al\mbox{.}(2010)]%
        {levine2010gesture}
\bibfield{author}{\bibinfo{person}{Sergey Levine}, \bibinfo{person}{Philipp
  Kr{\"a}henb{\"u}hl}, \bibinfo{person}{Sebastian Thrun}, {and}
  \bibinfo{person}{Vladlen Koltun}.} \bibinfo{year}{2010}\natexlab{}.
\newblock \showarticletitle{Gesture Controllers}.
\newblock \bibinfo{journal}{\emph{ACM Trans. Graph.}} \bibinfo{volume}{29},
  \bibinfo{number}{4}, Article \bibinfo{articleno}{124} (\bibinfo{year}{2010}),
  \bibinfo{numpages}{11}~pages.
\newblock
\urldef\tempurl%
\url{https://doi.org/10.1145/1778765.1778861}
\showDOI{\tempurl}


\bibitem[Lhommet et~al\mbox{.}(2015)]%
        {lhommet2015cerebella}
\bibfield{author}{\bibinfo{person}{Margot Lhommet}, \bibinfo{person}{Yuyu Xu},
  {and} \bibinfo{person}{Stacy Marsella}.} \bibinfo{year}{2015}\natexlab{}.
\newblock \showarticletitle{Cerebella: Automatic Generation of Nonverbal
  Behavior for Virtual Humans}. In \bibinfo{booktitle}{\emph{Proceedings of the
  AAAI Conference on Artificial Intelligence}} \emph{(\bibinfo{series}{AAAI
  '15}, \bibinfo{number}{1})}.
\newblock
\urldef\tempurl%
\url{https://doi.org/10.1609/aaai.v29i1.9778}
\showDOI{\tempurl}


\bibitem[Li et~al\mbox{.}(2022)]%
        {li2022danceformer}
\bibfield{author}{\bibinfo{person}{Buyu Li}, \bibinfo{person}{Yongchi Zhao},
  \bibinfo{person}{Shi Zhelun}, {and} \bibinfo{person}{Lu Sheng}.}
  \bibinfo{year}{2022}\natexlab{}.
\newblock \showarticletitle{{D}ance{F}ormer: Music Conditioned 3{D} Dance
  Generation with Parametric Motion {T}ransformer}. In
  \bibinfo{booktitle}{\emph{Proceedings of the AAAI Conference on Artificial
  Intelligence}} \emph{(\bibinfo{series}{AAAI '22},
  Vol.~\bibinfo{volume}{36})}. \bibinfo{pages}{1272--1279}.
\newblock
\urldef\tempurl%
\url{https://doi.org/10.1609/aaai.v36i2.20014}
\showDOI{\tempurl}


\bibitem[Li et~al\mbox{.}(2020)]%
        {li2020learning}
\bibfield{author}{\bibinfo{person}{Jiaman Li}, \bibinfo{person}{Yihang Yin},
  \bibinfo{person}{Hang Chu}, \bibinfo{person}{Yi Zhou},
  \bibinfo{person}{Tingwu Wang}, \bibinfo{person}{Sanja Fidler}, {and}
  \bibinfo{person}{Hao Li}.} \bibinfo{year}{2020}\natexlab{}.
\newblock \showarticletitle{Learning to Generate Diverse Dance Motions with
  {T}ransformer}.
\newblock \bibinfo{journal}{\emph{arXiv preprint arXiv:2008.08171}}
  (\bibinfo{year}{2020}).
\newblock
\urldef\tempurl%
\url{https://arxiv.org/abs/2008.08171}
\showURL{%
\tempurl}


\bibitem[Li et~al\mbox{.}(2021)]%
        {li2021ai}
\bibfield{author}{\bibinfo{person}{Ruilong Li}, \bibinfo{person}{Shan Yang},
  \bibinfo{person}{David~A. Ross}, {and} \bibinfo{person}{Angjoo Kanazawa}.}
  \bibinfo{year}{2021}\natexlab{}.
\newblock \showarticletitle{{AI} {C}horeographer: Music Conditioned 3{D} Dance
  Generation with {AIST}++}. In \bibinfo{booktitle}{\emph{Proceedings of the
  IEEE/CVF International Conference on Computer Vision}}
  \emph{(\bibinfo{series}{ICCV '21})}. \bibinfo{pages}{13401--13412}.
\newblock
\urldef\tempurl%
\url{https://doi.org/10.1109/ICCV48922.2021.01315}
\showDOI{\tempurl}


\bibitem[Ling et~al\mbox{.}(2020)]%
        {ling2020character}
\bibfield{author}{\bibinfo{person}{Hung~Yu Ling}, \bibinfo{person}{Fabio
  Zinno}, \bibinfo{person}{George Cheng}, {and} \bibinfo{person}{Michiel van~de
  Panne}.} \bibinfo{year}{2020}\natexlab{}.
\newblock \showarticletitle{Character Controllers Using Motion {VAE}s}.
\newblock \bibinfo{journal}{\emph{ACM Trans. Graph.}} \bibinfo{volume}{39},
  \bibinfo{number}{4}, Article \bibinfo{articleno}{40} (\bibinfo{year}{2020}),
  \bibinfo{numpages}{12}~pages.
\newblock
\urldef\tempurl%
\url{https://doi.org/10.1145/3386569.3392422}
\showDOI{\tempurl}


\bibitem[Liu et~al\mbox{.}(2022b)]%
        {liu2022beat}
\bibfield{author}{\bibinfo{person}{Haiyang Liu}, \bibinfo{person}{Zihao Zhu},
  \bibinfo{person}{Naoya Iwamoto}, \bibinfo{person}{Yichen Peng},
  \bibinfo{person}{Zhengqing Li}, \bibinfo{person}{You Zhou},
  \bibinfo{person}{Elif Bozkurt}, {and} \bibinfo{person}{Bo Zheng}.}
  \bibinfo{year}{2022}\natexlab{b}.
\newblock \showarticletitle{{BEAT}: A Large-Scale Semantic and Emotional
  Multi-Modal Dataset for Conversational Gestures Synthesis}. In
  \bibinfo{booktitle}{\emph{Proceedings of the European Conference on Computer
  Vision}} \emph{(\bibinfo{series}{ECCV '22})}. \bibinfo{publisher}{Springer},
  \bibinfo{pages}{612--630}.
\newblock
\urldef\tempurl%
\url{https://doi.org/10.1007/978-3-031-20071-7_36}
\showDOI{\tempurl}


\bibitem[Liu et~al\mbox{.}(2022a)]%
        {liu2022compositional}
\bibfield{author}{\bibinfo{person}{Nan Liu}, \bibinfo{person}{Shuang Li},
  \bibinfo{person}{Yilun Du}, \bibinfo{person}{Antonio Torralba}, {and}
  \bibinfo{person}{Joshua~B. Tenenbaum}.} \bibinfo{year}{2022}\natexlab{a}.
\newblock \showarticletitle{Compositional Visual Generation with Composable
  Diffusion Models}. In \bibinfo{booktitle}{\emph{Proceedings of the European
  Conference on Computer Vision}} \emph{(\bibinfo{series}{ECCV '22})}.
  \bibinfo{publisher}{Springer}, \bibinfo{pages}{423--439}.
\newblock
\urldef\tempurl%
\url{https://doi.org/10.1007/978-3-031-19790-1_26}
\showDOI{\tempurl}


\bibitem[Liu et~al\mbox{.}(2021)]%
        {liu2021speech}
\bibfield{author}{\bibinfo{person}{Yu Liu}, \bibinfo{person}{Gelareh
  Mohammadi}, \bibinfo{person}{Yang Song}, {and} \bibinfo{person}{Wafa Johal}.}
  \bibinfo{year}{2021}\natexlab{}.
\newblock \showarticletitle{Speech-Based Gesture Generation for Robots and
  Embodied Agents: A Scoping Review}. In \bibinfo{booktitle}{\emph{Proceedings
  of the International Conference on Human Agent Interaction}}
  \emph{(\bibinfo{series}{HAI '21})}. \bibinfo{publisher}{ACM},
  \bibinfo{pages}{31--38}.
\newblock
\urldef\tempurl%
\url{https://doi.org/10.1145/3472307.3484167}
\showDOI{\tempurl}


\bibitem[Ma et~al\mbox{.}(2022)]%
        {ma2022pretrained}
\bibfield{author}{\bibinfo{person}{Jianxin Ma}, \bibinfo{person}{Shuai Bai},
  {and} \bibinfo{person}{Chang Zhou}.} \bibinfo{year}{2022}\natexlab{}.
\newblock \showarticletitle{Pretrained Diffusion Models for Unified Human
  Motion Synthesis}.
\newblock \bibinfo{journal}{\emph{arXiv preprint arXiv:2212.02837}}
  (\bibinfo{year}{2022}).
\newblock
\urldef\tempurl%
\url{https://arxiv.org/abs/2212.02837}
\showURL{%
\tempurl}


\bibitem[MacKay(2003)]%
        {mackay2003information}
\bibfield{author}{\bibinfo{person}{David J.~C. MacKay}.}
  \bibinfo{year}{2003}\natexlab{}.
\newblock \bibinfo{booktitle}{\emph{Information Theory, Inference and Learning
  Algorithms}}.
\newblock \bibinfo{publisher}{Cambridge University Press}.
\newblock
\urldef\tempurl%
\url{https://www.cambridge.org/0521642981}
\showURL{%
\tempurl}


\bibitem[Mason et~al\mbox{.}(2022)]%
        {mason2022realtime}
\bibfield{author}{\bibinfo{person}{Ian Mason}, \bibinfo{person}{Sebastian
  Starke}, {and} \bibinfo{person}{Taku Komura}.}
  \bibinfo{year}{2022}\natexlab{}.
\newblock \showarticletitle{Real-Time Style Modelling of Human Locomotion via
  Feature-Wise Transformations and Local Motion Phases}.
\newblock \bibinfo{journal}{\emph{Proc. ACM Comput. Graph. Interact. Tech.}}
  \bibinfo{volume}{5}, \bibinfo{number}{1}, Article \bibinfo{articleno}{6}
  (\bibinfo{year}{2022}), \bibinfo{numpages}{18}~pages.
\newblock
\urldef\tempurl%
\url{https://doi.org/10.1145/3522618}
\showDOI{\tempurl}


\bibitem[Meng et~al\mbox{.}(2022)]%
        {meng2022distillation}
\bibfield{author}{\bibinfo{person}{Chenlin Meng}, \bibinfo{person}{Ruiqi Gao},
  \bibinfo{person}{Diederik~P. Kingma}, \bibinfo{person}{Stefano Ermon},
  \bibinfo{person}{Jonathan Ho}, {and} \bibinfo{person}{Tim Salimans}.}
  \bibinfo{year}{2022}\natexlab{}.
\newblock \showarticletitle{On Distillation of Guided Diffusion Models}. In
  \bibinfo{booktitle}{\emph{Proceedings of the NeurIPS Workshop on Score-Based
  Methods}} \emph{(\bibinfo{series}{NeurIPS '22 Workshop})}.
\newblock
\urldef\tempurl%
\url{https://openreview.net/forum?id=6QHpSQt6VR-}
\showURL{%
\tempurl}


\bibitem[Miller et~al\mbox{.}(2013)]%
        {miller2013you}
\bibfield{author}{\bibinfo{person}{Jared~E. Miller}, \bibinfo{person}{Laura~A.
  Carlson}, {and} \bibinfo{person}{J.~Devin McAuley}.}
  \bibinfo{year}{2013}\natexlab{}.
\newblock \showarticletitle{When What You Hear Influences When You See:
  Listening to an Auditory Rhythm Influences the Temporal Allocation of Visual
  Attention}.
\newblock \bibinfo{journal}{\emph{Psychol. Sci.}} \bibinfo{volume}{24},
  \bibinfo{number}{1} (\bibinfo{year}{2013}), \bibinfo{pages}{11--18}.
\newblock
\urldef\tempurl%
\url{https://doi.org/10.1177/0956797612446707}
\showDOI{\tempurl}


\bibitem[M{\"u}ller et~al\mbox{.}(2005)]%
        {muller2005chroma}
\bibfield{author}{\bibinfo{person}{Meinard M{\"u}ller}, \bibinfo{person}{Frank
  Kurth}, {and} \bibinfo{person}{Michael Clausen}.}
  \bibinfo{year}{2005}\natexlab{}.
\newblock \showarticletitle{Chroma-Based Statistical Audio Features for Audio
  Matching}. In \bibinfo{booktitle}{\emph{Proceedings of the IEEE Workshop on
  Applications of Signal Processing to Audio and Acoustics}}
  \emph{(\bibinfo{series}{WASPAA '05})}. \bibinfo{publisher}{IEEE},
  \bibinfo{pages}{275--278}.
\newblock
\urldef\tempurl%
\url{https://doi.org/10.1109/ASPAA.2005.1540223}
\showDOI{\tempurl}


\bibitem[M\"{u}ller et~al\mbox{.}(2005)]%
        {Mueller2005Efficientcontentbased}
\bibfield{author}{\bibinfo{person}{Meinard M\"{u}ller}, \bibinfo{person}{Tido
  R\"{o}der}, {and} \bibinfo{person}{Michael Clausen}.}
  \bibinfo{year}{2005}\natexlab{}.
\newblock \showarticletitle{Efficient Content-Based Retrieval of Motion Capture
  Data}.
\newblock \bibinfo{journal}{\emph{ACM Trans. Graph.}} \bibinfo{volume}{24},
  \bibinfo{number}{3} (\bibinfo{year}{2005}), \bibinfo{pages}{677--685}.
\newblock
\urldef\tempurl%
\url{https://doi.org/10.1145/1073204.1073247}
\showDOI{\tempurl}


\bibitem[Neff et~al\mbox{.}(2010)]%
        {neff2010evaluating}
\bibfield{author}{\bibinfo{person}{Michael Neff}, \bibinfo{person}{Yingying
  Wang}, \bibinfo{person}{Rob Abbott}, {and} \bibinfo{person}{Marilyn Walker}.}
  \bibinfo{year}{2010}\natexlab{}.
\newblock \showarticletitle{Evaluating the Effect of Gesture and Language on
  Personality Perception in Conversational Agents}. In
  \bibinfo{booktitle}{\emph{Proceedings of the International Conference on
  Intelligent Virtual Agents}} \emph{(\bibinfo{series}{IVA '10})}.
  \bibinfo{publisher}{Springer}, \bibinfo{pages}{222--235}.
\newblock
\urldef\tempurl%
\url{https://doi.org/10.1007/978-3-642-15892-6_24}
\showDOI{\tempurl}


\bibitem[Nichol and Dhariwal(2021)]%
        {nichol2021improved}
\bibfield{author}{\bibinfo{person}{Alexander~Quinn Nichol} {and}
  \bibinfo{person}{Prafulla Dhariwal}.} \bibinfo{year}{2021}\natexlab{}.
\newblock \showarticletitle{Improved Denoising Diffusion Probabilistic Models}.
  In \bibinfo{booktitle}{\emph{Proceedings of the International Conference on
  Machine Learning}} \emph{(\bibinfo{series}{ICML '21})}.
  \bibinfo{pages}{8162--8171}.
\newblock
\urldef\tempurl%
\url{https://proceedings.mlr.press/v139/nichol21a.html}
\showURL{%
\tempurl}


\bibitem[Nichol et~al\mbox{.}(2022)]%
        {nichol2021glide}
\bibfield{author}{\bibinfo{person}{Alexander~Quinn Nichol},
  \bibinfo{person}{Prafulla Dhariwal}, \bibinfo{person}{Aditya Ramesh},
  \bibinfo{person}{Pranav Shyam}, \bibinfo{person}{Pamela Mishkin},
  \bibinfo{person}{Bob Mcgrew}, \bibinfo{person}{Ilya Sutskever}, {and}
  \bibinfo{person}{Mark Chen}.} \bibinfo{year}{2022}\natexlab{}.
\newblock \showarticletitle{{GLIDE}: Towards Photorealistic Image Generation
  and Editing with Text-Guided Diffusion Models}. In
  \bibinfo{booktitle}{\emph{Proceedings of the International Conference on
  Machine Learning}} \emph{(\bibinfo{series}{ICML -22})}.
  \bibinfo{pages}{16784--16804}.
\newblock
\urldef\tempurl%
\url{https://proceedings.mlr.press/v162/nichol22a.html}
\showURL{%
\tempurl}


\bibitem[Normoyle et~al\mbox{.}(2013)]%
        {normoyle2013effect}
\bibfield{author}{\bibinfo{person}{Aline Normoyle}, \bibinfo{person}{Fannie
  Liu}, \bibinfo{person}{Mubbasir Kapadia}, \bibinfo{person}{Norman~I. Badler},
  {and} \bibinfo{person}{Sophie J{\"o}rg}.} \bibinfo{year}{2013}\natexlab{}.
\newblock \showarticletitle{The Effect of Posture and Dynamics on the
  Perception of Emotion}. In \bibinfo{booktitle}{\emph{Proceedings of the ACM
  Symposium on Applied Perception}} \emph{(\bibinfo{series}{SAP '13})}.
  \bibinfo{publisher}{ACM}, \bibinfo{pages}{91--98}.
\newblock
\urldef\tempurl%
\url{https://doi.org/10.1145/2492494.2492500}
\showDOI{\tempurl}


\bibitem[Novack and Goldin-Meadow(2015)]%
        {novack2015learning}
\bibfield{author}{\bibinfo{person}{Miriam Novack} {and} \bibinfo{person}{Susan
  Goldin-Meadow}.} \bibinfo{year}{2015}\natexlab{}.
\newblock \showarticletitle{Learning from Gesture: How Our Hands Change Our
  Minds}.
\newblock \bibinfo{journal}{\emph{Educ. Psychol. Rev.}} \bibinfo{volume}{27},
  \bibinfo{number}{3} (\bibinfo{year}{2015}), \bibinfo{pages}{405--412}.
\newblock
\urldef\tempurl%
\url{https://doi.org/10.1007/s10648-015-9325-3}
\showDOI{\tempurl}


\bibitem[Nyatsanga et~al\mbox{.}(2023)]%
        {nyatsanga2023comprehensive}
\bibfield{author}{\bibinfo{person}{Simbarashe Nyatsanga},
  \bibinfo{person}{Taras Kucherenko}, \bibinfo{person}{Chaitanya Ahuja},
  \bibinfo{person}{Gustav~Eje Henter}, {and} \bibinfo{person}{Michael Neff}.}
  \bibinfo{year}{2023}\natexlab{}.
\newblock \showarticletitle{A Comprehensive Review of Data-Driven Co-Speech
  Gesture Generation}.
\newblock \bibinfo{journal}{\emph{Comput. Graph. Forum}}
  (\bibinfo{year}{2023}).
\newblock
\urldef\tempurl%
\url{https://arxiv.org/abs/2301.05339}
\showURL{%
\tempurl}


\bibitem[Ofli et~al\mbox{.}(2011)]%
        {ofli2011learn2dance}
\bibfield{author}{\bibinfo{person}{Ferda Ofli}, \bibinfo{person}{Engin Erzin},
  \bibinfo{person}{Y{\"u}cel Yemez}, {and} \bibinfo{person}{A.~Murat Tekalp}.}
  \bibinfo{year}{2011}\natexlab{}.
\newblock \showarticletitle{{L}earn2{D}ance: Learning Statistical
  Music-to-Dance Mappings for Choreography Synthesis}.
\newblock \bibinfo{journal}{\emph{IEEE T. Multimedia}} \bibinfo{volume}{14},
  \bibinfo{number}{3} (\bibinfo{year}{2011}), \bibinfo{pages}{747--759}.
\newblock
\urldef\tempurl%
\url{https://doi.org/10.1109/TMM.2011.2181492}
\showDOI{\tempurl}


\bibitem[Onuma et~al\mbox{.}(2008)]%
        {onuma2008fmdistance}
\bibfield{author}{\bibinfo{person}{Kensuke Onuma}, \bibinfo{person}{Christos
  Faloutsos}, {and} \bibinfo{person}{Jessica~K. Hodgins}.}
  \bibinfo{year}{2008}\natexlab{}.
\newblock \showarticletitle{{FMDistance}: A Fast and Effective Distance
  Function for Motion Capture Data}. In \bibinfo{booktitle}{\emph{Proceedings
  of the Annual Conference of the European Association for Computer Graphics --
  Short Papers}} \emph{(\bibinfo{series}{EUROGRAPHICS '08})},
  \bibfield{editor}{\bibinfo{person}{Katerina Mania} {and}
  \bibinfo{person}{Eric Reinhard}} (Eds.). \bibinfo{publisher}{The Eurographics
  Association}.
\newblock
\urldef\tempurl%
\url{https://doi.org/10.2312/egs.20081027}
\showDOI{\tempurl}


\bibitem[Papillon et~al\mbox{.}(2023)]%
        {papillon2022pirounet}
\bibfield{author}{\bibinfo{person}{Mathilde Papillon}, \bibinfo{person}{Mariel
  Pettee}, {and} \bibinfo{person}{Nina Miolane}.}
  \bibinfo{year}{2023}\natexlab{}.
\newblock \showarticletitle{{P}irou{N}et: Creating Dance Through Artist-Centric
  Deep Learning}. In \bibinfo{booktitle}{\emph{Proceedings of the EAI
  International Conference ArtsIT, Interactivity and Game Creation}}
  \emph{(\bibinfo{series}{ArtsIT '23})}. \bibinfo{publisher}{Springer Nature},
  \bibinfo{pages}{447--465}.
\newblock
\urldef\tempurl%
\url{https://doi.org/10.1007/978-3-031-28993-4_31}
\showDOI{\tempurl}


\bibitem[Parizet et~al\mbox{.}(2005)]%
        {parizet2005comparison}
\bibfield{author}{\bibinfo{person}{Etienne Parizet}, \bibinfo{person}{Nacer
  Hamzaoui}, {and} \bibinfo{person}{Guillaume Sabati{\'e}}.}
  \bibinfo{year}{2005}\natexlab{}.
\newblock \showarticletitle{Comparison of Some Listening Test Methods: A Case
  Study}.
\newblock \bibinfo{journal}{\emph{Acta Acust. United Ac.}}
  \bibinfo{volume}{91}, \bibinfo{number}{2} (\bibinfo{year}{2005}),
  \bibinfo{pages}{356--364}.
\newblock
\urldef\tempurl%
\url{https://www.ingentaconnect.com/content/dav/aaua/2005/00000091/00000002/art00018}
\showURL{%
\tempurl}


\bibitem[Perez et~al\mbox{.}(2018)]%
        {perez2018film}
\bibfield{author}{\bibinfo{person}{Ethan Perez}, \bibinfo{person}{Florian
  Strub}, \bibinfo{person}{Harm De~Vries}, \bibinfo{person}{Vincent Dumoulin},
  {and} \bibinfo{person}{Aaron Courville}.} \bibinfo{year}{2018}\natexlab{}.
\newblock \showarticletitle{{FiLM}: Visual Reasoning with a General
  Conditioning Layer}. In \bibinfo{booktitle}{\emph{Proceedings of the AAAI
  Conference on Artificial Intelligence}} \emph{(\bibinfo{series}{AAAI '18},
  \bibinfo{number}{1})}.
\newblock
\urldef\tempurl%
\url{https://doi.org/10.1609/aaai.v32i1.11671}
\showDOI{\tempurl}


\bibitem[Petrovich et~al\mbox{.}(2022)]%
        {petrovich2022temos}
\bibfield{author}{\bibinfo{person}{Mathis Petrovich},
  \bibinfo{person}{Michael~J. Black}, {and} \bibinfo{person}{G{\"u}l Varol}.}
  \bibinfo{year}{2022}\natexlab{}.
\newblock \showarticletitle{{TEMOS}: Generating Diverse Human Motions from
  Textual Descriptions}. In \bibinfo{booktitle}{\emph{Proceedings of the
  European Conference on Computer Vision}} \emph{(\bibinfo{series}{ECCV '22})}.
  \bibinfo{publisher}{Springer}, \bibinfo{pages}{480--497}.
\newblock
\urldef\tempurl%
\url{https://doi.org/10.1007/978-3-031-20047-2_28}
\showDOI{\tempurl}


\bibitem[Pouw et~al\mbox{.}(2021)]%
        {pouw2021multilevel}
\bibfield{author}{\bibinfo{person}{Wim Pouw}, \bibinfo{person}{Shannon
  Proksch}, \bibinfo{person}{Linda Drijvers}, \bibinfo{person}{Marco Gamba},
  \bibinfo{person}{Judith Holler}, \bibinfo{person}{Christopher Kello},
  \bibinfo{person}{Rebecca~S. Schaefer}, {and} \bibinfo{person}{Geraint~A.
  Wiggins}.} \bibinfo{year}{2021}\natexlab{}.
\newblock \showarticletitle{Multilevel Rhythms in Multimodal Communication}.
\newblock \bibinfo{journal}{\emph{Philos. T. R. Soc. B}} \bibinfo{volume}{376},
  \bibinfo{number}{1835} (\bibinfo{year}{2021}), \bibinfo{pages}{20200334}.
\newblock
\urldef\tempurl%
\url{https://doi.org/10.1098/rstb.2020.0334}
\showDOI{\tempurl}


\bibitem[Press et~al\mbox{.}(2022)]%
        {press2022train}
\bibfield{author}{\bibinfo{person}{Ofir Press}, \bibinfo{person}{Noah Smith},
  {and} \bibinfo{person}{Mike Lewis}.} \bibinfo{year}{2022}\natexlab{}.
\newblock \showarticletitle{Train Short, Test Long: Attention with Linear
  Biases Enables Input Length Extrapolation}. In
  \bibinfo{booktitle}{\emph{Proceedings of the International Conference on
  Learning Representations}} \emph{(\bibinfo{series}{ICLR '22})}.
\newblock
\urldef\tempurl%
\url{https://openreview.net/forum?id=R8sQPpGCv0}
\showURL{%
\tempurl}


\bibitem[Raffel et~al\mbox{.}(2020)]%
        {raffel2020exploring}
\bibfield{author}{\bibinfo{person}{Colin Raffel}, \bibinfo{person}{Noam
  Shazeer}, \bibinfo{person}{Adam Roberts}, \bibinfo{person}{Katherine Lee},
  \bibinfo{person}{Sharan Narang}, \bibinfo{person}{Michael Matena},
  \bibinfo{person}{Yanqi Zhou}, \bibinfo{person}{Wei Li}, {and}
  \bibinfo{person}{Peter~J. Liu}.} \bibinfo{year}{2020}\natexlab{}.
\newblock \showarticletitle{Exploring the Limits of Transfer Learning with a
  Unified Text-to-Text Transformer}.
\newblock \bibinfo{journal}{\emph{J. Mach. Learn. Res.}} \bibinfo{volume}{21},
  \bibinfo{number}{140} (\bibinfo{year}{2020}), \bibinfo{pages}{1--67}.
\newblock
\urldef\tempurl%
\url{http://jmlr.org/papers/v21/20-074.html}
\showURL{%
\tempurl}


\bibitem[Ramesh et~al\mbox{.}(2022)]%
        {ramesh2022hierarchical}
\bibfield{author}{\bibinfo{person}{Aditya Ramesh}, \bibinfo{person}{Prafulla
  Dhariwal}, \bibinfo{person}{Alex Nichol}, \bibinfo{person}{Casey Chu}, {and}
  \bibinfo{person}{Mark Chen}.} \bibinfo{year}{2022}\natexlab{}.
\newblock \showarticletitle{Hierarchical Text-Conditional Image Generation with
  {CLIP} Latents}.
\newblock \bibinfo{journal}{\emph{arXiv preprint arXiv:2204.06125}}
  (\bibinfo{year}{2022}).
\newblock
\urldef\tempurl%
\url{https://arxiv.org/abs/2204.06125}
\showURL{%
\tempurl}


\bibitem[Rombach et~al\mbox{.}(2022)]%
        {rombach2022high}
\bibfield{author}{\bibinfo{person}{Robin Rombach}, \bibinfo{person}{Andreas
  Blattmann}, \bibinfo{person}{Dominik Lorenz}, \bibinfo{person}{Patrick
  Esser}, {and} \bibinfo{person}{Bj{\"o}rn Ommer}.}
  \bibinfo{year}{2022}\natexlab{}.
\newblock \showarticletitle{High-Resolution Image Synthesis with Latent
  Diffusion Models}. In \bibinfo{booktitle}{\emph{Proceedings of the IEEE/CVF
  Conference on Computer Vision and Pattern Recognition}}
  \emph{(\bibinfo{series}{CVPR '22})}. \bibinfo{pages}{10684--10695}.
\newblock
\urldef\tempurl%
\url{https://doi.org/10.1109/CVPR52688.2022.01042}
\showDOI{\tempurl}


\bibitem[Sadoughi and Busso(2019)]%
        {sadoughi2019speech}
\bibfield{author}{\bibinfo{person}{Najmeh Sadoughi} {and}
  \bibinfo{person}{Carlos Busso}.} \bibinfo{year}{2019}\natexlab{}.
\newblock \showarticletitle{Speech-Driven Animation with Meaningful Behaviors}.
\newblock \bibinfo{journal}{\emph{Speech Commun.}}  \bibinfo{volume}{110}
  (\bibinfo{year}{2019}), \bibinfo{pages}{90--100}.
\newblock
\urldef\tempurl%
\url{https://doi.org/10.1016/j.specom.2019.04.005}
\showDOI{\tempurl}


\bibitem[Saharia et~al\mbox{.}(2022)]%
        {saharia2022photorealistic}
\bibfield{author}{\bibinfo{person}{Chitwan Saharia}, \bibinfo{person}{William
  Chan}, \bibinfo{person}{Saurabh Saxena}, \bibinfo{person}{Lala Li},
  \bibinfo{person}{Jay Whang}, \bibinfo{person}{Emily Denton},
  \bibinfo{person}{Seyed Kamyar~Seyed Ghasemipour},
  \bibinfo{person}{Burcu~Karagol Ayan}, \bibinfo{person}{S.~Sara Mahdavi},
  \bibinfo{person}{Rapha~Gontijo Lopes}, {et~al\mbox{.}}}
  \bibinfo{year}{2022}\natexlab{}.
\newblock \showarticletitle{Photorealistic Text-to-Image Diffusion Models with
  Deep Language Understanding}. In \bibinfo{booktitle}{\emph{Advances in Neural
  Information Processing Systems}} \emph{(\bibinfo{series}{NeurIPS '22})}.
  \bibinfo{pages}{36479--36494}.
\newblock
\urldef\tempurl%
\url{https://proceedings.neurips.cc/paper_files/paper/2022/file/ec795aeadae0b7d230fa35cbaf04c041-Paper-Conference.pdf}
\showURL{%
\tempurl}


\bibitem[Salem et~al\mbox{.}(2012)]%
        {salem2012generation}
\bibfield{author}{\bibinfo{person}{Maha Salem}, \bibinfo{person}{Stefan Kopp},
  \bibinfo{person}{Ipke Wachsmuth}, \bibinfo{person}{Katharina Rohlfing}, {and}
  \bibinfo{person}{Frank Joublin}.} \bibinfo{year}{2012}\natexlab{}.
\newblock \showarticletitle{Generation and Evaluation of Communicative Robot
  Gesture}.
\newblock \bibinfo{journal}{\emph{Int. J. Soc. Robot.}} \bibinfo{volume}{4},
  \bibinfo{number}{2} (\bibinfo{year}{2012}), \bibinfo{pages}{201--217}.
\newblock
\urldef\tempurl%
\url{https://doi.org/10.1007/s12369-011-0124-9}
\showDOI{\tempurl}


\bibitem[Salimans and Ho(2022)]%
        {salimans2022progressive}
\bibfield{author}{\bibinfo{person}{Tim Salimans} {and}
  \bibinfo{person}{Jonathan Ho}.} \bibinfo{year}{2022}\natexlab{}.
\newblock \showarticletitle{Progressive Distillation for Fast Sampling of
  Diffusion Models}. In \bibinfo{booktitle}{\emph{Proceedings of the
  International Conference on Learning Representations}}
  \emph{(\bibinfo{series}{ICLR '22})}.
\newblock
\urldef\tempurl%
\url{https://openreview.net/forum?id=TIdIXIpzhoI}
\showURL{%
\tempurl}


\bibitem[Shannon et~al\mbox{.}(2011)]%
        {shannon2011effect}
\bibfield{author}{\bibinfo{person}{Matt Shannon}, \bibinfo{person}{Heiga Zen},
  {and} \bibinfo{person}{William Byrne}.} \bibinfo{year}{2011}\natexlab{}.
\newblock \showarticletitle{The Effect of Using Normalized Models in
  Statistical Speech Synthesis}. In \bibinfo{booktitle}{\emph{Proceedings of
  the Annual Conference of the International Speech Communication Association}}
  \emph{(\bibinfo{series}{Interspeech '11})}. \bibinfo{publisher}{ISCA},
  \bibinfo{pages}{121--124}.
\newblock
\urldef\tempurl%
\url{https://doi.org/10.21437/Interspeech.2011-31}
\showDOI{\tempurl}


\bibitem[Siyao et~al\mbox{.}(2022)]%
        {siyao2022bailando}
\bibfield{author}{\bibinfo{person}{Li Siyao}, \bibinfo{person}{Weijiang Yu},
  \bibinfo{person}{Tianpei Gu}, \bibinfo{person}{Chunze Lin},
  \bibinfo{person}{Quan Wang}, \bibinfo{person}{Chen Qian},
  \bibinfo{person}{Chen~Change Loy}, {and} \bibinfo{person}{Ziwei Liu}.}
  \bibinfo{year}{2022}\natexlab{}.
\newblock \showarticletitle{Bailando: 3{D} Dance Generation by Actor-Critic
  {GPT} with Choreographic Memory}. In \bibinfo{booktitle}{\emph{Proceedings of
  the IEEE/CVF Conference on Computer Vision and Pattern Recognition}}
  \emph{(\bibinfo{series}{CVPR '22})}. \bibinfo{pages}{11050--11059}.
\newblock
\urldef\tempurl%
\url{https://doi.org/10.1109/CVPR52688.2022.01077}
\showDOI{\tempurl}


\bibitem[Smith and Neff(2017)]%
        {smith2017understanding}
\bibfield{author}{\bibinfo{person}{Harrison~Jesse Smith} {and}
  \bibinfo{person}{Michael Neff}.} \bibinfo{year}{2017}\natexlab{}.
\newblock \showarticletitle{Understanding the Impact of Animated Gesture
  Performance on Personality Perceptions}.
\newblock \bibinfo{journal}{\emph{ACM Trans. Graph.}} \bibinfo{volume}{36},
  \bibinfo{number}{4}, Article \bibinfo{articleno}{49} (\bibinfo{year}{2017}),
  \bibinfo{numpages}{12}~pages.
\newblock
\urldef\tempurl%
\url{https://doi.org/10.1145/3072959.3073697}
\showDOI{\tempurl}


\bibitem[Sohl-Dickstein et~al\mbox{.}(2015)]%
        {sohl2015deep}
\bibfield{author}{\bibinfo{person}{Jascha Sohl-Dickstein},
  \bibinfo{person}{Eric Weiss}, \bibinfo{person}{Niru Maheswaranathan}, {and}
  \bibinfo{person}{Surya Ganguli}.} \bibinfo{year}{2015}\natexlab{}.
\newblock \showarticletitle{Deep Unsupervised Learning Using Nonequilibrium
  Thermodynamics}. In \bibinfo{booktitle}{\emph{Proceedings of the
  International Conference on Machine Learning}} \emph{(\bibinfo{series}{ICML
  '15})}. \bibinfo{pages}{2256--2265}.
\newblock
\urldef\tempurl%
\url{https://proceedings.mlr.press/v37/sohl-dickstein15.html}
\showURL{%
\tempurl}


\bibitem[Song and Ermon(2019)]%
        {song2019generative}
\bibfield{author}{\bibinfo{person}{Yang Song} {and} \bibinfo{person}{Stefano
  Ermon}.} \bibinfo{year}{2019}\natexlab{}.
\newblock \showarticletitle{Generative Modeling by Estimating Gradients of the
  Data Distribution}. In \bibinfo{booktitle}{\emph{Advances in Neural
  Information Processing Systems}} \emph{(\bibinfo{series}{NeurIPS '19})}.
\newblock
\urldef\tempurl%
\url{https://proceedings.neurips.cc/paper_files/paper/2019/file/3001ef257407d5a371a96dcd947c7d93-Paper.pdf}
\showURL{%
\tempurl}


\bibitem[Song et~al\mbox{.}(2021)]%
        {song2021score}
\bibfield{author}{\bibinfo{person}{Yang Song}, \bibinfo{person}{Jascha
  Sohl-Dickstein}, \bibinfo{person}{Diederik~P Kingma},
  \bibinfo{person}{Abhishek Kumar}, \bibinfo{person}{Stefano Ermon}, {and}
  \bibinfo{person}{Ben Poole}.} \bibinfo{year}{2021}\natexlab{}.
\newblock \showarticletitle{Score-Based Generative Modeling Through Stochastic
  Differential Equations}. In \bibinfo{booktitle}{\emph{Proceedings of the
  International Conference on Learning Representations}}
  \emph{(\bibinfo{series}{ICLR '21})}.
\newblock
\urldef\tempurl%
\url{https://openreview.net/forum?id=PxTIG12RRHS}
\showURL{%
\tempurl}


\bibitem[Starke et~al\mbox{.}(2022)]%
        {starke2022deep}
\bibfield{author}{\bibinfo{person}{Sebastian Starke}, \bibinfo{person}{Ian
  Mason}, {and} \bibinfo{person}{Taku Komura}.}
  \bibinfo{year}{2022}\natexlab{}.
\newblock \showarticletitle{{D}eep{P}hase: Periodic Autoencoders for Learning
  Motion Phase Manifolds}.
\newblock \bibinfo{journal}{\emph{ACM Trans. Graph.}} \bibinfo{volume}{41},
  \bibinfo{number}{4}, Article \bibinfo{articleno}{136} (\bibinfo{year}{2022}),
  \bibinfo{numpages}{13}~pages.
\newblock
\urldef\tempurl%
\url{https://doi.org/10.1145/3528223.3530178}
\showDOI{\tempurl}


\bibitem[Starke et~al\mbox{.}(2020)]%
        {starke2020local}
\bibfield{author}{\bibinfo{person}{Sebastian Starke}, \bibinfo{person}{Yiwei
  Zhao}, \bibinfo{person}{Taku Komura}, {and} \bibinfo{person}{Kazi Zaman}.}
  \bibinfo{year}{2020}\natexlab{}.
\newblock \showarticletitle{Local Motion Phases for Learning Multi-Contact
  Character Movements}.
\newblock \bibinfo{journal}{\emph{ACM Trans. Graph.}} \bibinfo{volume}{39},
  \bibinfo{number}{4}, Article \bibinfo{articleno}{54} (\bibinfo{year}{2020}),
  \bibinfo{numpages}{14}~pages.
\newblock
\urldef\tempurl%
\url{https://doi.org/10.1145/3386569.3392450}
\showDOI{\tempurl}


\bibitem[Takeuchi et~al\mbox{.}(2017)]%
        {takeuchi2017speech}
\bibfield{author}{\bibinfo{person}{Kenta Takeuchi}, \bibinfo{person}{Dai
  Hasegawa}, \bibinfo{person}{Shinichi Shirakawa}, \bibinfo{person}{Naoshi
  Kaneko}, \bibinfo{person}{Hiroshi Sakuta}, {and} \bibinfo{person}{Kazuhiko
  Sumi}.} \bibinfo{year}{2017}\natexlab{}.
\newblock \showarticletitle{Speech-to-Gesture Generation: A Challenge in Deep
  Learning Approach with Bi-Directional {LSTM}}. In
  \bibinfo{booktitle}{\emph{Proceedings of the International Conference on
  Human Agent Interaction}} \emph{(\bibinfo{series}{HAI '17})}.
  \bibinfo{publisher}{ACM}.
\newblock
\urldef\tempurl%
\url{https://doi.org/10.1145/3125739.3132594}
\showDOI{\tempurl}


\bibitem[Tang et~al\mbox{.}(2018)]%
        {tang2018dance}
\bibfield{author}{\bibinfo{person}{Taoran Tang}, \bibinfo{person}{Jia Jia},
  {and} \bibinfo{person}{Hanyang Mao}.} \bibinfo{year}{2018}\natexlab{}.
\newblock \showarticletitle{Dance with Melody: An {LSTM}-Autoencoder Approach
  to Music-Oriented Dance Synthesis}. In \bibinfo{booktitle}{\emph{Proceedings
  of the ACM International Conference on Multimedia}}
  \emph{(\bibinfo{series}{MM '18})}. \bibinfo{publisher}{ACM},
  \bibinfo{pages}{1598--1606}.
\newblock
\urldef\tempurl%
\url{https://doi.org/10.1145/3240508.3240526}
\showDOI{\tempurl}


\bibitem[Taylor et~al\mbox{.}(2021)]%
        {taylor2021speech}
\bibfield{author}{\bibinfo{person}{Sarah Taylor}, \bibinfo{person}{Jonathan
  Windle}, \bibinfo{person}{David Greenwood}, {and} \bibinfo{person}{Iain
  Matthews}.} \bibinfo{year}{2021}\natexlab{}.
\newblock \showarticletitle{Speech-Driven Conversational Agents Using
  Conditional {F}low-{VAEs}}. In \bibinfo{booktitle}{\emph{Proceedings of the
  ACM European Conference on Visual Media Production}}
  \emph{(\bibinfo{series}{CVMP '21})}. \bibinfo{publisher}{ACM},
  \bibinfo{pages}{6:1--6:9}.
\newblock
\urldef\tempurl%
\url{https://doi.org/10.1145/3485441.3485647}
\showDOI{\tempurl}


\bibitem[Tevet et~al\mbox{.}(2023)]%
        {tevet2023human}
\bibfield{author}{\bibinfo{person}{Guy Tevet}, \bibinfo{person}{Sigal Raab},
  \bibinfo{person}{Brian Gordon}, \bibinfo{person}{Yonatan Shafir},
  \bibinfo{person}{Daniel Cohen-Or}, {and} \bibinfo{person}{Amit~H. Bermano}.}
  \bibinfo{year}{2023}\natexlab{}.
\newblock \showarticletitle{Human Motion Diffusion Model}. In
  \bibinfo{booktitle}{\emph{Proceedings of the International Conference on
  Learning Representations}} \emph{(\bibinfo{series}{ICLR '23})}.
\newblock
\urldef\tempurl%
\url{https://openreview.net/forum?id=SJ1kSyO2jwu}
\showURL{%
\tempurl}


\bibitem[Theis et~al\mbox{.}(2016)]%
        {theis2016note}
\bibfield{author}{\bibinfo{person}{Lucas Theis}, \bibinfo{person}{A{\"a}ron
  van~den Oord}, {and} \bibinfo{person}{Matthias Bethge}.}
  \bibinfo{year}{2016}\natexlab{}.
\newblock \showarticletitle{A Note on the Evaluation of Generative Models}.
\newblock \bibinfo{journal}{\emph{Proceedings of the International Conference
  on Learning Representations}} (\bibinfo{year}{2016}).
\newblock
\urldef\tempurl%
\url{https://arxiv.org/abs/1511.01844}
\showURL{%
\tempurl}


\bibitem[Tomczak and Welling(2018)]%
        {tomczak2018vae}
\bibfield{author}{\bibinfo{person}{Jakub Tomczak} {and} \bibinfo{person}{Max
  Welling}.} \bibinfo{year}{2018}\natexlab{}.
\newblock \showarticletitle{{VAE} with a {V}amp{P}rior}. In
  \bibinfo{booktitle}{\emph{Proceedings of the International Conference on
  Artificial Intelligence and Statistics}} \emph{(\bibinfo{series}{AISTATS
  '18})}. \bibinfo{pages}{1214--1223}.
\newblock
\urldef\tempurl%
\url{https://proceedings.mlr.press/v84/tomczak18a.html}
\showURL{%
\tempurl}


\bibitem[Tseng et~al\mbox{.}(2023)]%
        {tseng2023edge}
\bibfield{author}{\bibinfo{person}{Jonathan Tseng}, \bibinfo{person}{Rodrigo
  Castellon}, {and} \bibinfo{person}{C.~Karen Liu}.}
  \bibinfo{year}{2023}\natexlab{}.
\newblock \showarticletitle{{EDGE}: Editable Dance Generation from Music}. In
  \bibinfo{booktitle}{\emph{Proceedings of the IEEE/CVF Conference on Computer
  Vision and Pattern Recognition}} \emph{(\bibinfo{series}{CVPR '23})}.
\newblock
\urldef\tempurl%
\url{https://arxiv.org/abs/2211.10658}
\showURL{%
\tempurl}


\bibitem[Valle-P{\'e}rez et~al\mbox{.}(2021)]%
        {valle2021transflower}
\bibfield{author}{\bibinfo{person}{Guillermo Valle-P{\'e}rez},
  \bibinfo{person}{Gustav~Eje Henter}, \bibinfo{person}{Jonas Beskow},
  \bibinfo{person}{Andre Holzapfel}, \bibinfo{person}{Pierre-Yves Oudeyer},
  {and} \bibinfo{person}{Simon Alexanderson}.} \bibinfo{year}{2021}\natexlab{}.
\newblock \showarticletitle{{T}ransflower: {P}robabilistic Autoregressive Dance
  Generation with Multimodal Attention}.
\newblock \bibinfo{journal}{\emph{ACM Trans. Graph.}} \bibinfo{volume}{40},
  \bibinfo{number}{6} (\bibinfo{year}{2021}), \bibinfo{pages}{1:1--1:13}.
\newblock
\urldef\tempurl%
\url{https://doi.org/10.1145/3478513.3480570}
\showDOI{\tempurl}


\bibitem[van~den Oord et~al\mbox{.}(2016)]%
        {oord2016wavenet}
\bibfield{author}{\bibinfo{person}{A{\"a}ron van~den Oord},
  \bibinfo{person}{Sander Dieleman}, \bibinfo{person}{Heiga Zen},
  \bibinfo{person}{Karen Simonyan}, \bibinfo{person}{Oriol Vinyals},
  \bibinfo{person}{Alex Graves}, \bibinfo{person}{Nal Kalchbrenner},
  \bibinfo{person}{Andrew Senior}, {and} \bibinfo{person}{Koray Kavukcuoglu}.}
  \bibinfo{year}{2016}\natexlab{}.
\newblock \showarticletitle{{W}ave{N}et: A Generative Model for Raw Audio}.
\newblock \bibinfo{journal}{\emph{arXiv preprint arXiv:1609.03499}}
  (\bibinfo{year}{2016}).
\newblock


\bibitem[Vaswani et~al\mbox{.}(2017)]%
        {vaswani2017attention}
\bibfield{author}{\bibinfo{person}{Ashish Vaswani}, \bibinfo{person}{Noam
  Shazeer}, \bibinfo{person}{Niki Parmar}, \bibinfo{person}{Jakob Uszkoreit},
  \bibinfo{person}{Llion Jones}, \bibinfo{person}{Aidan~N. Gomez},
  \bibinfo{person}{{\L}ukasz Kaiser}, {and} \bibinfo{person}{Illia
  Polosukhin}.} \bibinfo{year}{2017}\natexlab{}.
\newblock \showarticletitle{Attention Is All You Need}. In
  \bibinfo{booktitle}{\emph{Advances in Neural Information Processing Systems}}
  \emph{(\bibinfo{series}{NIPS '17})}. \bibinfo{pages}{5998--6008}.
\newblock
\urldef\tempurl%
\url{https://papers.nips.cc/paper/7181-attention-is-all-you-need}
\showURL{%
\tempurl}


\bibitem[Voleti et~al\mbox{.}(2022)]%
        {voleti2022mcvd}
\bibfield{author}{\bibinfo{person}{Vikram Voleti}, \bibinfo{person}{Alexia
  Jolicoeur-Martineau}, {and} \bibinfo{person}{Christopher Pal}.}
  \bibinfo{year}{2022}\natexlab{}.
\newblock \showarticletitle{{MCVD} -- Masked Conditional Video Diffusion for
  Prediction, Generation, and Interpolation}. In
  \bibinfo{booktitle}{\emph{Advances in Neural Information Processing Systems}}
  \emph{(\bibinfo{series}{NeurIPS '22})}.
\newblock
\urldef\tempurl%
\url{https://proceedings.neurips.cc/paper_files/paper/2022/file/944618542d80a63bbec16dfbd2bd689a-Paper-Conference.pdf}
\showURL{%
\tempurl}


\bibitem[Wagner et~al\mbox{.}(2014)]%
        {wagner2014gesture}
\bibfield{author}{\bibinfo{person}{Petra Wagner}, \bibinfo{person}{Zofia
  Malisz}, {and} \bibinfo{person}{Stefan Kopp}.}
  \bibinfo{year}{2014}\natexlab{}.
\newblock \showarticletitle{Gesture and Speech in Interaction: An Overview}.
\newblock \bibinfo{journal}{\emph{Speech Commun.}}  \bibinfo{volume}{57}
  (\bibinfo{year}{2014}), \bibinfo{pages}{209--232}.
\newblock
\urldef\tempurl%
\url{https://doi.org/10.1016/j.specom.2013.09.008}
\showDOI{\tempurl}


\bibitem[Wang et~al\mbox{.}(2021)]%
        {wang2021integrated}
\bibfield{author}{\bibinfo{person}{Siyang Wang}, \bibinfo{person}{Simon
  Alexanderson}, \bibinfo{person}{Joakim Gustafson}, \bibinfo{person}{Jonas
  Beskow}, \bibinfo{person}{Gustav~Eje Henter}, {and} \bibinfo{person}{{\'E}va
  Sz{\'e}kely}.} \bibinfo{year}{2021}\natexlab{}.
\newblock \showarticletitle{Integrated Speech and Gesture Synthesis}. In
  \bibinfo{booktitle}{\emph{Proceedings of the ACM International Conference on
  Multimodal Interaction}} \emph{(\bibinfo{series}{ICMI '21})}.
  \bibinfo{publisher}{ACM}, \bibinfo{pages}{177--185}.
\newblock
\urldef\tempurl%
\url{https://doi.org/10.1145/3462244.3479914}
\showDOI{\tempurl}


\bibitem[Wennberg and Henter(2021)]%
        {wennberg2021case}
\bibfield{author}{\bibinfo{person}{Ulme Wennberg} {and}
  \bibinfo{person}{Gustav~Eje Henter}.} \bibinfo{year}{2021}\natexlab{}.
\newblock \showarticletitle{The Case for Translation-Invariant Self-Attention
  in Transformer-Based Language Models}. In
  \bibinfo{booktitle}{\emph{Proceedings of the Annual Meeting of the
  Association for Computational Linguistics and the International Joint
  Conference on Natural Language Processing}}
  \emph{(\bibinfo{series}{ACL-IJCNLP '21}, Vol.~\bibinfo{volume}{2})}.
  \bibinfo{publisher}{ACL}, \bibinfo{pages}{130--140}.
\newblock
\urldef\tempurl%
\url{https://doi.org/10.18653/v1/2021.acl-short.18}
\showDOI{\tempurl}


\bibitem[Wu et~al\mbox{.}(2021a)]%
        {wu2021modeling}
\bibfield{author}{\bibinfo{person}{Bowen Wu}, \bibinfo{person}{Chaoran Liu},
  \bibinfo{person}{Carlos~T. Ishi}, {and} \bibinfo{person}{Hiroshi Ishiguro}.}
  \bibinfo{year}{2021}\natexlab{a}.
\newblock \showarticletitle{Modeling the Conditional Distribution of Co-Speech
  Upper Body Gesture Jointly Using Conditional-{GAN} and Unrolled-{GAN}}.
\newblock \bibinfo{journal}{\emph{Electronics}} \bibinfo{volume}{10},
  \bibinfo{number}{3} (\bibinfo{year}{2021}), \bibinfo{pages}{228}.
\newblock
\urldef\tempurl%
\url{https://doi.org/10.3390/electronics10030228}
\showDOI{\tempurl}


\bibitem[Wu et~al\mbox{.}(2021b)]%
        {wu2021probabilistic}
\bibfield{author}{\bibinfo{person}{Bowen Wu}, \bibinfo{person}{Chaoran Liu},
  \bibinfo{person}{Carlos~T. Ishi}, {and} \bibinfo{person}{Hiroshi Ishiguro}.}
  \bibinfo{year}{2021}\natexlab{b}.
\newblock \showarticletitle{Probabilistic Human-Like Gesture Synthesis from
  Speech Using {GRU}-Based {WGAN}}. In \bibinfo{booktitle}{\emph{Companion
  Publication of the International Conference on Multimodal Interaction}}
  \emph{(\bibinfo{series}{ICMI '21 Companion})}. \bibinfo{publisher}{ACM},
  \bibinfo{pages}{194--201}.
\newblock
\urldef\tempurl%
\url{https://doi.org/10.1145/3461615.3485407}
\showDOI{\tempurl}


\bibitem[Xie et~al\mbox{.}(2022)]%
        {xie2022learning}
\bibfield{author}{\bibinfo{person}{Zhaoming Xie}, \bibinfo{person}{Sebastian
  Starke}, \bibinfo{person}{Hung~Yu Ling}, {and} \bibinfo{person}{Michiel
  van~de Panne}.} \bibinfo{year}{2022}\natexlab{}.
\newblock \showarticletitle{Learning Soccer Juggling Skills with Layer-Wise
  Mixture-of-Experts}. In \bibinfo{booktitle}{\emph{ACM Special Interest Group
  on Computer Graphics and Interactive Techniques Conference Proceedings}}
  \emph{(\bibinfo{series}{SIGGRAPH '22})}. \bibinfo{publisher}{ACM}, Article
  \bibinfo{articleno}{25}, \bibinfo{numpages}{9}~pages.
\newblock
\urldef\tempurl%
\url{https://doi.org/10.1145/3528233.3530735}
\showDOI{\tempurl}


\bibitem[Yazdian et~al\mbox{.}(2022)]%
        {yazdian2022gesture2vec}
\bibfield{author}{\bibinfo{person}{Payam~Jome Yazdian}, \bibinfo{person}{Mo
  Chen}, {and} \bibinfo{person}{Angelica Lim}.}
  \bibinfo{year}{2022}\natexlab{}.
\newblock \showarticletitle{{G}esture2{V}ec: Clustering Gestures Using
  Representation Learning Methods for Co-Speech Gesture Generation\balance}. In
  \bibinfo{booktitle}{\emph{Proceedings of the IEEE/RSJ International
  Conference on Intelligent Robots and Systems}} \emph{(\bibinfo{series}{IROS
  '22})}. \bibinfo{publisher}{IEEE}.
\newblock
\urldef\tempurl%
\url{https://doi.org/10.1109/IROS47612.2022.9981117}
\showDOI{\tempurl}


\bibitem[Ye et~al\mbox{.}(2021)]%
        {ye2021human}
\bibfield{author}{\bibinfo{person}{Zijie Ye}, \bibinfo{person}{Haozhe Wu},
  {and} \bibinfo{person}{Jia Jia}.} \bibinfo{year}{2021}\natexlab{}.
\newblock \showarticletitle{Human Motion Modeling with Deep Learning: A
  Survey}.
\newblock \bibinfo{journal}{\emph{AI Open}}  \bibinfo{volume}{3}
  (\bibinfo{year}{2021}), \bibinfo{pages}{35--39}.
\newblock
\urldef\tempurl%
\url{https://doi.org/10.1016/j.aiopen.2021.12.002}
\showDOI{\tempurl}


\bibitem[Ye et~al\mbox{.}(2020)]%
        {ye2020choreonet}
\bibfield{author}{\bibinfo{person}{Zijie Ye}, \bibinfo{person}{Haozhe Wu},
  \bibinfo{person}{Jia Jia}, \bibinfo{person}{Yaohua Bu}, \bibinfo{person}{Wei
  Chen}, \bibinfo{person}{Fanbo Meng}, {and} \bibinfo{person}{Yanfeng Wang}.}
  \bibinfo{year}{2020}\natexlab{}.
\newblock \showarticletitle{{C}horeo{N}et: Towards Music to Dance Synthesis
  with Choreographic Action Unit}. In \bibinfo{booktitle}{\emph{Proceedings of
  the ACM International Conference on Multimedia}} \emph{(\bibinfo{series}{MM
  '20})}. \bibinfo{publisher}{ACM}, \bibinfo{pages}{744--752}.
\newblock
\urldef\tempurl%
\url{https://doi.org/10.1145/3394171.3414005}
\showDOI{\tempurl}


\bibitem[Yin et~al\mbox{.}(2023)]%
        {yin2022dance}
\bibfield{author}{\bibinfo{person}{Wenjie Yin}, \bibinfo{person}{Hang Yin},
  \bibinfo{person}{Kim Baraka}, \bibinfo{person}{Danica Kragic}, {and}
  \bibinfo{person}{Mårten Björkman}.} \bibinfo{year}{2023}\natexlab{}.
\newblock \showarticletitle{Dance Style Transfer with Cross-Modal Transformer}.
  In \bibinfo{booktitle}{\emph{Proceedings of the IEEE/CVF Winter Conference on
  Applications of Computer Vision}} \emph{(\bibinfo{series}{WACV '23})}.
  \bibinfo{pages}{5047--5056}.
\newblock
\urldef\tempurl%
\url{https://doi.org/10.1109/WACV56688.2023.00503}
\showDOI{\tempurl}


\bibitem[Yoon et~al\mbox{.}(2020)]%
        {yoon2020speech}
\bibfield{author}{\bibinfo{person}{Youngwoo Yoon}, \bibinfo{person}{Bok Cha},
  \bibinfo{person}{Joo-Haeng Lee}, \bibinfo{person}{Minsu Jang},
  \bibinfo{person}{Jaeyeon Lee}, \bibinfo{person}{Jaehong Kim}, {and}
  \bibinfo{person}{Geehyuk Lee}.} \bibinfo{year}{2020}\natexlab{}.
\newblock \showarticletitle{Speech Gesture Generation from the Trimodal Context
  of Text, Audio, and Speaker Identity}.
\newblock \bibinfo{journal}{\emph{ACM Trans. Graph.}} \bibinfo{volume}{39},
  \bibinfo{number}{6} (\bibinfo{year}{2020}).
\newblock


\bibitem[Yoon et~al\mbox{.}(2022)]%
        {yoon2022genea}
\bibfield{author}{\bibinfo{person}{Youngwoo Yoon}, \bibinfo{person}{Pieter
  Wolfert}, \bibinfo{person}{Taras Kucherenko}, \bibinfo{person}{Carla Viegas},
  \bibinfo{person}{Teodor Nikolov}, \bibinfo{person}{Mihail Tsakov}, {and}
  \bibinfo{person}{Gustav~Eje Henter}.} \bibinfo{year}{2022}\natexlab{}.
\newblock \showarticletitle{{T}he {GENEA} {C}hallenge 2022: {A} Large
  Evaluation of Data-Driven Co-Speech Gesture Generation}. In
  \bibinfo{booktitle}{\emph{Proceedings of the ACM International Conference on
  Multimodal Interaction}} \emph{(\bibinfo{series}{ICMI '22})}.
  \bibinfo{publisher}{ACM}, \bibinfo{pages}{736--747}.
\newblock
\urldef\tempurl%
\url{https://doi.org/10.1145/3536221.3558058}
\showDOI{\tempurl}


\bibitem[Zhang et~al\mbox{.}(2023)]%
        {zhang2023diffmotion}
\bibfield{author}{\bibinfo{person}{Fan Zhang}, \bibinfo{person}{Naye Ji},
  \bibinfo{person}{Fuxing Gao}, {and} \bibinfo{person}{Yongping Li}.}
  \bibinfo{year}{2023}\natexlab{}.
\newblock \showarticletitle{{D}iff{M}otion: Speech-Driven Gesture Synthesis
  Using Denoising Diffusion Model}. In \bibinfo{booktitle}{\emph{Proceedings of
  the International Conference on Multimedia Modeling}}
  \emph{(\bibinfo{series}{MMM '23})}. \bibinfo{pages}{231--242}.
\newblock
\urldef\tempurl%
\url{https://doi.org/10.1007/978-3-031-27077-2_18}
\showDOI{\tempurl}


\bibitem[Zhang et~al\mbox{.}(2018)]%
        {zhang2018mode}
\bibfield{author}{\bibinfo{person}{He Zhang}, \bibinfo{person}{Sebastian
  Starke}, \bibinfo{person}{Taku Komura}, {and} \bibinfo{person}{Jun Saito}.}
  \bibinfo{year}{2018}\natexlab{}.
\newblock \showarticletitle{Mode-Adaptive Neural Networks for Quadruped Motion
  Control}.
\newblock \bibinfo{journal}{\emph{ACM Trans. Graph.}} \bibinfo{volume}{37},
  \bibinfo{number}{4}, Article \bibinfo{articleno}{145} (\bibinfo{year}{2018}),
  \bibinfo{numpages}{11}~pages.
\newblock
\urldef\tempurl%
\url{https://doi.org/10.1145/3197517.3201366}
\showDOI{\tempurl}


\bibitem[Zhang et~al\mbox{.}(2022a)]%
        {zhang2022motiondiffuse}
\bibfield{author}{\bibinfo{person}{Mingyuan Zhang}, \bibinfo{person}{Zhongang
  Cai}, \bibinfo{person}{Liang Pan}, \bibinfo{person}{Fangzhou Hong},
  \bibinfo{person}{Xinying Guo}, \bibinfo{person}{Lei Yang}, {and}
  \bibinfo{person}{Ziwei Liu}.} \bibinfo{year}{2022}\natexlab{a}.
\newblock \showarticletitle{{M}otion{D}iffuse: Text-Driven Human Motion
  Generation with Diffusion Model}.
\newblock \bibinfo{journal}{\emph{arXiv preprint arXiv:2208.15001}}
  (\bibinfo{year}{2022}).
\newblock
\urldef\tempurl%
\url{https://arxiv.org/abs/2208.15001}
\showURL{%
\tempurl}


\bibitem[Zhang et~al\mbox{.}(2022b)]%
        {zhang2022music}
\bibfield{author}{\bibinfo{person}{Mingao Zhang}, \bibinfo{person}{Changhong
  Liu}, \bibinfo{person}{Yong Chen}, \bibinfo{person}{Zhenchun Lei}, {and}
  \bibinfo{person}{Mingwen Wang}.} \bibinfo{year}{2022}\natexlab{b}.
\newblock \showarticletitle{Music-to-Dance Generation with Multiple
  {C}onformer}. In \bibinfo{booktitle}{\emph{Proceedings of the International
  Conference on Multimedia Retrieval}} \emph{(\bibinfo{series}{ICMR '22})}.
  \bibinfo{publisher}{ACM}, \bibinfo{pages}{34--38}.
\newblock
\urldef\tempurl%
\url{https://doi.org/10.1145/3512527.3531430}
\showDOI{\tempurl}


\bibitem[Zhao et~al\mbox{.}(2022)]%
        {zhao2022egsde}
\bibfield{author}{\bibinfo{person}{Min Zhao}, \bibinfo{person}{Fan Bao},
  \bibinfo{person}{Chongxuan Li}, {and} \bibinfo{person}{Jun Zhu}.}
  \bibinfo{year}{2022}\natexlab{}.
\newblock \showarticletitle{{EGSDE}: Unpaired Image-to-Image Translation via
  Energy-Guided Stochastic Differential Equations}. In
  \bibinfo{booktitle}{\emph{Advances in Neural Information Processing Systems}}
  \emph{(\bibinfo{series}{NeurIPS '22})}. \bibinfo{pages}{3609--3623}.
\newblock
\urldef\tempurl%
\url{https://proceedings.neurips.cc/paper_files/paper/2022/file/177d68f4adef163b7b123b5c5adb3c60-Paper-Conference.pdf}
\showURL{%
\tempurl}


\bibitem[Zhou et~al\mbox{.}(2022)]%
        {zhou2022gesturemaster}
\bibfield{author}{\bibinfo{person}{Chi Zhou}, \bibinfo{person}{Tengyue Bian},
  {and} \bibinfo{person}{Kang Chen}.} \bibinfo{year}{2022}\natexlab{}.
\newblock \showarticletitle{{G}esture{M}aster: Graph-Based Speech-Driven
  Gesture Generation}. In \bibinfo{booktitle}{\emph{Proceedings of the ACM
  International Conference on Multimodal Interaction}}
  \emph{(\bibinfo{series}{ICMI '22})}. \bibinfo{publisher}{ACM},
  \bibinfo{pages}{764--770}.
\newblock
\urldef\tempurl%
\url{https://doi.org/10.1145/3536221.3558063}
\showDOI{\tempurl}


\bibitem[Zhu et~al\mbox{.}(2022)]%
        {zhu2022motionbert}
\bibfield{author}{\bibinfo{person}{Wentao Zhu}, \bibinfo{person}{Xiaoxuan Ma},
  \bibinfo{person}{Zhaoyang Liu}, \bibinfo{person}{Libin Liu},
  \bibinfo{person}{Wayne Wu}, {and} \bibinfo{person}{Yizhou Wang}.}
  \bibinfo{year}{2022}\natexlab{}.
\newblock \showarticletitle{{M}otion{BERT}: Unified Pretraining for Human
  Motion Analysis}.
\newblock \bibinfo{journal}{\emph{arXiv preprint arXiv:2210.06551}}
  (\bibinfo{year}{2022}).
\newblock
\urldef\tempurl%
\url{https://arxiv.org/abs/2210.06551}
\showURL{%
\tempurl}


\bibitem[Zhuang et~al\mbox{.}(2022)]%
        {zhuang2022music2dance}
\bibfield{author}{\bibinfo{person}{Wenlin Zhuang}, \bibinfo{person}{Congyi
  Wang}, \bibinfo{person}{Jinxiang Chai}, \bibinfo{person}{Yangang Wang},
  \bibinfo{person}{Ming Shao}, {and} \bibinfo{person}{Siyu Xia}.}
  \bibinfo{year}{2022}\natexlab{}.
\newblock \showarticletitle{{M}usic2{D}ance: {D}ance{N}et for Music-Driven
  Dance Generation}.
\newblock \bibinfo{journal}{\emph{ACM T.. Multim. Comput.}}
  \bibinfo{volume}{18}, \bibinfo{number}{2}, Article \bibinfo{articleno}{65}
  (\bibinfo{year}{2022}), \bibinfo{numpages}{21}~pages.
\newblock
\urldef\tempurl%
\url{https://doi.org/10.1145/3485664}
\showDOI{\tempurl}


\end{thebibliography}

\appendix

\clearpage
\section{Experimental details}
\subsection{Data processing and model training}
Each and every input and output feature was standardised to mean zero and standard deviation one, save for binary features on $\{0,\,1\}$.
For the gesture datasets we used the original skeletons in the corresponding database, while the other datasets were retargeted to all use a shared skeleton.
The data was augmented by lateral mirroring, and by time stretching the sequences by a random factor chosen uniformly on $[0.9,\,1.1]$ for 20\% of the training updates on the dancing and MMA data, and 10\% of the updates on other datasets.
Lateral mirroring was not applied to the data from the specific locomotion styles that distinguish between left and right.
\balance

Finger data, if present, were not included in the modelling, since finger motion capture is
difficult and often suffers from quality issues.
We instead used a fixed hand pose in all experiments.
This is quite standard even for gestures, with all submissions to the 2020 GENEA gesture-generation challenge \cite{kucherenko2021large} and a majority of submissions to the 2022 GENEA challenge \cite{yoon2022genea} (including the best-performing deep generative model \cite{ghorbani2022exemplar}) using a fixed hand pose in their generated gestures.

During hyperparameter tuning, we also noticed a subtle trade-off between the number of residual blocks and the number of Conformers stacked within each block, when keeping the total number of parameters approximately fixed.
When using $L=10$ blocks with 4 Conformer layers in each, motion was more deliberate.
When instead using $L=20$ blocks with 2 Conformer layers each, the overall amount of motion increased.
Subjectively, we found more deliberate motion to be appropriate in most scenarios (where we thus used the former hyperparameter setting), whereas dance instead benefited from an increased amount of motion overall (where we thus used the latter hyperparameter setting).
However, the actual difference is quite subtle, and both settings give similar-looking results and similar motion quality on both data types.

Our final models after hyperparameter tuning used a $\beta_n$ range of 0.0074 to 0.78 with linear noise schedule across $N=100$ diffusion steps on the gesture datasets and a range of 0.01 to 0.7 and $N=150$ steps on the other datasets.
Except for on the MMA data, our proposed models used 8 heads, 256 attention channels, 512-dimensional embeddings, and 1024 channels in the feedforward and Conformer networks, and a dilation cycle of length $\bar{l}=3$.
(The one model we trained using the original DiffWave architecture, and thus without Conformers, used a dilation cycle of length 10 instead.)
The model trained on the MMA data used 512 channels in the feedforward and Conformer networks and a reduced embedding dimension, as an adaptation to the smaller dataset size.
Training used the Adam optimiser \cite{kingma2015adam} with $\mathrm{lr}_{\mathrm{max}}=\num{6e-4}$, 3k warm-up steps and a learning rate decay factor of \num{0.5e-5} per 10 iterations, meaning that the learning rate was multiplied by $(1-0.5\cdot10^{-5})$ every ten model updates.

Unlike the proposed systems, the ZE baseline in the evaluation on the ZeroEGGS data does not use one-hot encoding for style input, but instead requires a motion clip as its style-conditioning input.
This clip is then translated into a latent vector by the encoder.
To best match the ZE training procedure, we derived these style inputs from randomly sampled sequences of lengths 256--512 taken from a training video of the given style.

\subsection{User studies}
User study participants were recruited from the US, Canada, UK, Ireland, Australia, and New Zealand using Prolific.
Participants were required to be fluent in English.
They were asked to use headphones unless the videos in the experiment were silent.
All experiments ran in a web browser and presented 36 20-second comparison videos to each participant, unless stated otherwise.
The user studies were run in a web browser via the jsPsych package\footnote{\href{https://jspsych.org}{https://jspsych.org}}.

Except where stated otherwise, each participant was exposed to motion from each audio clip once, with the order of these clips and of the two videos in each pair being randomised for each participant.
The median completion time for the experiments was 15 minutes.
The median hourly compensation was approximately 12 GBP.

Attention checks occurred at two random points in each experiment, consisting of the spoken message ``Attention: please select the rightmost option'', but for experiments where audio was omitted from the stimuli
the same message was instead displayed as text in the lower part of the video during the second half of the clip.
Subjects that failed both attention checks were disqualified and their data not used.
(Prolific's policies do not permit disqualifying subjects based on a single failed attention check in these tests.) 
Responses given to attention-check stimuli were not included in the analyses.

\end{document}